\documentclass[10pt,twocolumn,letterpaper]{article}

\usepackage[dvipsnames]{xcolor}
\usepackage{colortbl} 
\usepackage{graphicx}
\usepackage{booktabs}
\usepackage{multirow}
\usepackage{multicol}
\usepackage{caption}
\usepackage{subcaption}
\usepackage{amssymb}
\usepackage{pifont}
\usepackage{gensymb}
\usepackage{wrapfig}
\usepackage{arydshln}

\definecolor{colorid}{RGB}{255, 245, 180}   
\definecolor{colorcd}{RGB}{180, 220, 255}   
\definecolor{colorod}{RGB}{255, 200, 200}   
\definecolor{trainhighlight}{RGB}{255, 245, 180} 
\newcommand{\dsid}[1]{\colorbox{colorid}{\textbf{#1}}}  
\newcommand{\dscd}[1]{\colorbox{colorcd}{\textbf{#1}}}  
\newcommand{\dsod}[1]{\colorbox{colorod}{\textbf{#1}}}  

\newcommand{\method}{GGPT\xspace}
\usepackage{bm} 
\usepackage[pagenumbers]{cvpr} 

\usepackage{xparse}
\usepackage{arydshln}

\definecolor{cvprblue}{rgb}{0.21,0.49,0.74}
\usepackage[pagebackref,breaklinks,colorlinks,allcolors=cvprblue]{hyperref}

\def\paperID{40744} 
\def\confName{CVPR}
\def\confYear{2026}

\title{\method: Geometry-Grounded Point Transformer}

\author{Yutong Chen$^1$, Yiming Wang$^1$, Xucong Zhang$^{1,2}$, Sergey Prokudin$^{1}$, Siyu Tang$^1$
\\ETH Zurich$^1$; Delft University of Technology$^2$}

\begin{document}

\twocolumn[{%
  \renewcommand\twocolumn[1][]{#1}
  \maketitle
  \begin{center}
    \includegraphics[width=0.99\textwidth]{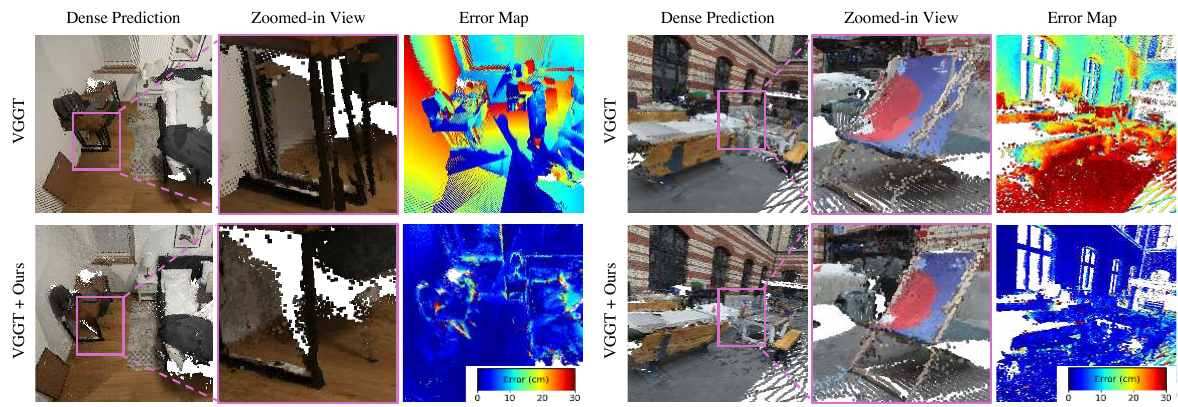}
    \captionof{figure}{Top row: the dense point maps predicted by feed-forward methods (\eg VGGT~\cite{wang2025vggt}) struggle with multi-view geometric consistency resulting in large error.
    Bottom row: with geometric guidance, our \method refines the dense point maps to enhance global alignment and 3D consistency, substantially reducing the reconstruction error.}
    \label{fig:teaser}
  \end{center}
}]

\begin{abstract}

\noindent Recent feed-forward networks have achieved remarkable progress in sparse-view 3D reconstruction by predicting dense point maps directly from RGB images. 
However, they often suffer from geometric inconsistencies and limited fine-grained accuracy due to the absence of explicit multi-view constraints. 
We introduce the \textbf{G}eometry-\textbf{G}rounded \textbf{P}oint \textbf{T}ransformer (\method), a framework that augments feed-forward reconstruction with reliable sparse geometric guidance. 
We first propose an improved Structure-from-Motion pipeline based on dense feature matching and lightweight geometric optimisation to efficiently estimate accurate camera poses and partial 3D point clouds from sparse input views.
Building on this foundation, we propose a geometry-guided 3D point transformer that refines dense point maps under explicit partial-geometry supervision using an optimised guidance encoding. 
Extensive experiments demonstrate that our method provides a principled mechanism for integrating geometric priors with dense feed-forward predictions, producing reconstructions that are both geometrically consistent and spatially complete, recovering fine structures and filling gaps in textureless areas.
Trained solely on ScanNet++ with VGGT predictions, \method generalises across architectures and datasets, substantially outperforming state-of-the-art feed-forward 3D reconstruction models in both in-domain and out-of-domain settings. Project website: \url{https://chenyutongthu.github.io/research/ggpt}.

\end{abstract} 
\section{Introduction}
\label{sec:intro}
In recent years, feed-forward 3D reconstruction networks have rapidly emerged, aiming to recover complete scene geometry directly from sparse RGB inputs. Starting with DUSt3R~\cite{wang2024dust3r}, which first demonstrated that a single transformer can jointly predict camera poses and dense 3D point maps from uncalibrated image pairs, subsequent works such as MASt3R~\cite{mast3r_eccv24} and VGGT~\cite{wang2025vggt} have further advanced this paradigm. Trained on large-scale multi-view datasets, these vision transformers can now produce dense, visually coherent reconstructions in a single forward pass. As a result, feed-forward 3D reconstruction has become one of the most promising directions in 3D vision, offering fast, scalable, and unified reconstruction pipelines.

Yet, despite their impressive visual results, closer inspection reveals that these models still struggle with multi-view geometric consistency~(see Fig.~\ref{fig:teaser}). Their reconstructions often exhibit multi-layer artifacts and deviate from the ground truth, particularly when applied outside their training distribution. In practice, although these networks can predict plausible 3D structures from limited views, they often fail to recover geometry that is sufficiently accurate and consistent across viewpoints. This limitation becomes evident in out-of-domain scenarios, such as medical or surgical scenes, or human data, where feed-forward predictions can deviate substantially from the true underlying geometry.

In contrast, Structure-from-Motion (SfM)~\cite{revisitedsfm} remains firmly grounded in geometric principles~\cite{hartley2003multiple}, producing geometrically consistent reconstructions. However, SfM pipelines remain fragile under wide baselines, low overlap, or limited viewpoints, and typically recover only sparse geometry structure. This contrast highlights a clear opportunity: to combine the completeness and efficiency of feed-forward 3D reconstruction with the geometric accuracy and generalisation of SfM.

Building on this motivation, several recent studies have explored similar ideas~\cite{guo2025murre, zuo2025omni, pow3r_cvpr25, keetha2025mapanything}, showing that incorporating sparse geometric guidance can improve dense feed-forward reconstructions. However, existing approaches remain limited in two key aspects. {\it First}, they often rely on unrealistic SfM guidance, such as pseudo SfM points sampled from ground truth~\cite{pow3r_cvpr25,zuo2025omni,keetha2025mapanything} or SfM results obtained from densely captured video sequences~\cite{guo2025murre}, which are rarely available in real-world sparse-view scenarios.
This dependency largely reflects the lack of robustness and efficiency of conventional SfM pipelines under limited input views. {\it Second}, prior methods refine predictions in 2D image space via depth-map diffusion~\cite{guo2025murre} or image transformers~\cite{pow3r_cvpr25,keetha2025mapanything}, which inherently constrain the model to view-dependent reasoning.
As a result, they fail to exploit the explicit 3D structure of the scene and cannot enforce true cross-view geometric consistency.

In this work, we introduce \method(\textbf{G}eometry-\textbf{G}rounded \textbf{P}oint \textbf{T}ransformer), a geometry-guided framework that refines dense feed-forward reconstructions using accurate geometric guidance obtained from an improved SfM pipeline, and does so directly in 3D space.

{\it First,} we revisit SfM under limited input views and introduce an improved pipeline that integrates dense matchers~\cite{edstedt2024roma, zhang2025ufm} with a lightweight optimisation procedure.
Concretely, we estimate camera poses from a compact set of high-confidence correspondences and then triangulate all valid matches using a direct linear transform. Compared with state-of-the-art SfM methods designed for unconstrained captures, including VGGT+BA~\cite{wang2024vggsfm,wang2025vggt} and MASt3R-SfM~\cite{duisterhof2025mastrsfm}, our SfM pipeline achieves higher accuracy thanks to the rich geometric constraints provided by the dense correspondences, and also improved efficiency thanks to the separation of non-linear BA optimisation and linear triangulation. Its robustness and efficiency also enable the construction of realistic SfM supervision without requiring densely captured multi-view sequences.

{\it Second,} we introduce a variant of a lightweight 3D Point Transformer~\cite{wu2024ptv3} that jointly processes dense point maps from feed-forward models and geometrically grounded partial point cloud from our SfM pipeline, predicting residual corrections for every point. By reasoning directly in a global 3D coordinate space, \method propagates the geometric accuracy of triangulated SfM points to dense but noisy feed-forward predictions. In contrast to prior depth completion methods that operate on 2D image tokens and channel-wise fuse geometry and image features~\cite{keetha2025mapanything,pow3r_cvpr25,zuo2025omni,guo2025murre}, \method performs attention on the two point clouds directly in 3D, where spatial proximity, not pixel coordinates, defines receptive fields. This design explicitly enforces multi-view geometric consistency and produces globally aligned and metrically coherent dense reconstructions.

We conduct extensive experiments to analyse both our improved SfM pipeline and Geometry-Grounded Point Transformer. Our SfM framework achieves better performance and efficiency than state-of-the-art SfM alternatives~\cite{wang2024vggsfm, duisterhof2025mastrsfm}, making it a practical and robust solution for applications beyond the scope of this work. When conditioned on the same SfM guidance, our proposed point transformer predicts significantly more accurate dense points than prior depth completion networks~\cite{guo2025murre,zuo2025omni,pow3r_cvpr25,keetha2025mapanything}.

Overall, thanks to its intuitive and modular design, our method can be seamlessly integrated with various feed-forward 3D reconstruction models at inference time, without requiring any fine-tuning.
Remarkably, despite being trained only on ScanNet++~\cite{yeshwanth2023scannet++} using VGGT~\cite{wang2025vggt} dense predictions, GGPT generalises effectively across architectures and datasets, substantially improving the performance of state-of-the-art feed-forward models in both in-domain and out-of-domain settings. The strong generalisation results further suggest that it can serve as a broadly useful tool for a wide range of 3D reconstruction applications, extending well beyond conventional benchmark scenarios.

\section{Related Work}
\label{sec:related_work}
\noindent\textbf{Feed-forward 3D reconstruction models.} 
Recent multi-view image models ~\cite{wang2024dust3r,spann3r,wang2025pi3,wang2025vggt,keetha2025mapanything,cut3r,zhang2024monst3r,dens3r,lu2024align3r} predict camera poses and depths directly from RGBs. Trained on large-scale 3D datasets, these models offer greater robustness and efficiency than traditional SfM~\cite{revisitedsfm} in sparse-view settings. However, lacking explicit geometric constraints, their predictions often show multi-view inconsistency and global spatial drift, especially in out-of-domain scenarios. Our method alleviates these issues by leveraging geometry estimation to ground the dense predictions.

\noindent\textbf{Structure-from-Motion.} 
COLMAP~\cite{revisitedsfm}, the traditional incremental SfM, is the standard approach to 3D reconstruction from sufficient overlapping views.  
MP-SfM~\cite{pataki2025mpsfm} enhances COLMAP with monocular priors, focusing on improving the camera pose in low-overlap cases. VGG-SfM~\cite{wang2024vggsfm} replaces the incremental pipeline with a simplified global optimisation initialised by feed-forward predicitions. More recently, VGGT+BA~\cite{wang2025vggt} improves the results by using VGGT as initialisation for a global bundle adjustment, but it only estimates very sparse points due to the expensiveness of dense BA. MASt3R-SfM~\cite{duisterhof2025mastrsfm} adopts MASt3R~\cite{mast3r_eccv24} predictions for sparse matching and initialisation, and jointly optimize cameras and depths to produce dense prediction. However, its points estimation has limited accuracy due to the sparse matches. Our SfM adopts a global optimisation instead of incremental reconstruction~\cite{revisitedsfm,pataki2025mpsfm}, which greatly simplifies the algorithm. In contrast to previous global SfM methods~\cite{wang2024vggsfm,wang2025vggt,duisterhof2025mastrsfm}, our SfM pipeline combines efficient dense matching regressors~\cite{zhang2025ufm,edstedt2024roma} with a lightweight sparse BA and dense linear triangulation, achieving improved accuracy and efficiency.

\noindent\textbf{Geometry-conditioned 3D reconstruction.}  
Previous work typically frames geometry-conditioned 3D reconstruction as a 2D depth completion task. Monocular depth completion models~\cite{Bae2022irondepth,lin2024promptda,wang2025priorda,guo2025murre,zuo2025omni} inject sparse depth maps into pretrained depth networks, but their predictions suffer from multi-view inconsistency when applied to multi-view reconstruction. Recent multi-view transformers~\cite{pow3r_cvpr25,keetha2025mapanything} accept additional depth maps, yet depend on pseudo-SfM points sampled from ground truth. In contrast, we address a more practical and challenging setting, enhancing dense predictions using SfM points derived \emph{solely} from the input RGB views. Unlike prior approaches that operate in 2D image space, our model employs a 3D point transformer to jointly process dense and sparse point clouds in 3D, yielding improved multi-view consistency and generalisation.

\noindent\textbf{3D point cloud processing.}
Unlike 2D image processing, 3D point clouds exhibit irregular and non-uniform spatial distributions and require specialised backbone designs, such as MLP-based PointNets~\cite{qi2016pointnet,qi2017pointnetplusplus}, sparse convolutional networks~\cite{minkowski}, and attention-based Point Transformers~\cite{zhao2021point,wu2024ptv3}. Early 3D networks mainly targeted scene understanding tasks, while recent work extends them to geometry-centric problems, including point cloud denoising~\cite{vogel2024p2p,rakotosaona2020pointcleannet,Liu2025OmniScene,unknownue2022pdflow,iterpfn}, completion~\cite{yu2021pointr,yuan2018pcn}, registration~\cite{Yew_2022_regtr,regformer}, sampling~\cite{yu2018punet,Ha2025HVPUNet,rong2024repkpu}, Gaussian splat processing~\cite{chen2024splatformer,lan2024gaussiananything,DiffGS}, and point tracking~\cite{rajic2025mvtracker}. 
We use point transformers to refine dense predictions under sparse geometric guidance, introducing dedicated encodings that capture positional relations and correspondence offsets between the two inputs.
This bridges feed-forward image-based reconstruction and geometry-aware 3D reasoning into a unified, spatially grounded refinement framework.

\section{Method}
\label{section:method}
\begin{figure*}[t!]
    \centering
    \includegraphics[width=1.0\linewidth]{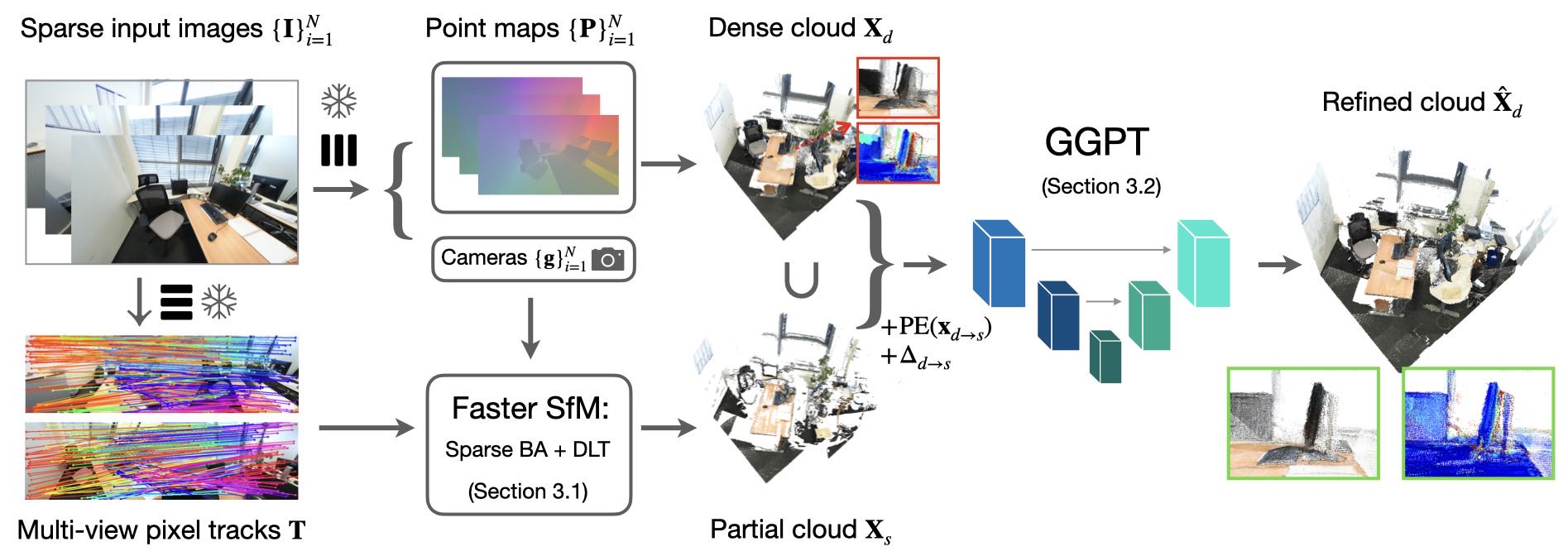}
\caption{
\emph{Overview of our method.}  
We utilise off-the-shelf feed-forward multi-view transformers~\cite{wang2025vggt,keetha2025mapanything} and dense matchers~\cite{edstedt2024roma,zhang2025ufm} to predict dense point maps $\{\mathbf{P}_i\!\in\!\mathbb{R}^{H\times W\times 3}\}_{i=1}^N$, camera parameters $\{\mathbf{g}_i\}_{i=1}^N$, and multi-view correspondences $\mathbf{T}\!\in\!\mathbb{R}^{N\times N\times H\times W\times 2}$.  
These correspondences are used for sparse bundle adjustment and multi-view DLT triangulation, producing a geometrically consistent yet incomplete sparse cloud $\mathbf{X}_s$ via a lightweight alternative to standard SfM pipelines (Section~\ref{subsec:fastba}).  
The dense yet multi-view inconsistent point cloud $\mathbf{X}_d$, obtained by combining the predicted per-view point maps, exhibits local misalignments (red boxes) that are then refined under the geometric guidance of $\mathbf{X}_s$ by our \emph{Geometry-Grounded Point Transformer (GGPT)} (Section~\ref{subsec:ptv3}), a modified point transformer \cite{wu2024ptv3} equipped with specialised geometric embeddings $\mathrm{PE}(\mathbf{x}_{d\rightarrow s})$ and $\Delta_{d\rightarrow s}$, which encode spatial relations between dense and sparse points to provide stable geometric guidance.  
The final output is a refined, globally aligned, and geometry-consistent dense point cloud $\hat{\mathbf{X}}_d$.
}

\label{fig:method}
\end{figure*}

\noindent
Our core idea is to refine dense point maps $\mathbf{X}_d$ predicted by feed-forward multi-view transformers~\cite{wang2025vggt,keetha2025mapanything,wang2025pi3} using geometrically accurate yet incomplete 3D point clouds $\mathbf{X}_s$.  As illustrated in Fig.~\ref{fig:method}, our method consists of two stages. (1) \textbf{Efficient and Robust SfM}: a lightweight and robust SfM pipeline with dense matchers and sparse BA to produce an incomplete but geometrically consistent point map $\mathbf{X}_s$; and (2) \textbf{Geometry-Grounded Point Transformer} (GGPT): a 3D point transformer to refine the feed-forward prediction $\mathbf{X}_d$ with the geometric guidance $\mathbf{X}_s$.

\subsection{Efficient and Robust SfM
}
\label{subsec:fastba}

\noindent
\textbf{Key insight.} Our SfM pipeline aims to efficiently estimate accurate camera poses and a sufficient number of points from limited observations. 
We initialise global optimisation using camera parameters and points predicted by feed-forward models, improving robustness and efficiency compared with traditional incremental SfM pipelines~\cite{revisitedsfm,pataki2025mpsfm}. 
We leverage recent dense matching regressors~\cite{edstedt2024roma,zhang2025ufm} to extract multi-view correspondences from pairwise matchings. 
Directly using all dense matchings for global non-linear optimisation is computationally expensive. Instead, we first perform sparse bundle adjustment (BA) on a compact set of high-confidence matches to estimate camera poses, followed by an efficient direct linear triangulation (DLT) to reconstruct dense points. 

\vspace{3pt}
\noindent\textbf{Initialisation from feed-forward models.}
Given N unposed RGB images 
$\{\mathbf{I}_i \in \mathbb{R}^{H\times W\times 3}\}_{i=1}^N$, a multi-view transformers $f_\theta$ can predict camera parameters and dense 3D point maps:
\begin{align}
\big\{\mathbf{g}_i,\, \mathbf{P}_i\big\}_{i=1}^N 
&= f_\theta\!\left(\{\mathbf{I}_i\}_{i=1}^N\right),
\end{align}
where each camera parameter vector
$\mathbf{g}_i = [\mathbf{q}_i,\, \mathbf{t}_i,\, \mathbf{f}_i]$ 
contains rotation (unit quaternion $\mathbf{q}_i$), translation $\mathbf{t}_i$, and intrinsics $\mathbf{f}_i = [f_x,f_y,c_x,c_y]$.  
Each dense point map 
$\mathbf{P}_i \in \mathbb{R}^{H\times W\times 3}$ 
assigns to every pixel $(u,v)$ a predicted 3D coordinate 
$\mathbf{p}_{i,uv}\in\mathbb{R}^3$ 
in the global scene frame.  
The union of all $\mathbf{P}_i$ forms the initial dense prediction 
$\mathbf{X}_d=\{\mathbf{P}_i\}$, 
later refined by our transformer-based stage.

\vspace{3pt}
\noindent\textbf{Dense feature matching.} We adopt RoMa~\cite{edstedt2024roma} and UFM~\cite{zhang2025ufm} to obtain a global correspondence tensor across all image pairs
\begin{align}
\mathbf{T}\in\mathbb{R}^{N\times N\times W\times H\times 2},\quad
\mathbf{C}\in[0,1]^{N\times N\times W\times H},
\end{align}
where $\mathbf{T}[i,k,u,v]$ denotes the predicted location in view $k$ corresponding to pixel $(u,v)$ in view $i$, and $\mathbf{C}[i,k,u,v]$ its confidence.  
Invalid or inconsistent correspondences are removed by enforcing cycle consistency:
\begin{align}
\big\|\mathbf{T}[k,i,\mathbf{T}[i,k,\mathbf{u}]] - \mathbf{u}\big\|_2 < \epsilon,
\end{align}
producing a binary mask $\mathbf{M}\in\{0,1\}^{N\times N\times W\times H}$ that defines the set of geometrically valid matches.

\vspace{3pt}
\noindent\textbf{Sparse bundle adjustment.}
From $\mathbf{T}$, $\mathbf{C}$, and $\mathbf{M}$ we derive two confidence-masked subsets for bundle adjustment and triangulation:
\begin{align}
\mathbf{M}_{\text{BA}} &= \big[\mathbf{C} > \epsilon_{\text{BA}}\big] \odot \mathbf{M}, \ 
\mathbf{M}_{\text{DLT}} &= \big[\mathbf{C} > \epsilon_{\text{DLT}}\big] \odot \mathbf{M},
\end{align}
with $\epsilon_{\text{BA}} > \epsilon_{\text{DLT}}$, ensuring that BA uses only high-confidence tracks while DLT benefits from denser correspondences.  
Applying $\mathbf{M}_{\text{BA}}$ to $\mathbf{T}$ yields the masked track set 
$\mathbf{T}_{\text{BA}} = \mathbf{T} \odot \mathbf{M}_{\text{BA}}$. Each track $\mathbf{T}[i,:,u,v]$ matches an anchor pixel $(u,v)$ in view $i$ to pixels in other views. We use $j$ to index the track anchored at $(i,u,v)$ and denote its visible views as
\[
\mathcal{V}^j = \{\,k~|~\mathbf{M}_{\text{BA}}[i,k,u,v]=1\,\},
\]
As $\mathbf{T}_{\text{BA}}$ can still contain large numbers of tracks, we further select a maximal number of anchor pixels from each view and use their associated tracks for our BA. We choose $n_{\textrm{BA}}$ anchor pixels per image with highest SuperPoint saliency~\cite{superpoint} and visible in at least two views. Using the maximal $N\times n_{\textrm{BA}}$ tracks, we jointly refine cameras and the anchor points by minimising the 2D reprojection loss:
\begin{align}
\{\mathbf{G}^*,\,\mathbf{X}^*\}
&=
\arg\min_{\{\mathbf{g}_i\},\,\{\mathbf{x}^j\}}
\sum_j \sum_{i\in\mathcal{V}^j}
\mathcal{L}_{\textrm{r}}\!\bigl(
\pi(\mathbf{g}_i,\mathbf{x}^j)-\mathbf{u}_i^j
\bigr).
\end{align}
where $\pi(\mathbf{g}_i,\,\mathbf{x}^j)$ projects 3D point $\mathbf{x}^j$ into view $i$ and $\mathbf{u}_i^j$ is the observation of track $j$ in the view $i$, $\mathcal{L}_{\textrm{r}}$ is the robust Cauchy loss~\cite{robustloss}.
The optimisation is initialised from $\{\mathbf{g}_i,\,\mathbf{P}_i\}$ and converges quickly thanks to the feed-forward prior.  
While both cameras and sparse points are refined, only the updated camera estimates 
$\mathbf{G}^*=\{\mathbf{g}_i^*\}$ 
are propagated to the next stage.

\vspace{3pt}
\noindent\textbf{Direct linear transform.}
Finally, we apply $\mathbf{M}_{\text{DLT}}$ to the correspondences to obtain $\mathbf{T}_{\text{DLT}} = \mathbf{T}\odot\mathbf{M}_{\text{DLT}}$ and reconstruct 3D points using all available valid correspondences jointly via a multi-view variant of the Direct Linear Transform~\cite{triangulation}.  
Each track $\mathbf{T}^j$ across views $\mathcal{V}^j$ is triangulated into a 3D point $\mathbf{x}^j$ using the refined cameras $\mathbf{G}^*$.  
Points with large reprojection error or small triangulation angle are discarded.  
The resulting set
\begin{align}
\mathbf{X}_s
= \big\{\,
\mathbf{x}^j ~\big|~
\mathbf{T}^j \subseteq \mathbf{T}_{\text{DLT}}
\,\big\}
\end{align}
constitutes our final geometrically consistent scene representation. Note that $\mathbf{X}_s$ is incomplete for regions which are only observed in single views or do not have valid matchings due to lack of saliency. Additionally, as each $\mathbf{x}^j$ associates with pixels in $\mathbf{T}^j$, it has multiple correspondences (at least two) in the feed-forward multi-view point map $\mathbf{X}_d$. Next,  we will use $\mathbf{X}_s$ as the geometric guidance to refine the inaccurate dense predictions $\mathbf{X}_d$.

\subsection{Geometry-Grounded Point Transformer}
\label{subsec:ptv3}



\noindent\textbf{Key insight.}
The geometry reconstruction $\mathbf{X}_s$ is geometrically accurate but incomplete, particularly lacking coverage in textureless or occluded regions, whereas the dense prediction $\mathbf{X}_d$ from feed-forward models is complete yet can be multi-view inconsistent. To combine their complementary strengths, we introduce the \emph{Geometry-Grounded Point Transformer}, which refines $\mathbf{X}_d$ under the guidance of $\mathbf{X}_s$. By jointly reasoning over both in a shared 3D space, GGPT transfers the geometric reliability of triangulated points to dense but inaccurate predictions. Unlike prior refinement methods~\cite{keetha2025mapanything,pow3r_cvpr25} that attend to 2D image tokens, GGPT performs attention directly in 3D, where spatial proximity rather than pixel location defines receptive fields, yielding 3D consistent dense reconstructions.

\vspace{4pt}
\noindent\textbf{Input embeddings.}
Given the two point clouds $\mathbf{X}_d\in \mathbb{R}^{NHW \times 3}$ and $\mathbf{X}_s \in \mathbb{R}^{N_s \times 3}$, we first align them using the Kabsch–Umeyama transform~\cite{umeyama2002least} and embed each point $\mathbf{x}_i$ into an initial feature vector $\mathbf{z}_i^{(0)}$.  
For $\mathbf{x}_s\in \mathbf{X}_s$,
\begin{equation}
\mathbf{z}_{s}^{(0)} = [\mathrm{PE}(\mathbf{x}_{s}),\, \mathbf{e}_{\mathrm{type}(s)}],
\end{equation}
where $\mathrm{PE}(\mathbf{x}_{s})$ is a sinusoidal positional encoding~\cite{vaswani2017attention,zhao2021point} with a frequency of 4 and $\mathbf{e}_{\mathrm{type}(s)}\in \mathbb{R}^{16}$ a learnable token marking guidance points. Some points $\mathbf{x}_d~\in \mathbf{X}_d$ have corresponding points in the guidance, denoted as $\mathbf{x}_{d\rightarrow s} \in \mathbf{X}_s$, since they originate from the same image pixels. To make the network aware of this relation, we define
\begin{equation}
\mathbf{z}_{d}^{(0)} = [\mathrm{PE}(\mathbf{x}_{d}),\, \mathbf{e}_{\mathrm{type}(d)},\, \mathrm{PE}(\mathbf{x}_{d\rightarrow s}),\, \Delta_{d\rightarrow s}],
\end{equation}
where $\mathbf{e}_{\mathrm{type}(d)}$ is a learnable token for dense points and $\Delta_{d\rightarrow s}=\mathbf{x}_{d\rightarrow s}-\mathbf{x}_{s}$ encodes positional offsets.  
For $\mathbf{x}_d$ without correspondences, we use only the first two terms.  
All embeddings are zero-padded to a uniform dimension, yielding the combined input $\mathbf{Z}^{(0)}=\{\mathbf{z}_s^{(0)}\}\cup\{\mathbf{z}_d^{(0)}\}$.

\vspace{4pt}
\noindent\textbf{Point transformer backbone.}
The embeddings $\mathbf{Z}^{(0)}$ are processed by a 3D point transformer $g_\gamma(\cdot)$ with $L$ layers:
\begin{equation}
\mathbf{Z}^{(L)} = g_\gamma(\mathbf{Z}^{(0)}),
\end{equation}
We use an $L{=}8$-layer Point Transformer~V3~\cite{wu2024ptv3} (PTv3), which applies patch-wise self-attention over spatial neighbourhood to capture fine-scale structure and long-range dependencies.  
This models point interaction by 3D proximity rather than exhaustive 2D grid attention, fusing $\mathbf{X}_d$ and $\mathbf{X}_s$ into 3D-aware features $\mathbf{Z}^{(L)}$. Our PTv3 backbone has 53M parameters in total, considerably lighter than 2D vision transformers (around 300M) used in prior geometry-conditioned methods~\cite{guo2025murre,keetha2025mapanything}.

\vspace{4pt}
\noindent\textbf{Prediction head.}
The geometry-aware features $\mathbf{Z}^{(L)}$ are decoded by a shared MLP $h_\psi(\cdot)$ with ReLU activations:
\begin{equation}
[\boldsymbol{\delta},\, \tilde{c}] = h_\psi(\mathbf{Z}^{(L)}),
\end{equation}
where $\boldsymbol{\delta}\in\mathbb{R}^3$ is the predicted residual displacement and $\tilde{c}\in\mathbb{R}$ the raw confidence.  
The refined coordinates are computed as $\hat{\mathbf{x}} = \mathbf{x} + \boldsymbol{\delta}$,
and the final confidence as $c = \exp(\tilde{c}) + 1$ following~\cite{kendall2017uncertainties,wang2024dust3r,wang2025vggt}.  
The refined dense and sparse sets are $\hat{\mathbf{X}}_d=\{\hat{\mathbf{x}}_d\}$ and $\hat{\mathbf{X}}_s=\{\hat{\mathbf{x}}_s\}$, with $\hat{\mathbf{X}}_d$ used as the final output.

\vspace{4pt}
\noindent\textbf{Patch-based processing.}
To handle large-scale point clouds efficiently, GGPT operates on spatially local patches instead of the full scene.  
We represent the dense point cloud as overlapping subsets $\mathbf{X}_d = \{\mathbf{X}_i\}_{i=1}^{N_p}$, where each patch $\mathbf{X}_i$ corresponds to a cubic region centered at a 3D location and normalised to the unit cube $[0,1]^3$.  
During \emph{training}, we randomly sample anchor points from $\mathbf{X}_s$ and extract local cubes around them. During \emph{inference}, we apply a sliding cube with some overlap to cover the whole point cloud.
Each patch is processed independently by the transformer and prediction head.  
For points included in multiple overlapping patches, their final 3D coordinates are computed by averaging the predictions in multiple patches.
Such patch-based decomposition, also used in prior point cloud denoising~\cite{vogel2024p2p}, balances computational efficiency with fine-grained geometric fidelity across large 3D scenes.

\vspace{4pt}
\noindent\textbf{Training objectives.}
GGPT is trained with the loss
\begin{equation}
\mathcal{L} = \mathcal{L}_{\mathrm{conf}}  + \lambda_{\mathrm{id}} \mathcal{L}_{\mathrm{id}}.
\end{equation}

\begin{table*}[ht!]
    \centering
    \setlength{\tabcolsep}{4pt} 
    \caption{\textbf{Multi-view 3D Reconstruction on Standard Test Sets.}
    We report AUC@5/10 cm (\% $\uparrow$). Our method (our SfM and GGPT) can improve dense predictions of various models. Despite GGPT being trained solely on ScanNet++ and VGGT's dense prediction, as we highlight the top-left cells as \dsid{within-domain}, it generalises well to \dscd{cross-domain datasets} and other methods' predictions (rows 3–10).}
    \label{tab:ma_benchmark_short}
\begin{tabular}{l ccc c ccc c ccc}
\toprule
& \multicolumn{3}{c}{\dsid{ScanNet++}~\cite{yeshwanth2023scannet++}} 
&& \multicolumn{3}{c}{\dscd{ETH3D}~\cite{schoeps2017eth3d}}
&& \multicolumn{3}{c}{\dscd{T\&T}~\cite{Knapitsch2017tandt}} \\
& 4 imgs & 8 imgs & 16 imgs && 4 imgs & 8 imgs & 16 imgs && 4 imgs & 8 imgs & 16 imgs \\
\midrule

VGGT~\cite{wang2025vggt} 
& \cellcolor{trainhighlight}23/37 & \cellcolor{trainhighlight}19/32 & \cellcolor{trainhighlight}16/29 
&& 27/41 & 23/36 & 19/32 
&& 26/40 & 25/39 & 24/38 \\
~\cite{wang2025vggt} + Ours 
& \cellcolor{trainhighlight}\textbf{38/53} & \cellcolor{trainhighlight}\textbf{45/60} & \cellcolor{trainhighlight}\textbf{50/66}
&& \textbf{41/55} & \textbf{47/61} & \textbf{49/63}
&& \textbf{34/47} & \textbf{42/57} & \textbf{43/57} \\
\midrule

Pi3~\cite{wang2025pi3} 
& \textbf{54/69} & \textbf{56}/71 & 58/\textbf{74} 
&& 31/47 & 25/41 & 23/38 
&& 25/39 & 26/42 & 25/40 \\
~\cite{wang2025pi3} + Ours
& \textbf{54}/68 & \textbf{56/72} & \textbf{59/74} 
&& \textbf{36/53} & \textbf{36/53} & \textbf{37/54} 
&& \textbf{27/43} & \textbf{32/50} & \textbf{33/50}  \\
\midrule

MapAnything~\cite{keetha2025mapanything} 
& 40/57 & 38/57 & 38/58 
&& 10/20 & 7/15 & 5/12 
&& 10/21 & 9/20 & 8/17   \\
~\cite{keetha2025mapanything} + Ours
& \textbf{44/61} & \textbf{48/64} & \textbf{52/68}
&& \textbf{32/43} & \textbf{33/45} & \textbf{34/47}
&& \textbf{29/43} & \textbf{40/55} & \textbf{42/56}   \\
\midrule

MAtCha~\cite{guedon2025matcha} 
& 12/19 & 15/26 & 18/32 
&& 40/52 & 41/53 & 42/56 
&& 34/47 & 36/50 & 33/47  \\
~\cite{guedon2025matcha} + Ours
& \textbf{27/37} & \textbf{40/52} & \textbf{48/63}
&& \textbf{42/55} & \textbf{47/60} & \textbf{50/65}
&& \textbf{35/48} & \textbf{43/57} & \textbf{43/57}  \\
\midrule

MASt3R-SfM~\cite{duisterhof2025mastrsfm}
& 12/22 & 14/35 & 16/31
&& 37/50 & 39/51 & 40/54
&& \textbf{34}/46 & 37/50 & 36/49  \\
~\cite{duisterhof2025mastrsfm} + Ours
& \textbf{30/41} & \textbf{39/52} & \textbf{48/64}
&& \textbf{41/55} & \textbf{46/59} & \textbf{48/63}
&& 33/\textbf{47} & \textbf{41/56} & \textbf{42/56}  \\
\bottomrule
\end{tabular}

\end{table*}

\begin{figure*}[t!]
    \centering
    \includegraphics[width=\linewidth, trim={0.2cm 0.1cm 0.2cm 0.0cm}, clip]{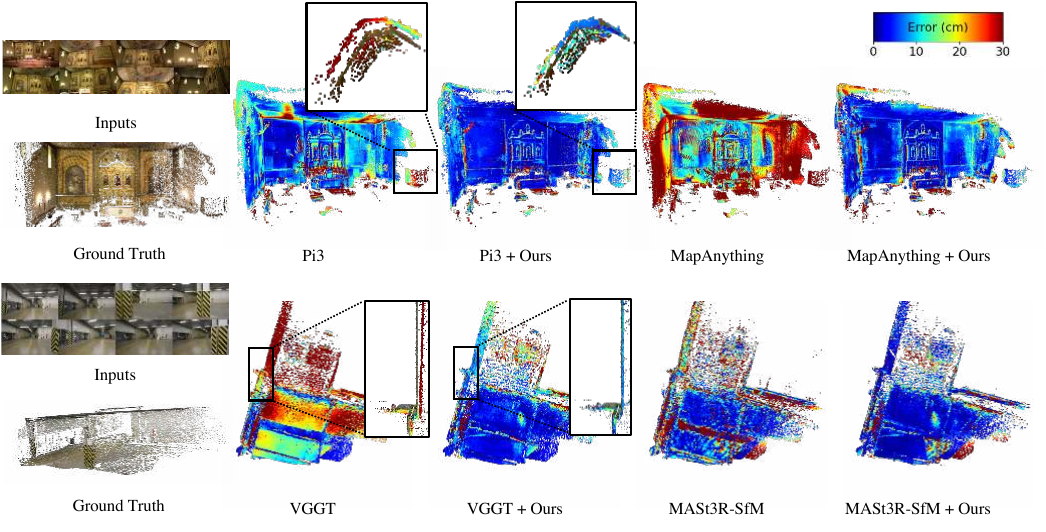}
    \caption{\textbf{3D reconstruction results on T\&T~\cite{Knapitsch2017tandt} and ETH3D~\cite{schoeps2017eth3d}}.  Points with confidence above the 90\% quantile are visualised. We compare the error maps of the reconstruction before and after our refinement. In the zoomed-in regions, the ground truth (colored by input RGBs) is overlaid with the predictions (colored by error) to highlight how our method corrects the misalignment of input points.}
    \label{fig:vis_tab1}
\end{figure*}

\noindent\textbf{Confidence-weighted regression.}
We use a heteroscedastic formulation~\cite{kendall2017uncertainties}:
\begin{equation}
\mathcal{L}_{\mathrm{conf}} =
\sum_{\mathbf{x}\in\mathbf{X}_d\cup\mathbf{X}_s}
c\,\|\hat{\mathbf{x}}-\mathbf{x}_{\mathrm{GT}}\| - \alpha\log c.
\end{equation}
The predicted confidence $c$ modulates each residual, reducing the effect of uncertain regions while the $-\alpha\log c$ term penalises trivial solution with overly low confidence.  
Ground-truth points $\mathbf{X}_{\mathrm{GT}}$ are pre-aligned to $\mathbf{X}_s$ via Umeyama alignment, ensuring supervision focuses on local corrections rather than global shifts.


\noindent\textbf{Identity consistency.}
If a point $\hat{\mathbf{x}} \in \mathbf{X}_d$ has a valid correspondence in the guidance, \ie $\mathbf{x}_{d\rightarrow s}\in \mathbf{X}_s$, we apply an anchoring term to its prediction $\hat{\mathbf{x}}$: 
\begin{equation}
\mathcal{L}_{\mathrm{id}} =
\sum_{\mathbf{x}\in\mathbf{X}_d^{d\rightarrow s}}
\|\hat{\mathbf{x}}-\mathbf{x}_{d\rightarrow s}\|,
\end{equation}
where $\mathbf{X}_d^{d\rightarrow s}$ denotes the set of points with geometry guidance. This term encourages the model to predict points aligned with the geometry guidance.

\vspace{4pt}
\noindent\textbf{Key architectural insights.}
We provide ablation studies in the supplementary to highlight two factors improving the performance of our point transformer design:
(i) incorporating partial geometry $\mathbf{X}_s$ as guidance, both as an auxiliary input and via encodings $\mathrm{PE}(\mathbf{x}_{d\rightarrow s})$ and $\Delta_{d\rightarrow s}$, and
(ii) adopting patch-based processing, which enhances efficiency while retaining fine geometric detail.

\section{Experiments }
\label{sec:exp}

\subsection{Implementation Details}
\noindent \textbf{SfM configurations.} To filter matches for BA and DLT, we set $\epsilon=4$, $\epsilon_\textrm{BA}=0.6$, $\epsilon_\textrm{DLT}=0.1, n_\textrm{BA}=2048$. After DLT, we filter out points with reprojection error above 4 pixels and maximal triangulation angle below 3 degrees.

\noindent \textbf{GGPT training.} The training set has 20k multi-view sequences sampled from 856 training scenes in ScanNet++~\cite{yeshwanth2023scannet++}. We use VGGT~\cite{wang2025vggt} for $\textbf{X}_d$ prediction and SfM initialisation. Each input patch has a half width of $0.4\times$ the scene radius. Each forward pass processes up to 400k points. We set $\lambda_{id}=1, \alpha=0.2$. Training is performed on 8 NVIDIA GH200 GPUs for one day. 

\noindent \textbf{Inference runtime.} We provide a detailed runtime breakdown for different input views in the Sec. C of the supplementary, which shows that dense matching~\cite{edstedt2024roma,zhang2025ufm} dominates the runtime while our proposed BA, DLT, and GGPT refinement add minor overhead. 

\noindent \textbf{Evaluation protocols.}
We align the prediction to the metric ground truth with robust Umeyama implemented by pycolmap~\cite{colmap}, compute Euclidean error between predicted point and its corresponding ground truth, and calculate the area under the curve at varying error thresholds. 


\subsection{Within-domain and Cross-domain Evaluation}
\begin{figure}[t!]
    \centering
    \begin{minipage}{0.48\columnwidth}
        \centering
        \includegraphics[width=\linewidth, trim={0cm 0cm 0cm 0cm}, clip]{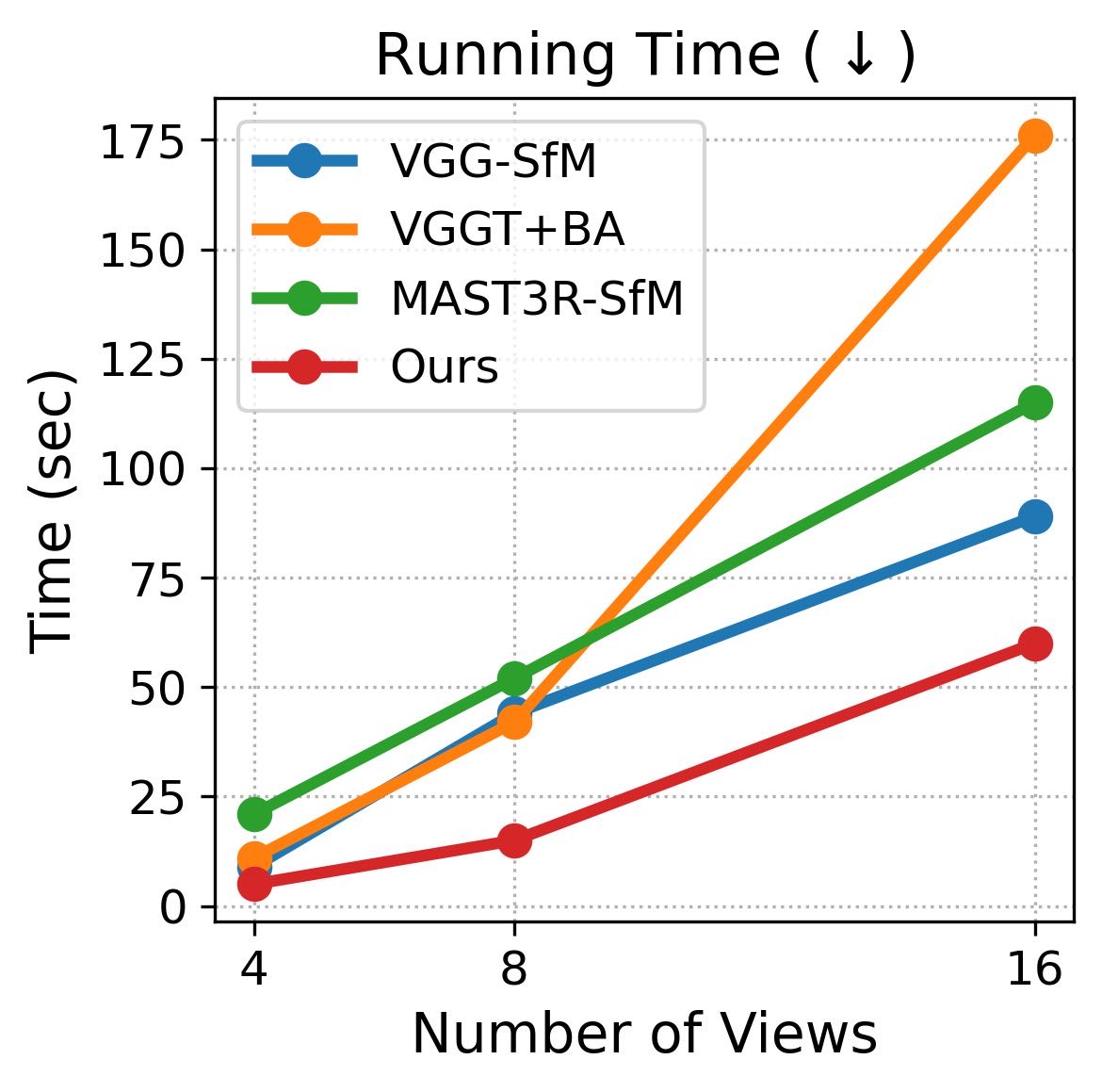}
    \end{minipage}
    \hfill
    \begin{minipage}{0.48\columnwidth}
        \centering
        \includegraphics[width=\linewidth, trim={0cm 0cm 0cm 0cm}, clip]{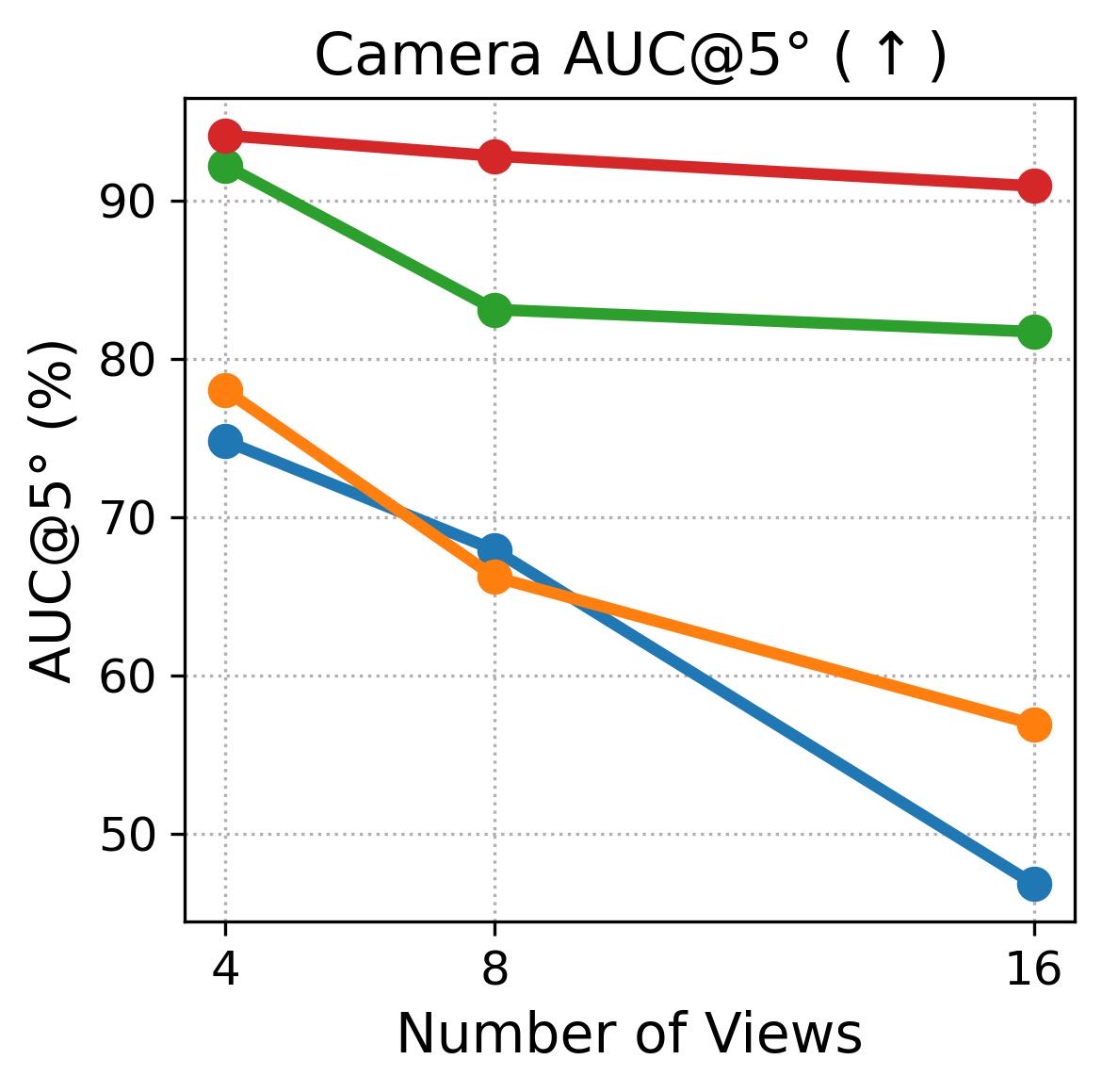}
    \end{minipage}

    \vspace{0.5em} 

    \begin{minipage}{0.48\columnwidth}
        \centering
        \includegraphics[width=\linewidth, trim={0cm 0cm 0cm 0cm}, clip]{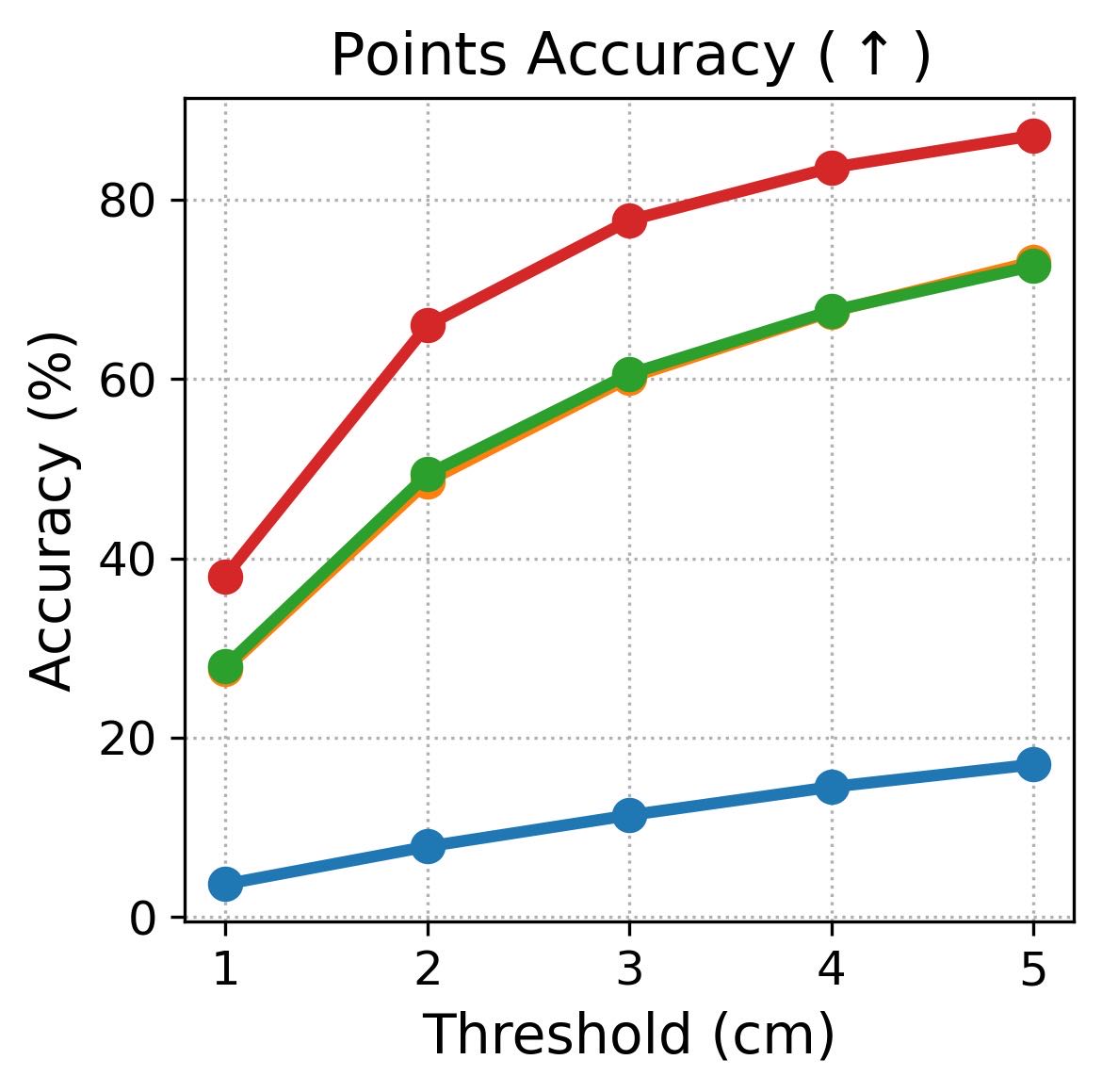}
    \end{minipage}
    \hfill
    \begin{minipage}{0.48\columnwidth}
        \centering
        \includegraphics[width=\linewidth, trim={0cm 0cm 0cm 0cm}, clip]{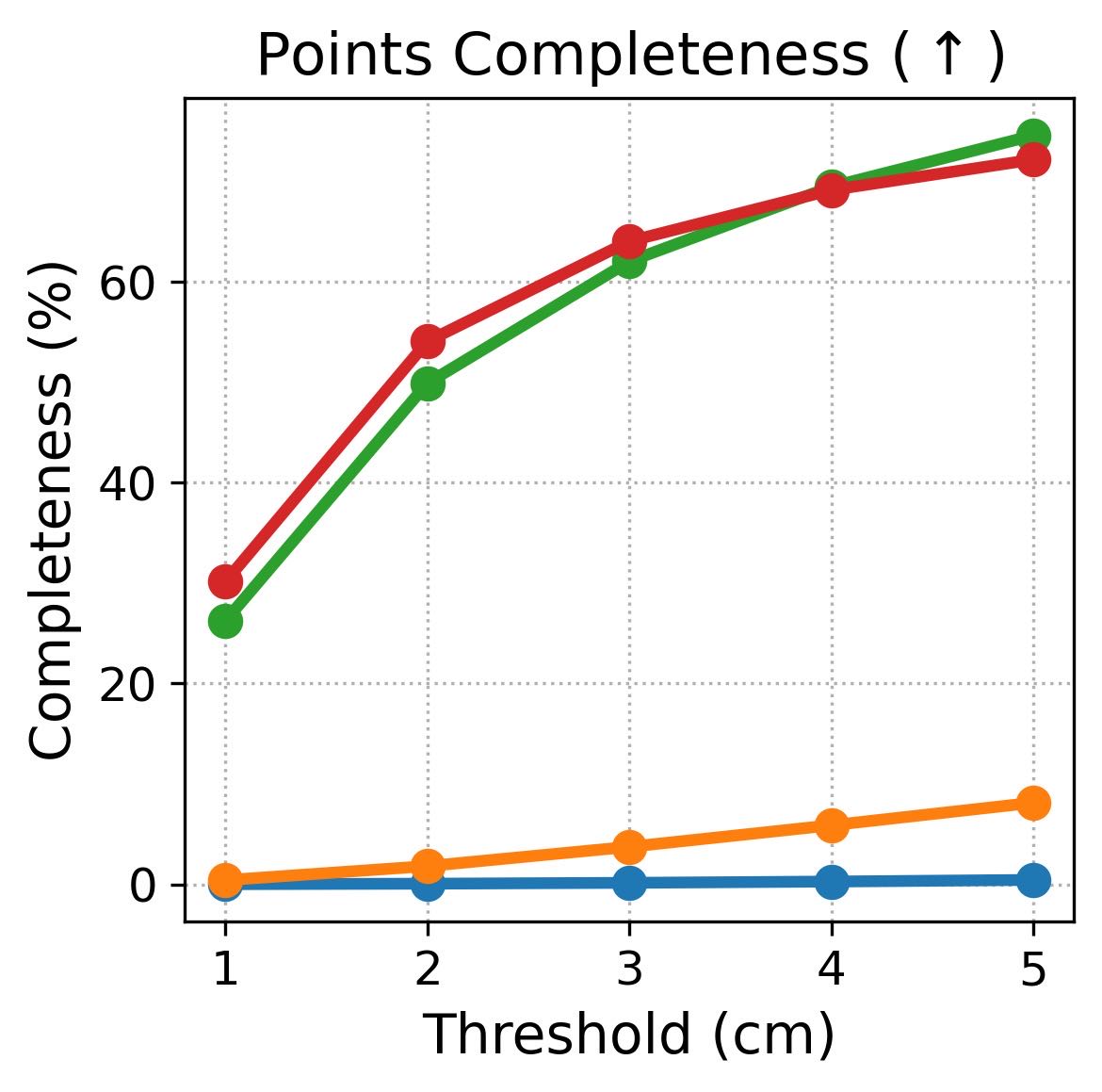}
    \end{minipage}

    \caption{\textbf{Comparison between SfMs on ETH3D~\cite{schoeps2017eth3d}.} Across 4/8/16-view setups, our SfM pipeline achieves consistently better camera pose accuracy, points accuracy, and good points completeness, while retaining the shortest running time.}
    \vspace{-5mm}
    \label{fig:sfm}
\end{figure}

\begin{figure*}[ht!]
    \centering
    \includegraphics[width=\linewidth]{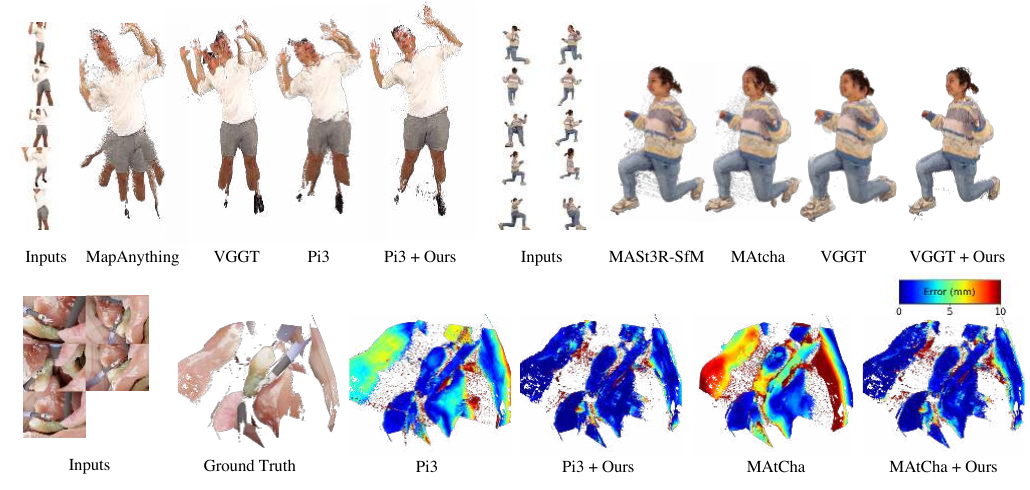}
    \caption{\textbf{Examples on out-of-domain \dsod{4D-DRESS}~\cite{wang20244ddress}} with visualisation (top row), and \dsod{MV-dVRK}~\cite{dvrk-smv} with both visualisation and reconstruction error (bottom row), predicted by feed-forward networks and enhanced by our method. Our GGPT can significantly improve the multi-view consistency in the reconstruction result.}
    \label{fig:vis_tab2}
\end{figure*}
\begin{table}[t!]
    \centering
    \small
\caption{\textbf{Multi-view 3D reconstruction on \dsod{out-of-domain} datasets.}  
AUC@1/5 cm (\% $\uparrow$) for human body reconstruction (4D-DRESS~\cite{wang20244ddress}) and AUC@1/5 mm (\% $\uparrow$) for surgical scene reconstruction (MV-dVRK~\cite{dvrk-smv}).}
  
    \label{tab:ood_benchmark}
    \vspace{-1em}

    \begin{tabular}{l c c}
    \toprule
    &\footnotesize\textbf{\dsod{4D-DRESS}~\cite{wang20244ddress}}  &  \footnotesize\textbf{\dsod{MV-dVRK}~\cite{dvrk-smv}}\\
%
\midrule
VGGT~\cite{wang2025vggt} & 10/45 & \ 8/33 \\
~\cite{wang2025vggt} + Ours & \textbf{66/77} &\textbf{45/61}   \\
\midrule
Pi3~\cite{wang2025pi3} & \ 8/50 & 18/51 \\
~\cite{wang2025pi3} + Ours & \textbf{63/80} &\textbf{40/67} \\
\midrule
MapAnything~\cite{keetha2025mapanything} & \ 2/12 &\ 3/13\\
~\cite{keetha2025mapanything} + Ours & \textbf{42/52} &\textbf{35/47} \\
\midrule
MAtCha~\cite{guedon2025matcha} & 48/68 &37/62  \\
~\cite{guedon2025matcha} + Ours & \textbf{62/75} &\textbf{50/67} \\
\midrule
MASt3R-SfM~\cite{duisterhof2025mastrsfm} &54/71 &39/62 \\
~\cite{duisterhof2025mastrsfm} + Ours &\textbf{64/75} &\textbf{49/66}  \\

    \bottomrule
    \end{tabular}



\end{table}

We evaluate methods on three benchmarks: ScanNet++~\cite{yeshwanth2023scannet++}, ETH3D~\cite{schoeps2017eth3d}, and T\&T~\cite{Knapitsch2017tandt}.mEach dataset contains varying baseline coverages with 4, 8, and 16 input views and each view split has 32 sequences. For ScanNet++ and ETH3D, we use a subset of test sets in~\cite{keetha2025mapanything}. For T\&T, we adopt the same view-sampling protocol~\cite{keetha2025mapanything} to construct a test set.
Since \method is exclusively trained on ScanNet++ using VGGT predictions, evaluations on unseen ScanNet++ scenes are referred to as \dsid{within-domain}, while results on ETH3D and T\&T are referred to as \dscd{cross-domain}. 

Despite GGPT is trained solely on VGGT's dense prediction, it can be applied to improve the dense prediction of other approaches, including feed-forward methods~\cite{wang2025pi3,keetha2025mapanything} and forward-optimization hybrid methods~\cite{duisterhof2025mastrsfm,guedon2025matcha}. Concretely, we replace VGGT's BA initialisation and $\mathbf{X}_d$ with the alternative prediction while using the same SfM pipeline and GGPT model weights.  

Tab.~\ref{tab:ma_benchmark_short} shows that our method can seamlessly enhance a range of advanced 3D reconstruction approaches, on both within-domain and cross-domain datasets, across different view setups. 
%
Pi3~\cite{wang2025pi3}, a concurrent work, is trained on ScanNet++ alongside a vast number of similar indoor scenes, achieving highly optimised performance on the ScanNet++ dataset.
%
Nevertheless, \method can still largely improve Pi3 on cross-domain datasets.
Fig.~\ref{fig:vis_tab1} shows that our refinement reduces geometric misalignments and mitigates multi-layer artifacts frequently observed in feed-forward predictions.
\subsection{Out-of-domain Evaluation}
\label{sec:ood_test}

To further test the generalisation of our method, we also evaluate it on \dsod{out-of-domain} datasets where the input images differ substantially in appearance or geometry from the training data of both our model and existing 3D reconstruction methods.
%
%
We consider two challenging out-of-distribution domains: human bodies and robotic abdominal surgery. We render high-resolution clothed body scans from 4D-DRESS~\cite{wang20244ddress}, and use photorealistic Blender renderings from the MV-dVRK~\cite{dvrk-smv} dataset. Each dataset contains 24 sequences, with 4–12 input views per sequence. 
%

Tab.~\ref{tab:ood_benchmark} shows our method achieves superior performance, particularly in AUC@1 cm/mm, which reflects fine-grained geometric accuracy. Fig.~\ref{fig:vis_tab2} illustrates that our approach effectively corrects distortions and misalignments produced by feed-forward models~\cite{wang2025pi3,wang2025vggt,keetha2025mapanything}, yielding more accurate and consistent reconstructions than hybrid methods~\cite{guedon2025matcha,duisterhof2025mastrsfm}. 


\subsection{Evaluation of the SfM Pipeline}
\label{sec:cmp_sfm_efficiency}
%
%


\noindent\textbf{Direct comparison.} Following standard SfM evaluation protocols~\cite{lindenberger2021pixsfm,wang2024vggsfm}, we measure the camera pose with AUC@5$\degree$ and the partial point map with Accuracy/Completeness based on its Chamfer distance to the ground truth. 
Fig.~\ref{fig:sfm} shows that our SfM outperforms state-of-the-art global SfM methods ~\cite{duisterhof2025mastrsfm,wang2025vggt,wang2024vggsfm} in camera pose and points accuracy.
Compared with MASt3R-SfM~\cite{duisterhof2025mastrsfm}, which optimises a dense grid for near-complete scene coverage, our method attains competitive completeness without sacrificing accuracy. Our pipeline consistently achieves the lowest runtime across all input view counts. 


\noindent\textbf{Impacts on geometry-conditioned models.} 
To further assess the utility of our SfM results, we feed $\mathbf{X}_s$, from either our method or ~\cite{duisterhof2025mastrsfm}, into state-of-the-art geometry-guided models~\cite{guo2025murre,zuo2025omni,pow3r_cvpr25,keetha2025mapanything}, and compare their dense predictions using each geometric condition. As shown in Tab.~\ref{tab:cmp_completion}, replacing the SfM input from~\cite{duisterhof2025mastrsfm} with ours yields substantial performance gains across all models. This verifies that our SfM pipeline provides a stronger geometric signal for the dense reconstruction task.
 

\begin{table}[ht!]
    \centering
    \small
    \caption{\textbf{Comparison of geometry conditions and geometry-grounded models.}  
    Points AUC@5/10 cm (\% $\uparrow$) on the ETH3D~\cite{schoeps2017eth3d} 8-view test set. \textbf{Two conclusions can be drawn here.} (1) Column-wise, our SfM $\mathbf{X}_s$ provides a stronger geometry signal than prior SfM~\cite{duisterhof2025mastrsfm} (We \textcolor{teal}{highlight} the gain from using our SfM). (2)  Row-wise, given the same SfM condition, our 3D GGPT outperforms prior 2D-based methods.}
    \begin{tabular}{l c c }
    \toprule
    Method & $\mathbf{X}_s$ from~\cite{duisterhof2025mastrsfm}& $\mathbf{X}_s$ from our SfM   \\
    \midrule
    Murre~\cite{guo2025murre} & 9/23 & 26/40 \textcolor{teal}{(+17/+17)} \\
    OMNI-DC~\cite{zuo2025omni} & 25/44 & 31/44 \textcolor{teal}{(+6/+0)} \\
    POW3R~\cite{pow3r_cvpr25} & 13/29 & 32/45 \textcolor{teal}{(+19/+16)} \\
    MapAnything~\cite{keetha2025mapanything} & 13/32 & 27/40 \textcolor{teal}{(+14/+8)} \\
    VGGT~\cite{wang2025vggt} + GGPT & \textbf{36/50} & \textbf{47/61} \textcolor{teal}{(+11/+11)} \\
    
    \bottomrule
    \end{tabular}
    \label{tab:cmp_completion}
    \vspace{-2mm}
\end{table}

\subsection{Geometry-conditioned Models}
 We compare our geometry-guided 3D point transformer with 2D depth completion methods~\cite{guo2025murre,zuo2025omni,keetha2025mapanything,pow3r_cvpr25}). All models receive the same partial points $\mathbf{X}_s$ produced either by our SfM or by prior state-of-the-art SfM~\cite{duisterhof2025mastrsfm}. For monocular methods, we process each view independently and unproject per-view depths into a global point map using the SfM camera poses. Tab.~\ref{tab:cmp_completion} shows that given the same partial geometry guidance, \method consistently produces more accurate dense point maps than all baselines, regardless of the underlying SfM algorithm. This verifies the effectiveness of the 3D point transformer architecture design. 

\section{Conclusion}
We introduced GGPT, a framework that enhances feed-forward 3D reconstruction by employing reliable geometric guidance from SfM. Our efficient SfM uses dense feature matching and linear triangulation to recover accurate sparse geometry. 
We further introduce a lightweight 3D point transformer to refine the feed-forward dense predictions directly in 3D space.
This design provides an effective way to fuse sparse geometric priors with dense feed-forward predictions.
Our GGPT demonstrates strong generalisation to various feed-forward methods and diverse datasets, highlighting its practicality as a broadly applicable refinement module for sparse-view 3D reconstruction. 

\newpage
\section{Acknowledgements}
This study was conducted within the national “Proficiency” research project (No. PFFS-21-19) funded by the Swiss Innovation Agency Innosuisse in 2021 as one of 15 flagship initiatives. This work was supported by the Swiss AI Initiative under project IDs a136 and a144, funded through a grant from the ETH Domain. The authors gratefully acknowledge the computational resources provided by the Swiss National Supercomputing Centre (CSCS) under the Alps infrastructure. We thank Philipp Lindenberger, Shaohui Liu, Zador Pataki, Xudong Jiang, Frano Rajic, Linfei Pan, Johannes Weidenfeller, and Malte Prinzler for valuable discussions and support.


{
    \small
    \bibliographystyle{ieeenat_fullname}
    \bibliography{main}
}

\clearpage
\begingroup

\twocolumn[
\begin{center}
    {\Large \bf \Large{\bf \method: Geometry Grounded Point Transformer} \\ -- Supplementary Material -- \par}
  \vspace*{30pt}
\end{center}
]

\appendix

\section{Results with RoMa v2 and Comparison with Additional Baselines}

In the main paper, we ensemble the dense correspondences predicted by RoMa~\cite{edstedt2024roma} and UFM~\cite{zhang2025ufm} (Sec.~\ref{sec:sfm_impl}) within the SfM pipeline. This configuration is used to produce the results reported in the main paper and the ablation studies in Sec.~\ref{sec:ablate}. Following the initial submission, we observed that replacing RoMa and UFM with the more recent dense matcher, \ie RoMa v2~\cite{edstedt2025romav2}, further improves performance across all evaluation datasets. The updated results are reported in Tab.~\ref{tab:ma_benchmark_short_RoMav2} and Tab.~\ref{tab:ood_benchmark_RoMav2}. In addition, we include two recent feed-forward models for comparison: \ie Pi3X~\cite{wang2025pi3}\footnote{Pi3X is an enhanced version of Pi3 with engineering updates. The model weights are publicly available at \url{https://github.com/yyfz/Pi3}.} and DepthAnything3~\cite{depthanything3}. As shown in Tab.~\ref{tab:ma_benchmark_short_RoMav2} and  Tab.~\ref{tab:ood_benchmark_RoMav2}, our method GGPT consistently outperforms these feed-forward baselines, regardless of whether it employs RoMa + UFM or RoMa v2, with particularly notable gains on cross-domain and out-of-domain datasets.

\begin{table}[t!]
    \centering
    \small
\caption{\textbf{Multi-view 3D reconstruction on \dsod{out-of-domain} datasets.}  
AUC@1/5 cm (\% $\uparrow$) for human body reconstruction (4D-DRESS~\cite{wang20244ddress}) and AUC@1/5 mm (\% $\uparrow$) for surgical scene reconstruction (MV-dVRK~\cite{dvrk-smv}). \textbf{\textcolor{ForestGreen}{Here we report additional results with latest dense matcher RoMa v2~\cite{edstedt2025romav2}, which further boosts the performance.}}}
    \label{tab:ood_benchmark_RoMav2}
    \vspace{-1em}

    \begin{tabular}{l c c}
    \toprule
    &\footnotesize\textbf{\dsod{4D-DRESS}}  &  \footnotesize\textbf{\dsod{MV-dVRK}}\\
\midrule
VGGT~\cite{wang2025vggt} & 10/45 & \ 8/33 \\
~\cite{wang2025vggt} + Ours \footnotesize{(RoMa + UFM)} & \textbf{66/77} &\textbf{45/61}   \\
~\cite{wang2025vggt} + Ours \footnotesize{(RoMa v2)} &  \textcolor{ForestGreen}{\textbf{75}/\textbf{87}} & \textcolor{ForestGreen}{\textbf{62}/\textbf{77}} \\
\midrule
Pi3~\cite{wang2025pi3} & \ 8/50 & 18/51 \\
~\cite{wang2025pi3} + Ours \footnotesize{(RoMa + UFM)} & \textbf{63/80} &\textbf{40/67} \\
~\cite{wang2025pi3} + Ours \footnotesize{(RoMa v2)}  & \textcolor{ForestGreen}{\textbf{76}/\textbf{88}} & \textcolor{ForestGreen}{\textbf{67}/\textbf{82}} \\
\midrule
Pi3X~\cite{wang2025pi3} &3/12 & 4/22 \\
~\cite{wang2025pi3} + Ours \footnotesize{(RoMa + UFM)}  & \textbf{49/66} & \textbf{42/57} \\
~\cite{wang2025pi3} + Ours \footnotesize{(RoMa v2)}  & \textcolor{ForestGreen}{\textbf{69}/\textbf{83}} & \textcolor{ForestGreen}{\textbf{60}/\textbf{76}} \\
\midrule

MapAnything~\cite{keetha2025mapanything} & \ 2/12 &\ 3/13\\
~\cite{keetha2025mapanything} + Ours \footnotesize{(RoMa + UFM)}  & \textbf{42/52} &\textbf{35/47} \\
~\cite{keetha2025mapanything} + Ours \footnotesize{(RoMa v2)} & \textcolor{ForestGreen}{\textbf{54/75}} &\textcolor{ForestGreen}{\textbf{33/52}}   \\
\midrule
DepthAnything3~\cite{depthanything3}  & 2/5 & 19/52 \\
~\cite{depthanything3} + Ours \footnotesize{(RoMa + UFM)} & \textbf{29/47} &\textbf{47/68} \\
~\cite{depthanything3} + Ours \footnotesize{(RoMa v2)}  & \textcolor{ForestGreen}{\textbf{49}/\textbf{70}} & \textcolor{ForestGreen}{\textbf{67}/\textbf{80}} \\
\midrule
MAtCha~\cite{guedon2025matcha} & 48/68 &37/62  \\
~\cite{guedon2025matcha} + Ours \footnotesize{(RoMa + UFM)} & \textbf{62/75} &\textbf{50/67} \\
~\cite{guedon2025matcha} + Ours \footnotesize{(RoMa v2)} & 
\textcolor{ForestGreen}{\textbf{67}/\textbf{80}} & \textcolor{ForestGreen}{\textbf{64}/\textbf{80}} \\ 
\midrule
MASt3R-SfM~\cite{duisterhof2025mastrsfm} &54/71 &39/62 \\
~\cite{duisterhof2025mastrsfm} + Ours \footnotesize{(RoMa + UFM)} &\textbf{64/75} &\textbf{49/66}  \\
~\cite{duisterhof2025mastrsfm} + Ours \footnotesize{(RoMa v2)} &  
 \textcolor{ForestGreen}{\textbf{70}/\textbf{82}} & \textcolor{ForestGreen}{\textbf{63}/\textbf{79}} \\

    \bottomrule
    \end{tabular}



\end{table}

\begin{table*}[h!]
    \centering
    \setlength{\tabcolsep}{4pt} 
    \caption{\textbf{Multi-view 3D Reconstruction on Standard Test Sets.}
    We report AUC@5/10 cm (\% $\uparrow$). Our method (our SfM and GGPT) can improve dense predictions of various models. Despite GGPT is trained solely on ScanNet++ and VGGT's dense prediction, as we highlight the top-left cells as \dsid{within-domain}, it generalises well to \dscd{cross-domain datasets} and other methods' predictions (rows 3–10).  For the results reported in the main paper, we ensemble the outputs of RoMa~\cite{edstedt2024roma} and UFM~\cite{zhang2025ufm} for dense matchings. \textbf{\textcolor{ForestGreen}{Here we report additional results with latest dense matcher RoMa v2~\cite{edstedt2025romav2}, which further boosts the performance.}}}
    \centering
    \label{tab:ma_benchmark_short_RoMav2}
\begin{tabular}{l ccc c ccc c ccc}
\toprule
& \multicolumn{3}{c}{\dsid{ScanNet++}~\cite{yeshwanth2023scannet++}} 
&& \multicolumn{3}{c}{\dscd{ETH3D}~\cite{schoeps2017eth3d}}
&& \multicolumn{3}{c}{\dscd{T\&T}~\cite{Knapitsch2017tandt}} \\
& 4 imgs & 8 imgs & 16 imgs && 4 imgs & 8 imgs & 16 imgs && 4 imgs & 8 imgs & 16 imgs \\
\midrule
VGGT~\cite{wang2025vggt} 
& \cellcolor{trainhighlight}23/37 & \cellcolor{trainhighlight}19/32 & \cellcolor{trainhighlight}16/29 
&& 27/41 & 23/36 & 19/32 
&& 26/40 & 25/39 & 24/38 \\
~\cite{wang2025vggt} + Ours \footnotesize{(RoMa + UFM)}
& \cellcolor{trainhighlight}\textbf{38/53} & \cellcolor{trainhighlight}\textbf{45/60} & \cellcolor{trainhighlight}\textbf{50/66}
&& \textbf{41/55} & \textbf{47/61} & \textbf{49/63}
&& \textbf{34/47} & \textbf{42/57} & \textbf{43/57} \\

~\cite{wang2025vggt} + Ours \footnotesize{(RoMa v2)}
 & \textcolor{ForestGreen}{\textbf{47}/\textbf{62}} & \textcolor{ForestGreen}{\textbf{56}/\textbf{70}} & \textcolor{ForestGreen}{\textbf{62}/\textbf{76}} && \textcolor{ForestGreen}{\textbf{54}/\textbf{66}} & \textcolor{ForestGreen}{\textbf{57}/\textbf{69}} & \textcolor{ForestGreen}{\textbf{62}/\textbf{75}} && \textcolor{ForestGreen}{\textbf{39}/\textbf{52}} & \textcolor{ForestGreen}{\textbf{50}/\textbf{63}} & \textcolor{ForestGreen}{\textbf{49}/\textbf{62}} \\
\midrule

Pi3~\cite{wang2025pi3} 
& \textbf{54/69} & \textbf{56}/71 & 58/\textbf{74} 
&& 31/47 & 25/41 & 23/38 
&& 25/39 & 26/42 & 25/40 \\
~\cite{wang2025pi3} + Ours \footnotesize{(RoMa + UFM)}
& \textbf{54}/68 & \textbf{56/72} & \textbf{59/74} 
&& \textbf{36/53} & \textbf{36/53} & \textbf{37/54} 
&& \textbf{27/43} & \textbf{32/50} & \textbf{33/50}  \\
~\cite{wang2025pi3} + Ours \footnotesize{(RoMa v2)}  & \textcolor{ForestGreen}{\textbf{58}/\textbf{72}} & \textcolor{ForestGreen}{\textbf{61}/\textbf{75}} & \textcolor{ForestGreen}{\textbf{65}/\textbf{78}} && \textcolor{ForestGreen}{\textbf{44}/\textbf{59}} & \textcolor{ForestGreen}{\textbf{45}/\textbf{60}} & \textcolor{ForestGreen}{\textbf{46}/\textbf{63}} && \textcolor{ForestGreen}{\textbf{27}/\textbf{42}} & \textcolor{ForestGreen}{\textbf{35}/\textbf{53}} & \textcolor{ForestGreen}{\textbf{36}/\textbf{53}} \\

\midrule

Pi3X~\cite{wang2025pi3} & \textbf{51}/66 &\textbf{ 54}/70 & \textbf{55/72} && 34/49 & 29/45 & 26/41 && 25/40 & 24/40 & 23/39 \\
~\cite{wang2025pi3} + Ours \footnotesize{(RoMa + UFM)}& \textbf{51/67} & \textbf{54/71} & \textbf{55/72} && \textbf{38/54} & \textbf{42/57} & \textbf{49/64} && \textbf{31/44} & \textbf{34/51 }& \textbf{41/55 }\\
~\cite{wang2025pi3} + Ours \footnotesize{(RoMa v2)} &  \textcolor{ForestGreen}{\textbf{56}/\textbf{70}} & \textcolor{ForestGreen}{\textbf{63}/\textbf{76}} & \textcolor{ForestGreen}{\textbf{68}/\textbf{80}} && \textcolor{ForestGreen}{\textbf{60}/\textbf{71}} & \textcolor{ForestGreen}{\textbf{63}/\textbf{74}} & \textcolor{ForestGreen}{\textbf{69}/\textbf{79}} && \textcolor{ForestGreen}{\textbf{39}/\textbf{51}} & \textcolor{ForestGreen}{\textbf{50}/\textbf{63}} & \textcolor{ForestGreen}{\textbf{50}/\textbf{63}} \\

\midrule
MapAnything~\cite{keetha2025mapanything} 
& 40/57 & 38/57 & 38/58 
&& 10/20 & 7/15 & 5/12 
&& 10/21 & 9/20 & 8/17   \\
~\cite{keetha2025mapanything} + Ours \footnotesize{(RoMa + UFM)}
& \textbf{44/61} & \textbf{48/64} & \textbf{52/68}
&& \textbf{32/43} & \textbf{33/45} & \textbf{34/47}
&& \textbf{29/43} & \textbf{40/55} & \textbf{42/56}   \\
~\cite{keetha2025mapanything} + Ours \footnotesize{(RoMa v2)} &
\textcolor{ForestGreen}{\textbf{56}/\textbf{70}} & \textcolor{ForestGreen}{\textbf{62}/\textbf{75}} & \textcolor{ForestGreen}{\textbf{67}/\textbf{79}} && \textcolor{ForestGreen}{\textbf{45}/\textbf{57}} & \textcolor{ForestGreen}{\textbf{48}/\textbf{61}} & \textcolor{ForestGreen}{\textbf{51}/\textbf{64}} && \textcolor{ForestGreen}{\textbf{35}/\textbf{49}} & \textcolor{ForestGreen}{\textbf{46}/\textbf{61}} & \textcolor{ForestGreen}{\textbf{48}/\textbf{61}} 
\\

\midrule
DepthAnything3~\cite{depthanything3}
 & 10/20 & 7/16 & 5/13 && 28/42 & 25/39 & 23/37 && 10/21 & 8/19 & 8/18 \\
~\cite{depthanything3} + Ours \footnotesize{(RoMa + UFM)}& \textbf{36/50 }& \textbf{45/60} & \textbf{53/68} && \textbf{35/49 }& \textbf{41/57} & \textbf{45/60} && \textbf{30/43 }& \textbf{31/48 }& \textbf{38/53} \\
~\cite{depthanything3} + Ours \footnotesize{(RoMa v2)} 
 & \textcolor{ForestGreen}{\textbf{47}/\textbf{59}} & \textcolor{ForestGreen}{\textbf{57}/\textbf{70}} & \textcolor{ForestGreen}{\textbf{65}/\textbf{77}} && \textcolor{ForestGreen}{\textbf{53}/\textbf{64}} & \textcolor{ForestGreen}{\textbf{58}/\textbf{70}} & \textcolor{ForestGreen}{\textbf{61}/\textbf{72}} && \textcolor{ForestGreen}{\textbf{38}/\textbf{51}} & \textcolor{ForestGreen}{\textbf{46}/\textbf{61}} & \textcolor{ForestGreen}{\textbf{48}/\textbf{62}} \\

\midrule

MAtCha~\cite{guedon2025matcha} 
& 12/19 & 15/26 & 18/32 
&& 40/52 & 41/53 & 42/56 
&& 34/47 & 36/50 & 33/47  \\
~\cite{guedon2025matcha} + Ours \footnotesize{(RoMa + UFM)}
& \textbf{27/37} & \textbf{40/52} & \textbf{48/63}
&& \textbf{42/55} & \textbf{47/60} & \textbf{50/65}
&& \textbf{35/48} & \textbf{43/57} & \textbf{43/57}  \\
~\cite{guedon2025matcha} + Ours \footnotesize{(RoMa v2)} & \textcolor{ForestGreen}{\textbf{38}/\textbf{48}} & \textcolor{ForestGreen}{\textbf{47}/\textbf{60}} & \textcolor{ForestGreen}{\textbf{60}/\textbf{72}} && \textcolor{ForestGreen}{\textbf{53}/\textbf{65}} & \textcolor{ForestGreen}{\textbf{56}/\textbf{67}} & \textcolor{ForestGreen}{\textbf{62}/\textbf{74}} && \textcolor{ForestGreen}{\textbf{39}/\textbf{52}} & \textcolor{ForestGreen}{\textbf{50}/\textbf{63}} & \textcolor{ForestGreen}{\textbf{49}/\textbf{61}} \\
\midrule

MASt3R-SfM~\cite{duisterhof2025mastrsfm}
& 12/22 & 14/35 & 16/31
&& 37/50 & 39/51 & 40/54
&& \textbf{34}/46 & 37/50 & 36/49  \\
~\cite{duisterhof2025mastrsfm} + Ours \footnotesize{(RoMa + UFM)}
& \textbf{30/41} & \textbf{39/52} & \textbf{48/64}
&& \textbf{41/55} & \textbf{46/59} & \textbf{48/63}
&& 33/\textbf{47} & \textbf{41/56} & \textbf{42/56}  \\
~\cite{duisterhof2025mastrsfm} + Ours \footnotesize{(RoMa v2)}  & \textcolor{ForestGreen}{\textbf{40}/\textbf{50}} & \textcolor{ForestGreen}{\textbf{51}/\textbf{64}} & \textcolor{ForestGreen}{\textbf{60}/\textbf{73}} && \textcolor{ForestGreen}{\textbf{52}/\textbf{64}} & \textcolor{ForestGreen}{\textbf{55}/\textbf{66}} & \textcolor{ForestGreen}{\textbf{62}/\textbf{73}} && \textcolor{ForestGreen}{\textbf{39}/\textbf{52}} & \textcolor{ForestGreen}{\textbf{49}/\textbf{62}} & \textcolor{ForestGreen}{\textbf{48}/\textbf{61}} \\
\bottomrule
\end{tabular}

\end{table*}

\section{Ablation Studies}
\label{sec:ablate}
\subsection{Structure-from-Motion}
\label{sec:sfm_ablate}
We conduct experiments to validate the contribution of our three SfM design components. (1) \textit{Dense matchers}: Using dense matchers rather than sparse matchers. (2) \textit{Sparse BA}: Selecting a small subset of matches for Bundle Adjustment (BA) instead of a large set. (3) \textit{DLT triangulation}: Using Direct Linear Triangulation (DLT) rather than conventional RANSAC-based non-linear triangulation.

The ablation is conducted on the ETH3D test set~\cite{schoeps2017eth3d} with varying baseline coverages with 4, 8, and 16 input views. In Tab.~\ref{tab:ablate_sfm}, we report the running time of each step (\textit{Time}), and the accuracy of the camera poses (\textit{Cam.}) and sparse points estimation (\textit{Pts.}). We adhere to standard Structure-from-Motion (SfM) evaluation protocols~\cite{lindenberger2021pixsfm,wang2024vggsfm,pataki2025mpsfm}. For camera poses, we compare the relative poses to the ground truth, compute the maximum of rotational and translational angular errors, and report the Area Under the recall Curve (AUC) up to $1^\circ$, $5^\circ$, and $10^\circ$. For sparse points, we compute the chamfer distance between the triangulated points and the ground truth, report the accuracy and completeness of the reconstruction as the ratio of points within a given distance of each other. 

\begin{table*}[h!]
    \centering
    \small
\caption{\textbf{Ablation of SfM Components on ETH3D.} We evaluate design variants across three stages of our SfM—matching, bundle adjustment, and triangulation—and highlight the configuration with the best efficiency or accuracy. For each variant, we report: (1) \textbf{Time} denotes the running time (s $\downarrow$) of each stage. (2) \textbf{Cam.} denotes the AUC@1/5$\degree$ (\% $\uparrow$) for camera-pose estimation. (3) \textbf{Pts.} (\% $\uparrow$) denotes the Accuracy/Completeness@1 cm (\% $\uparrow$) of reconstructed points.}

    \label{tab:ablate_sfm} 
    \vspace{-0.5em}
\setlength{\tabcolsep}{0.5pt} 
    \begin{tabular*}{\linewidth}{@{\extracolsep{\fill}}l ccc c ccc c ccc @{}}
    \toprule
    & \multicolumn{3}{c}{4 Views} && \multicolumn{3}{c}{8 Views} && \multicolumn{3}{c}{16 Views} \\
    \cmidrule(lr){2-4} \cmidrule(lr){6-8} \cmidrule(lr){10-12}
    & Time (s)  & Cam. (\%)  & Pts. (\%)
    && Time (s) & Cam. (\%) & Pts. (\%)
    && Time (s) & Cam. (\%) & Pts. (\%) \\
    \midrule
        (a) \textbf{Dense Matcher}   \\
        MASt3R~\cite{mast3r_eccv24} (Sparse) & 8.4 & 75/86 & 19/ \ 2 && 21 & 75/85 & 21/ \ 4 && 57 & 76/86 & 26/ \ 8 \\
        RoMa~\cite{edstedt2024roma} (Dense) & 2.3 & \textbf{90/93} & \textbf{47}/20 && 8.8 & \textbf{87}/91 & \textbf{46}/25 && 36 & \textbf{89/93} & \textbf{46}/29 \\
        UFM~\cite{zhang2025ufm} (Dense) & \textbf{1.0} & 64/84 & 26/19 && \textbf{1.7} & 81/90 & 30/25 && \textbf{8.5} & 77/87 & 33/30 \\
        RoMa~\cite{edstedt2024roma} + UFM~\cite{zhang2025ufm} (Ours, Dense) & 3.3 & 86/\textbf{93} & 35/\textbf{22} && 11 & \textbf{87/92} & 37/\textbf{30} && 44 & 85/91 & 40/\textbf{35}  \\
        \hdashline
        (b) \textbf{Sparse Bundle Adjustment (BA)}   \\
        $n_{\textrm{BA}}=512$ & \textbf{0.2} & 85/\textbf{93} & \textbf{36/23} &&\textbf{ 0.5} & 85/90 & \textbf{38/31} && \textbf{1.0} & 84/90 & 40/35 \\        
        $n_{\textrm{BA}}=2048$ & 1.0 & 86/\textbf{93} & 35/22 && 1.9 & 87/\textbf{92} & 37/30 && 15 & \textbf{85/91} & 40/35  \\
        $n_{\textrm{BA}}=4096$ & 1.9 & \textbf{87/93} & \textbf{36/23} && 17 & \textbf{88/92} & \textbf{38/31} && 29 & \textbf{85/91} &\textbf{ 41/36} \\
        \hdashline
        (c) \textbf{Direct Linear Triangulation (DLT)}   \\
        Ransac-based non-linear triangulation ~\cite{revisitedsfm} & 14 & \textbf{86/93} & 34/\textbf{23}  && 64 &   \textbf{87/92} & 36/\textbf{31} && 272 & \textbf{85/91} & 39/\textbf{35} \\
        Direct linear triangulation & \textbf{0.03} & \textbf{86/93} & \textbf{35}/22 && \textbf{0.08} &  \textbf{87/92} & \textbf{37}/30 &&\textbf{ 0.3} & \textbf{85/91}  & \textbf{40/35} \\

        \bottomrule
    \end{tabular*}
\end{table*}

\noindent\textbf{Dense matchers.} We compare dense matching regressors~\cite{edstedt2024roma,zhang2025ufm} with the SOTA sparse keypoint-based MASt3R~\cite{mast3r_eccv24}, which uses a reciprocal nearest-neighbor (NN) searching algorithm to extract a sparse grid of correspondences between each pair. 
Tab.~\ref{tab:ablate_sfm} shows that the sparse MASt3R matcher is noticeably slower and produces less reliable correspondences than dense matching regressors, which leads to suboptimal camera and points accuracy.  due to the sparsity of the matchings, it can only produce a extremely sparse set of points, yielding a low completeness, \eg 4 \% vs 20 \% on the 8-view setup. 
For SOTA dense matchers, we find that RoMa~\cite{edstedt2024roma} offers better accuracy while UFM~\cite{zhang2025ufm} offers better completeness and efficiency. Ensembling the two dense matchers can achieve the balance between points accuracy and completeness with minimal computational overhead, as shown in the \textit{RoMa + UFM (Ours, Dense)} row of Tab.~\ref{tab:ablate_sfm}. In Sec.~\ref{sec:vary_sfm_density}, we will provide further experiments showing that this ensemble strategy also yields the best final dense reconstruction.

\noindent\textbf{Sparse BA.} We study the effects of $n_{\textrm{BA}}$, \ie the number of matches per image used for bundle adjustment (BA). Results show that reducing $n_{\textrm{BA}}$ from 4096 to 512 can considerably reduce the running time, particularly for $\ge 8$ input views, but achieves quite similar camera and points accuracy. Our default configuration uses $n_{\textrm{BA}}=2048$, though this ablation shows that $n_{\textrm{BA}}=512$ actually suffices for the ETH3D dataset.

\noindent\textbf{DLT triangulation.} For the final point triangulation stage, we emphasise the efficiency of direct linear triangulation (DLT)~\cite{hartley2003multiple}. We compare it with the standard COLMAP pipeline~\cite{revisitedsfm}, which conducts LO-RANSAC~\cite{loransac} for outlier rejection and a non-linear iterative optimization for refinement before and after linear triangulation respectively. For this method, we employ the implementation function \footnote{\url{https://colmap.github.io/pycolmap/pycolmap.html\#pycolmap.triangulate\_points}} provided by pycolmap~\cite{colmap}, using the same matches and camera poses as in our DLT implementation. The comparison shows that DLT is orders of magnitude faster than the RANSAC-based non-linear pipeline, as DLT can be executed directly as CUDA tensor operations (See Sec.\ref{sec:sfm_impl}). Remarkably, even without LO-RANSAC and without a subsequent non-linear refinement, DLT attains comparable and in many cases superior accuracy and completeness. This robustness stems from our prefitering of correspondences using cycle consistency and matching confidence.

\begin{table}[h!]
    \centering
    \small
    \caption{\textbf{Ablation Studies on GGPT.} We report AUC@5/10 cm on within-domain \dsid{ScanNet++} and cross-domain \dscd{ETH3D} 8-view splits and AUC@1/5 cm on out-of-domain \dsod{4D-DRESS}.}
    \label{tab:ablation_ggpt}
    \vspace{-0.5em}
\setlength{\tabcolsep}{1pt} 
    \begin{tabular*}{\linewidth}{@{\extracolsep{\fill}}lccc@{}}
        \toprule
        &  \makebox[1cm]{\dsid{SNt++}} 
&  \makebox[1cm]{\dscd{ETH3D}} 
&  \makebox[1cm]{\dsod{4DDS.}} \\
        \midrule\
        VGGT~\cite{wang2025vggt} & 19/32 & 23/36 & 10/45 \\
        \hdashline
        \textbf{(a) Refinement Architecture} \\
        \underline{2D Transformers} & & & \\
        \hspace{1em}$\mathbf{X}_s \rightarrow $ VGGT~\cite{wang2025vggt,pow3r_cvpr25} & \textbf{55/70} & 25/40 & 12/52 \\
        \hspace{1em}$\mathbf{X}_s \rightarrow $ MapAnything~\cite{keetha2025mapanything} & 39/57 & 11/21 & 5/34 \\
        \underline{3D Networks} & & & \\
        \hspace{1em}Minkowski~\cite{minkowski} & 39/53 & 47/60 & 63/73 \\
        \hspace{1em}PTv3~\cite{wu2024ptv3} (Ours) & 45/60 & \textbf{47/61} & \textbf{66/77} \\
        \midrule
        \textbf{(b) Input Encodings} \\
        w/o $\mathbf{X}_s$ & 21/35 & 20/34 & 13/42 \\
        w/o $\mathbf{X}_{d\rightarrow s}$ & 24/41 & 24/41 & 14/47 \\
        w/o $\mathbf{z}_s$ & 42/57 & 44/58 & 56/72 \\
        Full model & \textbf{45/60} & \textbf{47/61} & \textbf{66/77} \\
        \midrule
        \textbf{(c) Patch Size} \\
        r=0.05  & 30/44 & 37/50 & 30/56  \\
        r=0.1  & 39/53 & 43/57 &  60/72 \\
        r=0.2 & \textbf{45/60} & \textbf{47}/\textbf{61 }& \textbf{66/77} \\
        r=0.6 & 40/55 & 38/53 & 64/74 \\
        r=1.0 & 41/56 & 40/56 & 63/75 \\
        \bottomrule
    \end{tabular*}
    \vspace{-0.4cm}
\end{table}

\subsection{Geometry-Grounded Point Transformer}
In this section, we perform ablations on our Geometry-Grounded Point Transformer (\cref{tab:ablation_ggpt}).

\noindent\textbf{Refinement architecture.} We first verify the advantage of performing geometry-conditioned refinement in 3D space with a 3D point transformer rather than a 2D image transformer. We use two 2D baselines built on pre-trained VGGT~\cite{wang2025vggt} and MapAnything~\cite{keetha2025mapanything}. For VGGT, we follow POW3R~\cite{pow3r_cvpr25} to convert it into a geometry-conditioned model: $\mathbf{X}_s$ is patchified, encoded with an MLP, and injected into the first multi-view attention layer. MapAnything already supports encoding additional geometry signals. We fine-tune both models on the same ScanNet++ training set. Despite these 2D variants achieve the strongest performance on the \dsid{within-domain ScanNet++} test set, their accuracy degrades substantially on the \dscd{cross-domain ETH3D} and \dsod{out-of-domain 4D-DRESS}. In contrast, our 3D point transformer effectively generalises across diverse domains. Finally, for different 3D backbones, we find that Point Transformer v3 (PTv3)~\cite{wu2024ptv3} outperforms Minkowski Engine~\cite{minkowski}.

\noindent\textbf{Input encodings.} We evaluate the impact of our input encoding strategies using the partial point cloud $\mathbf{X}_s$ from our SfM pipeline and the dense prediction $\mathbf{X}_d$.
(1) Without using $\mathbf{X}_s$ as guidance, the network still learns to unconditionally denoise $\mathbf{X}_d$ on ScanNet++, but fails to generalise to other datasets. (2) Introducing $\mathbf{x}_{d\rightarrow s}$ (Eq.~8 of the main paper) to make the network aware of the correspondence between the guidance and initial prediction proves crucial. (3) Encoding $\mathbf{X}_s$ as $\mathbf{z}_s$ and providing it jointly with $\mathbf{z}_d$ stabilises training and improves overall performance. \cref{tab:ablation_ggpt} demonstrates that all introduced encodings are essential for achieving high-quality results.

\noindent\textbf{Patch-based processing.} We study the impact of GGPT's patch size by varying the ratio $r$ between the radii of the input patch and the scene.
For larger patch sizes $r\ge 0.6$, GPU memory becomes a bottleneck, and we must reduce the number of input points from 400k to 100k. In this case, we randomly subsample 100k points from each cropped region as the input chunk.
We find that a slightly smaller patch size $r=0.2$ shows optimal performance across different datasets, as shown in \cref{tab:ablation_ggpt}. 
This behaviour may stem from two factors: (1) smaller patches concentrate the model’s capacity on capturing fine-grained geometric structure, and (2) partitioning the scene into numerous small patches effectively serves as a form of data augmentation, increasing sample diversity and enhancing generalisation.

\begin{table*}[t] 
    \centering
    \small
    \caption{\textbf{Performance of GGPT across Varying SfM Input Densities.}
    We vary the SfM configuration, including the matcher type, DLT threshold $\epsilon_{\mathrm{DLT}}$, and reprojection threshold $\epsilon_{\mathrm{reproj}}$, to obtain SfM point clouds of different coverage (\%). We report AUC@5/10 cm on within-domain \dsid{ScanNet++} and cross-domain \dscd{ETH3D} 8-view splits and AUC@1/5 cm (\% $\uparrow$) on out-of-domain \dsod{4D-DRESS}. 
    Our GGPT is robust to SfM points of various densities, and can improve the complete reconstruction even when the SfM point is extremely sparse, \eg 8.5\% on ScanNet++.}
    \label{tab:ablate_sfmfilter}
    \vspace{0.5em}
    \setlength{\tabcolsep}{4pt} 
    \begin{tabular*}{\linewidth}{@{\extracolsep{\fill}} c c c l c c c c c c @{}}
        \toprule
        \multirow{2}{*}{\textbf{Dense Matcher}} &
        \multirow{2}{*}{$\epsilon_{\mathrm{DLT}}$} &
        \multirow{2}{*}{$\epsilon_{\mathrm{reproj}}$} &
        \multirow{2}{*}{\textbf{Point Cloud}} &
        \multicolumn{2}{c}{\dsid{SNt++}} &
        \multicolumn{2}{c}{\dscd{ETH3D}} &
        \multicolumn{2}{c}{\dsod{4DDS}} \\
        \cmidrule(lr){5-6} \cmidrule(lr){7-8} \cmidrule(lr){9-10}
         &  &  &  & 
        \textbf{Coverage (\%)} & \textbf{AUC (\%)} &
        \textbf{Coverage (\%)} & \textbf{AUC (\%)} &
        \textbf{Coverage (\%)} & \textbf{AUC (\%)} \\
        \midrule
        -- & -- & -- & VGGT~\cite{wang2025vggt}($\mathbf{X}_d$) & 100 & 19/32 & 100 & 23/36 & 100 & 10/45 \\
        \midrule
        RoMa & 0.6 & 2 & Our SfM ($\mathbf{X}_s$) & 8.5 & 6\ /\ 7 & 29 & 20/23 & 60 & 55/56 \\
        RoMa & 0.6 & 2 & + GGPT ($\hat{\mathbf{X}}_d$) & 100 & 26/41 & 100 & 39/53 & 100 & 59/67 \\
        \hdashline
        RoMa & 0.1 & 4 & Our SfM ($\mathbf{X}_s$) & 21 & 12/15 & 54 & 35/41 & 73 & 62/64 \\
        RoMa & 0.1 & 4 & + GGPT ($\hat{\mathbf{X}}_d$) & 100 & 31/45 & 100 & 46/59 & 100 & \textbf{66/77} \\
        \midrule
        RoMa+UFM & 0.6 & 2 & Our SfM ($\mathbf{X}_s$) & 60 & 33/41 & 63 & 35/44 & 70 & 52/58 \\
        RoMa+UFM & 0.6 & 2 & + GGPT ($\hat{\mathbf{X}}_d$) & 100 & 43/58 & 100 & 43/57 & 100 & 56/67 \\
        \hdashline
        RoMa+UFM & 0.1 & 4 & Our SfM ($\mathbf{X}_s$) & 66 & 36/45 & 76 & 43/54 & 84 & 61/69 \\
        RoMa+UFM & 0.1 & 4 & + GGPT ($\hat{\mathbf{X}}_d$) & 100 & \textbf{45/60} & 100 & \textbf{47/61} & 100 & 63/73 \\
        \bottomrule
    \end{tabular*}
    \vspace{-0.5em}
\end{table*}

\subsection{Performance Across Varying SfM Input Densities}
\label{sec:vary_sfm_density}

As discussed in Section~3.1, adjusting the number of matches and the filtering threshold allows us to control the density and accuracy of the SfM estimate $\mathbf{X}_{s}$, which serves as geometric guidance for GGPT. In this section, we examine how GGPT’s final dense predictions vary with different input densities of $\mathbf{X}_{s}$. The results show that GGPT remains robust and delivers strong performance across a broad spectrum of SfM input densities.

\noindent\textbf{Experimental setup.} We control the density and accuracy of $\textbf{X}_{s}$ using three factors. (1) Dense Matchers: As shown in Tab.~\ref{tab:ablate_sfm}, using only RoMa~\cite{edstedt2024roma} for matching leads to more accurate yet sparser points than using it together with UFM~\cite{zhang2025ufm}. (2) DLT Confidence Threshold ($\epsilon_{\mathrm{DLT}})$: the confidence threshold to pre-filter matches for DLT. A higher value yields more accurate yet sparser $\mathbf{X}_s$. (3) $\epsilon_{\mathrm{reproj}}$: the 2D reprojection error threshold to post-filter points after DLT. Increasing this threshold also results in more accurate but sparser $\mathbf{X}_s$.
These factors were combined to generate four sets of $\textbf{X}_{s}$ with various degrees of density as shown in Tab.~\ref{tab:ablate_sfmfilter}. These $\textbf{X}_{s}$ offer geometry guidance to different ratios of the total $N\times H\times W \times 3$ pixels. 
For comparison with the dense predictions, we also compute the AUC of each $\mathbf{X}_s$, which corresponds to the average recall at various error thresholds.
We then feed each $\mathbf{X}_s$ together with the VGGT prediction $\mathbf{X}_d$ into GGPT and evaluate the final AUC of GGPT's complete prediction $\hat{\mathbf{X}}_d$.

\noindent\textbf{Quantitative results.} Tab.~\ref{tab:ablate_sfmfilter} presents the pixel coverage and AUC of the VGGT dense prediction $\mathbf{X}_d$, the various SfM partial-point estimates $\mathbf{X}_s$, and the GGPT reconstruction $\hat{\mathbf{X}}_d$ obtained when combining VGGT and SfM inputs.
First, when using only RoMa matches with a strict filtering scheme, SfM yields a very small yet accurate subset of points. For example, with $\epsilon_{\textrm{DLT}}{=}0.6$ and $\epsilon_{\textrm{reproj}}{=}2$, SfM recovers points for merely 8.5\% of pixels, which contributes to only an AUC@1cm of 6\%. Despite such sparse guidance, GGPT still substantially enhances the dense reconstruction, raising VGGT's AUC@5 cm from 19\% to 26\%. Relaxing the filtering thresholds ($\epsilon_{\textrm{DLT}}{=}0.1$, $\epsilon_{\textrm{reproj}}{=}4$) increases SfM coverage (21\% on ScanNet++), albeit with additional noise. Even so, GGPT continues to improve the dense prediction, achieving an AUC@5 cm of 31\%. The same behaviour is observed across all datasets, demonstrating that GGPT is robust to both sparsity and noise in the SfM input. Fig.~\ref{fig:vis_sfm} provides qualitative examples showing how GGPT effectively exploits highly sparse SfM guidance to refine VGGT outputs.

Next, we consider combining RoMa and UFM matches (implementation details in Sec.~\ref{sec:sfm_impl}). This increases SfM point density but also introduces additional noise, consistent with our observation that UFM attains higher recall but lower accuracy than RoMa (Tab.~\ref{tab:ablate_sfm}). On 4D-DRESS, incorporating UFM increases SfM coverage yet slightly lowers AUC@1 cm, which in turn reduces GGPT performance compared with using RoMa alone—although GGPT still clearly surpasses both the raw SfM points and VGGT predictions. On ScanNet++ and ETH3D, however, GGPT benefits from the richer geometric cues and achieves the best dense reconstructions. Consequently, our default inference setting uses only RoMa on 4D-DRESS and the RoMa s+ UFM ensemble on other datasets.

\begin{figure*}[h]
    \centering
    \includegraphics[width=0.99\linewidth, trim={0cm 0cm 0cm 0.0cm}, clip]{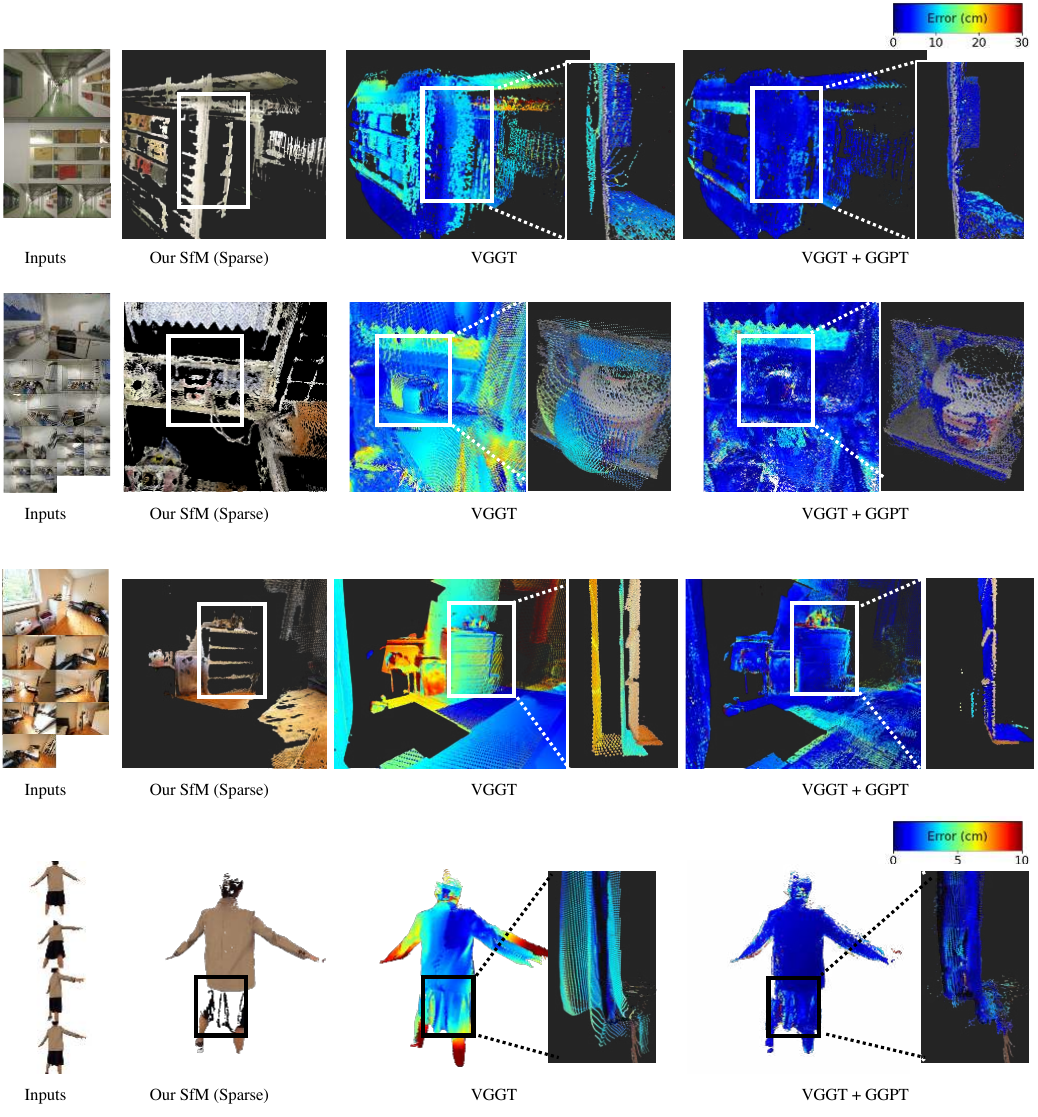}
    \caption{\textbf{GGPT's Robustness to Sparse Guidance.} Test examples from ETH3D~\cite{schoeps2017eth3d} (1st row), ScanNet++~\cite{yeshwanth2023scannet++} (2nd and 3rd rows), and 4D-DRESS~\cite{wang20244ddress} (4th row). Points with predicted confidence below the 10\% quantile are filtered. SfM points are produced with matchings from RoMa~\cite{edstedt2024roma}. We visualise the SfM points and compare error maps before and after GGPT refinement. In the zoomed-in regions, the ground truth (coloured by input RGBs) is overlaid with the predictions (coloured by error) to highlight how our method corrects misaligned input points. SfM guidance provides little support in textureless regions, such as white walls or black cloth, but GGPT effectively propagates sparse guidance to unguided areas, improving the overall accuracy of the point map. Quantitative evaluation can be found in Tab.~\ref{tab:ablate_sfmfilter}.}

    \label{fig:vis_sfm}
\end{figure*}

\begin{table*}[ht!]
    \centering
    \setlength{\tabcolsep}{1.6pt}
     \caption{\textbf{Comparison of GPU Memory Consumption and Inference Speed.} (a) We show the per-stage GPU memory and running time of our pipeline. (b) We compare our method with feed-forward method~\cite{wang2025vggt} and hybrid methods~\cite{duisterhof2025mastrsfm, guedon2025matcha} across varying view counts. }
    \label{tab:memory_speed}
    \begin{tabular}{l ccc c ccc c ccc}
    \toprule
    & \multicolumn{3}{c}{\textbf{Peak GPU Memory (GB)} $\downarrow$} 
    && \multicolumn{3}{c}{\textbf{Inference Time (s)} $\downarrow$}
    && \multicolumn{3}{c}{\textbf{AUC@5/10 cm (\%)} $\uparrow$} \\
    \cmidrule(lr){2-4} \cmidrule(lr){6-8} \cmidrule(lr){10-12}
    & 4 Views & 8 Views & 16 Views 
    && 4 Views & 8 Views & 16 Views
    && 4 Views & 8 Views & 16 Views \\
    \midrule
    \multicolumn{12}{l}{(a) \textbf{Pipeline Breakdown}} \\
    \addlinespace[0.05cm]

    VGGT Initialisation & 6.9 & 7.0 & 8.0 && 0.1 & 0.2 & 0.3 && -- & -- & -- \\

    Our SfM & & & && & & && & & \\
    \hspace{1em} Dense Matcher - RoMa~\cite{edstedt2024roma} 
        & 1.4 & 1.9 & 3.1 && 1.9 & 9.0 & 36 && -- & -- & -- \\
    \hspace{1em} Dense Matcher - RoMa v2~\cite{edstedt2025romav2} 
        & 4.9 & 5.3 & 6.1 && 1.5 & 3.1 & 10 && -- & -- & -- \\
    \hspace{1em} Dense Matcher - UFM~\cite{zhang2025ufm} 
        & 2.2 & 2.7 & 4.1 && 0.5 & 1.7 & 8.8 && -- & -- & -- \\
    \hspace{1em} Bundle Adjustment 
        & 0.3 & 0.8 & 2.1 && 1.0 & 1.9 & 15 && -- & -- & -- \\
    \hspace{1em} DLT Triangulation 
        & 0.7 & 3.1 & 5.9 && $<$0.1 & 0.1 & 0.3 && -- & -- & -- \\
    GGPT Refinement & 1.6 & 3.1 & 4.8 && 2.6 & 4.1 & 6.1 && -- & -- & -- \\
    
    \midrule
    \multicolumn{12}{l}{(b) \textbf{Comparison with Baselines}} \\

    MAST3R-SfM~\cite{duisterhof2025mastrsfm}
        & \textbf{3.3} & \textbf{3.3} & \textbf{3.3} && 21 & 52 & 115 && 37/50 & 39/51 & 40/54 \\
    MAtCha~\cite{guedon2025matcha}
        & 3.3 & 3.6 & 9.4 && 48 & 102 & 242 && 40/52 & 41/53 & 42/56 \\
    VGGT~\cite{wang2025vggt}
        & 6.9 & 7.0 & 8.0 && \textbf{0.1} & \textbf{0.2} & \textbf{0.3} && 27/41 & 23/36 & 19/32 \\
    VGGT + Ours (RoMa + UFM)
        & 6.9 & 7.0 & 8.0 && 7 & 18 & 70 && 41/55 & 47/61 & 49/63 \\
    VGGT + Ours (RoMa v2) 
        & 6.9 & 7.0 & 8.0 &&  5 & 9 & 34  && \textbf{56/66} & \textbf{58/68} & \textbf{63/73} \\
    \bottomrule
    \end{tabular}
\end{table*}

\begin{figure}[h!]
    \centering
    \begin{subfigure}[t]{0.78\columnwidth}
        \centering
        \includegraphics[width=\linewidth]{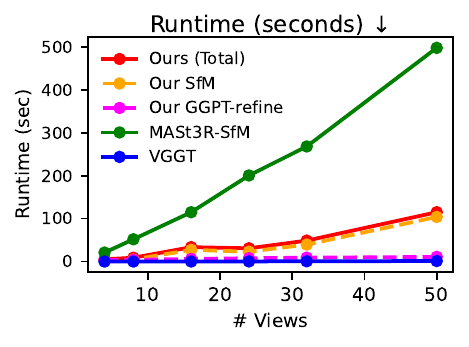}
        \label{fig:runtime}
    \end{subfigure}
    \begin{subfigure}[t]{0.78\columnwidth}
        \centering
        \includegraphics[width=\linewidth]{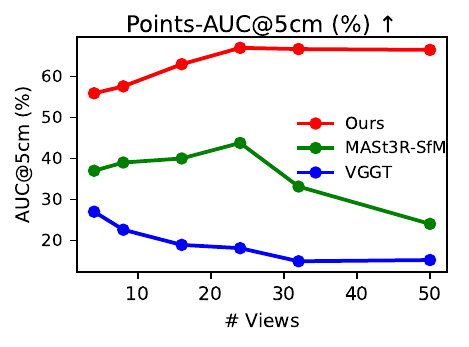}
        \label{fig:points_auc}
    \end{subfigure}
    \vspace{-0.8cm}
    \caption{\textbf{Performance vs. number of views on ETH3D.} Using the latest dense matcher RoMa v2~\cite{edstedt2025romav2}, our method consistently surpasses the baselines with a favourable runtime profile across a wide range of input views, scaling effectively up to 50.}
    \label{fig:scaleup}
\end{figure}

\newpage
\section{Efficiency Analysis}

We provide an analysis of GPU memory consumption and inference time of our method on the ETH3D test set in Tab.~\ref{tab:memory_speed}.

\subsection{Pipeline Breakdown} 

Our dense point map prediction pipeline comprises three stages: feed-forward initialisation, SfM reconstruction, and GGPT refinement. We use VGGT~\cite{wang2025vggt} as the representative feed-forward model for analysis, as other feed-forward approaches~\cite{wang2025pi3,keetha2025mapanything} exhibit comparable inference costs.

\noindent\textbf{GPU memory consumption.} VGGT demands the most GPU memory, driven by full self-attention across many multi-view patches and a deep hierarchy of attention layers. The dense matcher~\cite{edstedt2024roma,zhang2025ufm}, bundle adjustment, DLT triangulation, and GGPT refinement incur smaller memory footprints. The stages of DLT triangulation and GGPT refinement also support iterative batch processing: input images or matches can be divided into independent batches to reduce memory usage when handling larger numbers of views. 

\noindent\textbf{Inference time.} The dense matcher RoMa~\cite{edstedt2024roma} is the dominant source of running time in our pipeline, as it is applied to all image pairs. This cost can be reduced by restricting matching to overlapping views. Bundle adjustment is lightweight, requiring less than two seconds for up to eight views, and for sixteen or more views its runtime can be further reduced by lowering $n_{\textrm{BA}}$ to 512, as shown in Tab.~\ref{tab:ablate_sfm}. DLT triangulation adds only negligible overhead. GGPT refinement over all patches contributes additional seconds. After the initial submission, we find that adopting more recent RoMa v2~\cite{edstedt2025romav2} as the dense matcher further improves efficiency and accuracy.

\subsection{Comparison with baselines} 

Compared with feed-forward models such as VGGT~\cite{wang2025vggt}, our method maintains a similar memory footprint while introducing a modest increase in inference time, yielding substantially higher geometric accuracy. Although this makes the approach slower than pure feed-forward alternatives, we believe the additional runtime represents a worthwhile trade-off in applications where precise geometry is critical.

Additionally, with RoMa v2 as the dense matcher, our method can scale to more views while maintaining a favourable speed-accuracy trade-off. As shown in Fig.~\ref{fig:scaleup}, when scaling to 50 input views, our approach is five times faster than MASt3R-SfM and substantially outperforms the baselines across varying numbers of views.

\section{Comparison with Dense-SfM~\cite{densesfm}}

Similar to our SfM, Dense-SfM~\cite{densesfm} adopts RoMa dense matcher to extract pairwise correspondences. In contrast with our SfM, Dense-SfM integrates dense matchings into an incremental COLMAP pipeline, requires a Gaussian Splatting~\cite{kerbl20233d} process to extend multi-view tracks, and performs iterative track refinement and global BA. We compare our SfM with Dense-SfM to demonstrate the effectiveness and efficiency of our method.

Since the Dense-SfM code is not publicly available, we follow the evaluation protocol in their paper and evaluate our SfM on the same benchmarks, \ie the ETH3D~\cite{schoeps2017eth3d} sequences with complete input views and the IMC~\cite{imc} test set. We use RoMa for exhaustive matching, consistent with their setting, and compare camera pose accuracy with the results reported in Tab. 2 of their paper. Since both methods use the same matching procedure, we exclude matching time from the runtime comparison for clarity. Dense-SfM further reports around 5 minutes for 3D Gaussian training on ETH3D (Sec. 8.1 of their supplementary), in addition to iterative track refinement and BA, whereas our SfM requires only a single sparse BA. 

\begin{table}[h!]
    \centering
    \footnotesize
    \setlength{\tabcolsep}{3pt} 
    \caption{\textbf{Comparison with Dense-SfM.}}
    \vspace{-3mm}
    \begin{tabular}{@{}l cc  cc@{}}
    \toprule
      Method & \multicolumn{2}{c}{\textbf{ETH3D} ($\sim36$ views)}  &  \multicolumn{2}{c}{\textbf{IMC} ($\sim9$ views) }  \\
      & AUC@1/3$\degree$ & Runtime (s)  &  AUC@3/5$\degree$ & Runtime (s) \\
      \midrule
      Dense-SfM & 61/78 & $>$300 &48/61 &  unavailable     \\
      Our SfM   & \textbf{81/85} & \textbf{38} & \textbf{56/66} &\textbf{ 1.5} \\
      \bottomrule
    \end{tabular}
    \label{tab:cmp_dsfm}
\end{table}
\noindent Tab.~\ref{tab:cmp_dsfm} shows that our method achieves higher accuracy while being substantially faster, thanks to our feed-forward initialisation and effective track selection strategy.

\section{Additional Implementation Details}

\subsection{Structure From Motion}
\label{sec:sfm_impl}
\noindent\textbf{Image Resolutions.} All images are resized by setting a target width while preserving the original aspect ratio. For the feed-forward models~\cite{wang2025vggt,wang2025pi3,guedon2025matcha}, we use a width of 518 pixels. For the RoMa dense matcher~\cite{edstedt2024roma}, we provide full-HD inputs (width~=~1920), which we find improves matching accuracy. For the UFM matcher~\cite{zhang2025ufm}, we use the UFM-Refine model\footnote{\url{https://huggingface.co/infinity1096/UFM-Refine}}, which internally resizes inputs to width~=~560. All output matches are then resized to width~=~518. Bundle adjustment, DLT triangulation, GGPT refinement, and final evaluation are performed at this resolution.

\noindent\textbf{Dense matchers ensemble.} We combine the match predictions from RoMa~\cite{edstedt2024roma} and UFM~\cite{zhang2025ufm} to produce the global correspondences and confidence tensors $(\mathbf{T}, \mathbf{C})$ in Eq. (2) of the main paper. Concretely, we denote the tensors predicted by RoMa and UFM by $(\mathbf{T}_{\textrm{1}}, \mathbf{C}_{\textrm{1}})$ and $(\mathbf{T}_{\textrm{2}}, \mathbf{C}_{\textrm{2}})$ respectively. We compute their cycle consistency tensor (Eq. (3) of the main paper) and denote them as $\mathbf{\varepsilon}_{\textrm{1}}$ and $\mathbf{\varepsilon}_{\textrm{2}}$. 

\begin{align}
\mathbf{\varepsilon}_{\textrm{1}}[i,k,\mathbf{u}] = \big\|\mathbf{T}_{\textrm{1}}[k,i,\mathbf{T}_\textrm{1}[i,k,\mathbf{u}]] - \mathbf{u}\big\|_2 \\
\mathbf{\varepsilon}_{\textrm{2}}[i,k,\mathbf{u}] = \big\|\mathbf{T}_{\textrm{2}}[k,i,\mathbf{T}_\textrm{2}[i,k,\mathbf{u}]] - \mathbf{u}\big\|_2 
\end{align}

Each query pixel $\mathbf{u}$ in source view $i$ has two predicted correspondences in target view $k$ from the two matchers. We choose the correspondence with a smaller cycle consistency error. Formally,

\begin{align}
\mathbf{T} = [\mathbf{\varepsilon}_{\textrm{1}}<\mathbf{\varepsilon}_{\textrm{2}}]\odot \mathbf{T}_{\textrm{1}} + [\mathbf{\varepsilon}_{\textrm{1}}>\mathbf{\varepsilon}_{\textrm{2}}]\odot \mathbf{T}_{\textrm{2}} \\
\mathbf{C} = [\mathbf{\varepsilon}_{\textrm{1}}<\mathbf{\varepsilon}_{\textrm{2}}]\odot \mathbf{C}_{\textrm{1}} + [\mathbf{\varepsilon}_{\textrm{1}}>\mathbf{\varepsilon}_{\textrm{2}}]\odot \mathbf{C}_{\textrm{2}} 
\end{align}
Ensembling two matchers improves the completeness of SfM points, as shown in Sec.~\ref{sec:sfm_ablate} and Tab.~\ref{tab:ablate_sfm}, and it subsequently improves the final dense prediction, as shown in Sec.~\ref{sec:vary_sfm_density} and Tab.~\ref{tab:ablate_sfmfilter}. 


\noindent\textbf{Direct linear triangulation.} We provide the mathematical formulation of the direct linear triangulation (DLT) used in our SfM pipeline.
The objective of DLT is to estimate the 3D homogeneous points $\tilde{\mathbf{X}}\in \mathbb{R}^{M\times 4}$ that minimise the 2D reprojection error, given multi-view tracks in 2D homogeneous coordinates ${\mathbf{T}} \in \mathbb{R}^{N\times M\times 2}$ and its visibility mask $\mathbf{M} \in \mathbb{R}^{N\times M}$ \footnote{Here, $({\mathbf{T}},\mathbf{M})$ are derived from $({\mathbf{T}_{\textrm{DLT}}} \in \mathbb{R}^{N\times H\times M \times 2}, {\mathbf{M}_{\textrm{DLT}}} \in \mathbb{R}^{N\times H\times M})$ as introduced in Sec.~3.1 of the main paper.}, as well as the BA-estimated intrinsics  $\mathbf{K} \in \mathbb{R}^{N\times 3\times 3}$ and extrinsics $\mathbf{E} \in \mathbb{R}^{N\times 3\times 4}$. Here $N$ denotes the number of views, and $M$ denotes the number of multi-view tracks, which is also the number of points to estimate.

For each 3D point $\tilde{\mathbf{X}}_{[i]} \in \mathbb{R}^4$, its 2D reprojections in $N$ views can be represented in 2D homogeneous coordinates as $\mathbf{P}\tilde{\mathbf{X}}_{[i]} \in \mathbb{R}^{N\times 3}$, where $\mathbf{P}=\mathbf{K}\mathbf{E}$ is the multi-view projection matrix. As the 2D projections should be close to the observation ${\mathbf{T}}_{[:,i]}\in \mathbb{R}^{N\times 2}$ in visible views, the objective of DLT is to estimate

\begin{align}
\label{eq:dlt_func}
\tilde{\mathbf{X}}_{[i]}
= \operatorname*{arg\,min}_{\tilde{\mathbf{X}}_{[i]}}
\big\|
  \mathbf{M}_{[:, i]} \odot
  \big(\mathrm{deh}(\mathbf{P}\tilde{\mathbf{X}}_{[i]}) - \mathbf{T}_{[:,i]}\big)
\big\|
\end{align}
where $\mathrm{deh}(\cdot)$ is the de-homogenising operation.
\begin{align}
    \mathrm{deh}(\mathbf{Y}) = \mathbf{Y}_{[:,[0,1]]}/\mathbf{Y}_{[:,2]}
\end{align}
This can be approximated by a least-squares problem

\begin{align}
\tilde{\mathbf{X}}_{[i]} 
\;\sim\;
\operatorname*{arg\,min}_{\substack{\mathbf{x} \in \mathbb{R}^4 \\ \|\mathbf{x}\| = 1}}
\;\big\| \mathbf{A}_i \mathbf{x} \big\|.
\end{align}
\begin{align}
\mathbf{A}_i =
\left[
\begin{array}{c}
\vdots \\[2pt]
\mathbf{M}_{[n,i]}(\mathbf{T}_{[n,i,0]}\,\mathbf{P}_{[n,2,:]} - \mathbf{P}_{[n,0,:]}) \\[4pt]
\mathbf{M}_{[n,i]}(\mathbf{T}_{[n,i,1]}\,\mathbf{P}_{[n,2,:]} - \mathbf{P}_{[n,1,:]}) \\[2pt]
\vdots
\end{array}
\right]_{\substack{n = 1, \dots, N}}.
\end{align}
\noindent Here, $\sim$ denotes equality up to scale. The solution has a closed-form via singular value decomposition. Moreover, all $\tilde{\mathbf{X}}_{[i]}$ and $\mathbf{A}_i$ can be processed in parallel to solve for $\mathbf{X}$ efficiently using CUDA tensor operations in PyTorch.  

After DLT triangulation, for each estimated point, we compute its maximal 2D reprojection error and minimal triangulation angle across all visible views, both of which can be efficiently evaluated via tensor operations. Points with maximal 2D reprojection error exceeding 4 pixels or minimal triangulation angle below 3° are discarded. The remaining points constitute our final geometry guidance, denoted as $\mathbf{X}_s \in \mathbb{R}^{M'\times 3}$.  

To establish correspondences between the geometry estimate $\mathbf{X}_s \in \mathbb{R}^{M'\times 3}$ and the initial dense prediction $\mathbf{X}_d \in \mathbb{R}^{N\times H\times W \times 3}$, we project $\mathbf{X}_s$ back to their visible views, round the 2D coordinates, and assign the 3D coordinates to the corresponding integer pixels. For pixels receiving multiple projections, we compute the average of the 3D estimates.

\subsection{Geometry-Grounded Point Transformer}

\noindent\textbf{Architecture.}
The point transformer encoder begins with an MLP embedding layer, followed by five down-pooling and four up-pooling stages, producing features of dimensionality $V = 96$. The down-pooling stages include $(2, 2, 2, 6, 2)$ attention blocks with hidden dimensions $(64, 96, 128, 256, 512)$, with each stage followed by a down-sampling grid-pooling layer. The up-pooling stages contain $(2, 2, 2, 2)$ attention blocks with hidden dimensions $(256, 128, 96, 96)$, each stage preceded by an up-sampling grid-pooling layer. The point cloud is voxelised at a grid resolution of 384, with strides of $(2, 2, 2, 2)$ for the grid-pooling layers. For details on the attention blocks and grid pooling, see~\cite{wu2024ptv3}. Each point feature output by the point transformer backbone is concatenated with its input embedding $\mathbf{z}$ and passed through a three-layer MLP with two intermediate ReLU activations and 256 hidden dimensions, producing four outputs: three for 3D coordinate residuals and one for confidence. The complete GGPT model has a total of 53\,million parameters.

\noindent\textbf{Training dataset.} We curate the training dataset by sampling 20k multi-view sequences from the 856 training scenes of ScanNet++~\cite{yeshwanth2023scannet++}. Each multi-view sequence consists of 4-16 input views, which are randomly selected following the view-sampling algorithm in ~\cite{keetha2025mapanything}. For each sequence, we use VGGT~\cite{wang2025vggt} to predict dense predictions $\mathbf{X}_d$ and provide camera-pose initialisation for our SfM. We use the configurations described in Sec.~\ref{sec:sfm_impl} to run our SfM pipeline except we only use RoMa~\cite{edstedt2024roma} matcher for dense correspondence extraction. The training dataset generation used eight NVIDIA GH200 GPUs for one day.

\noindent\textbf{Training hyperparameters.} We set the weight of the identity consistency loss as $\lambda_{\textrm{id}}=1$, and the weight of the confidence term as $\alpha =0.2$. We use Adam optimiser with a learning rate of $1e-4$ and a batch size of 24. The model is trained for 100k iterations, which is performed on eight NVIDIA GH200 GPUS for one day.

\noindent\textbf{Patch-based processing.} We patchify the input point cloud during both training and inference. We calculate the scene radius $R$ as $3\times \textrm{std}(\mathbf{X}_s)$ and sample each input chunk with a half width of $r=0.2R$. To avoid out-of-memory issue, we input a maximal number of 400k points for each chunk. If the sampled chunk contains more points than this, which happens sporadically, we reduce the width by $\times 0.9$. During training, we randomly sample a chunk from each scene by sampling an anchor point from $\mathbf{X}_s$ and cropping a box centred at it. Each batch thus contains 24 chunks from 24 possibly different scenes. During inference, we cover the whole point cloud by iteratively sampling an anchor point from $\mathbf{X}_s$ and taking its surrounding points. At each iteration, we only sample anchors which are not covered previously, but still include other points in the box even if they have been processed. This ensures some overlapping between chunks while saving redundancy. sampling terminates once all points in $\mathbf{X}_s$ are covered. Empirically, this yields around 50 chunks for the whole scene on average. For unprocessed points in $\mathbf{X}_d$, which usually make up fewer than 5\% of the scene and are regions without any valid geometry constraints, we retain their initial dense predictions in the final output.

\subsection{Point Evaluation}
\label{sec:points_eval}
\noindent\textbf{Ground-truth reference.} We use the ground-truth metric point map $\mathbf{X}_{\textrm{gt}} \in \mathbb{R}^{N\times H \times W \times 3}$ at a resolution of $W=518$ as the reference for evaluation. For ETH3D~\cite{schoeps2017eth3d} and ScanNet++~\cite{yeshwanth2023scannet++}, the provided depth maps are first undistorted to linear camera models and then unprojected to obtain the ground-truth point maps. For T\&T~\cite{Knapitsch2017tandt}, we employ PyTorch3D to render the provided laser point clouds into 2D depth maps, remove occluded pixels, and subsequently unproject them back to 3D. For Blender-rendered datasets, including 4D-DRESS~\cite{wang20244ddress} and MV-dVRK~\cite{dvrk-smv}, we unproject the ground-truth depth maps directly output by Blender. Owing to the incompleteness of laser scans in real-world datasets or the invalid background regions in synthetic datasets, each $\mathbf{X}_{\textrm{gt}}$ is associated with a valid mask $\mathbf{M}_{\textrm{gt}} \in [0,1]^{N\times H \times W}$. Accuracy is evaluated exclusively within the valid regions defined by $\mathbf{M}_{\textrm{gt}}$.

\vspace{1ex}\noindent\textbf{Procrustes alignment.} Given the ground-truth metric point map $\mathbf{X}_{\textrm{gt}} \in \mathbb{R}^{N\times H \times W \times 3}$ and predicted scale-invariant point map $\mathbf{X}_{\textrm{pred}} \in \mathbb{R}^{N\times H \times W \times 3}$, we first estimate the optimal similarity transformation $(s, \mathbf{R} \in SO(3), \mathbf{t} \in \mathbb{R}^3)$ that aligns $\mathbf{X}_{\textrm{pred}}$ to $\mathbf{X}_{\textrm{gt}}$. Following prior work~\cite{wang2025vggt,wang2024dust3r,mast3r_eccv24}, we use the Umeyama~\cite{umeyama2002least} algorithm to compute the alignment using $\mathbf{M}_{\textrm{gt}} \odot \mathbf{X}_{\textrm{gt}}$ and $\mathbf{M}_{\textrm{gt}} \odot \mathbf{X}_{\textrm{pred}}$. To improve robustness against outliers in the prediction, we employ the RANSAC-based robust Umeyama implemented by pycolmap~\cite{colmap}\footnote{\url{https://colmap.github.io/pycolmap/pycolmap.html\#pycolmap.estimate\_sim3d_robust}} with a minimum inlier ratio of 0.8 and a maximum error of 3~cm. The aligned predicted points are denoted as $\mathbf{X'}_{\textrm{pred}} = s \mathbf{R} \mathbf{X}_{\textrm{pred}} + \mathbf{t}$.

\vspace{1ex}\noindent\textbf{Metrics.} 
For each pixel with valid ground truth, we compute the 3D Euclidean error as
\begin{align}
    \varepsilon[n, \mathbf{u}] = \|\mathbf{X}_{\textrm{gt}}[n, \mathbf{u}] - \mathbf{X}'_{\textrm{pred}}[n, \mathbf{u}]\|_2,
\end{align}
where $n$ indexes frames and $\mathbf{u}$ indexes pixels.  

Given an error threshold $\tau$, the recall at threshold $\tau$ (Recall@$\tau$) is defined as the fraction of pixels with error below $\tau$:
\begin{align}
    \mathrm{Recall}@\tau = \frac{1}{|M|} \sum_{(n,\mathbf{u}) \in M} \mathbf{1}\{ \varepsilon[n, \mathbf{u}] < \tau \},
\end{align}
where $M = \{(n,\mathbf{u}) \mid \mathbf{M}(n,\mathbf{u}) = 1\}$ is the set of pixels with valid ground truth, and $|M|$ is its cardinality.  

Finally, the area under the curve up to threshold $\tau$ (AUC@$\tau$) is computed by averaging recall over all integer thresholds $1,2,\dots,\tau$:
\begin{align}
    \mathrm{AUC}@\tau = \frac{1}{\tau} \sum_{k=1}^{\tau} \mathrm{Recall}@k.
\end{align}

\section{Depth Evaluation}
In addition to point-based metrics, we can also evaluate depth prediction. Specifically, after the Procrustes alignment in Sec.~\ref{sec:points_eval}, we use the estimated similarity transformation to transform the camera extrinsics into the same world coordinate system as $\mathbf{X}_{\textrm{pred}}'$. We then project $\mathbf{X}_{\textrm{pred}}'$ to each view to obtain per-view depth predictions, which are compared against ground truths. Note that no further scaling is applied to the predicted depths, as the similarity transformation already accounts for scale.

Following standard depth evaluation protocol~\cite{zuo2025omni}, we report the absolute relative error (Rel) and inlier ratio at 1.01 and 1.03 ($\tau$) before and after GGPT refinement. Tab.~\ref{tab:depth} shows that GGPT consistently improves the depths.
\begin{table}[h!]
    \centering
    \footnotesize
    \setlength{\tabcolsep}{3pt} 
    \caption{\textbf{Depth metrics.} We report relative absolute error (Rel \%) and inlier ratio at 1.01/1.03 ($\tau$ \%) on \dscd{cross-domain} and \dsod{out-of-domain} test sets.}
    \label{tab:depth}
\begin{tabular}{@{}l cc c cc c cc c cc@{}}
\toprule
& \multicolumn{2}{c}{\dscd{ETH3D}} 
&& \multicolumn{2}{c}{\dscd{T\&T}} 
&& \multicolumn{2}{c}{\dsod{4DDress}} 
&& \multicolumn{2}{c}{\dsod{MV-dVRK}}  \\

&  Rel$\downarrow$ & $\tau\uparrow$
&&Rel$\downarrow$ & $\tau\uparrow$ 
&&Rel$\downarrow$ & $\tau\uparrow$
&& Rel$\downarrow$ & $\tau\uparrow$ \\
\midrule

VGGT& 4.3&27/57& & 2.9&40/79& & 75.4&0/0 && 31.8&1/3 \\
 + Ours&\textbf{2.3}&\textbf{63/85}& & \textbf{2.3}&\textbf{58/85}& & \textbf{7.9}&\textbf{72/81} && \textbf{11.4}&\textbf{40/60} \\
\midrule
Pi3& 3.7&27/60& & 2.9&38/75& & 8.6&7/20 && 9.4&10/29 \\
 + Ours&\textbf{2.2}&\textbf{59/86}& & \textbf{2.1}&\textbf{58/87}& & \textbf{4.8}&\textbf{27/66} && \textbf{5.9}&\textbf{34/63} \\
\bottomrule
\end{tabular}
\end{table}

\section{Additional Qualitative Results}
We visualise additional results in Fig.~\ref{fig:supp_eth3d} (ETH3D), Fig.~\ref{fig:supp_scannetpp} (ScanNet++), Fig.~\ref{fig:supp_tandt} (T\&T), Fig.~\ref{fig:supp_mvdvrk} (MV-dVRK), and Fig.~\ref{fig:supp_4ddress} (4DDress). We apply our method to existing dense reconstruction methods~\cite{wang2025vggt,wang2025pi3,keetha2025mapanything,guedon2025matcha,duisterhof2025mastrsfm} and compare the error map of the point maps before and after our refinement. Note that we only visualise the error map for pixels with valid ground truth reference. Additionally, following prior work~\cite{wang2025vggt, wang2025pi3, keetha2025mapanything}, we filter out points with predicted confidence below the 10\% quantile.

Visual results demonstrate that our method can improve various baselines consistently across different datasets. On \dsid{ScanNet++}~\cite{yeshwanth2023scannet++}, the within-domain test set, Pi3~\cite{wang2025pi3} achieves near-saturated performance because it is trained on a large corpus of similar indoor scenes, leaving little room for improvement. However, our method can still noticeably improve its performance on other datasets, particularly out-of-domain \dsod{4DDress} and \dsod{MV-dVRK} datasets.

With these visualisations, we also acknowledge that our method shows limitations in refining local details, exhibiting red error patterns in some detailed areas, failing to correct geometry in regions without geometry guidance, and sometimes exhibiting patchy artefacts. For these limitations, we further discuss possible solutions and future directions in the next section.

\section{Limitations and Future Directions}

\vspace{2ex}\noindent\textbf{End-to-end 3D reconstruction.} Our pipeline currently consists of three sequential stages to obtain a multi-view consistent reconstruction. Because the SfM geometry guidance and the GGPT dense refinement operate in separate steps, any error in the SfM stage propagates directly into the refinement stage. As a result, GGPT struggles in regions where the geometry guidance is inaccurate or incomplete. A natural direction for future work is to develop an end-to-end approach that jointly estimates geometry based on multi-view constraints and predicts the dense reconstruction. One possibility is to integrate 2D image cues into the point embeddings and train the 3D point transformer to refine points directly through 2D photometric consistency.

\vspace{2ex}\noindent\textbf{Improving point transformer architecture.} As described in Sec. 3.2, our current GGPT adopts patch-based processing strategy and confidence-weighted fusion. This saves memory consumption and allows us to process large-scale scenes. It is also used by point cloud denoising networks~\cite{vogel2024p2p,rakotosaona2020pointcleannet}. Additionally, we find in our ablation (Tab.~\ref{tab:ablation_ggpt}) that a small patch size enables the network to improve both fine-level accuracy and generalisation. However, this strategy also has two downsides. (1) It may result in some patch-wise artefacts and non-smooth transitions between different patches. (2) It sacrifices long-range spatial dependency, since points in two non-overlapping patches cannot attend to each other. As a consequence, our prediction still exhibits error in regions which do not have geometry guidance and need to leverage guidance from distant areas. A future direction will be to develop a multi-scale hierarchical point transformer to improve the long-range dependency and global smoothness while maintaining the accuracy in local detail refinement.

\vspace{2ex}\noindent\textbf{Scaling up training datasets.} Our current GGPT is trained solely on indoor scenes from ScanNet++~\cite{yeshwanth2023scannet++}, yet already generalises to cross-domain~\cite{schoeps2017eth3d,Knapitsch2017tandt} and out-of-domain~\cite{wang20244ddress,dvrk-smv} datasets. Thanks to our efficient SfM pipeline and scalable point transformer architecture, training GGPT on larger datasets is practical. Future work could incorporate more diverse datasets, such as Tartan~\cite{tartanair2020iros} and BMVS~\cite{bmvs}, to further enrich the model with general 3D geometry priors, which is expected to improve sharp, fine-grained geometric details in the dense reconstruction.

\vspace{2ex}\noindent\textbf{Extending to large numbers of input views and dense captures.}
Our work targets 3D reconstruction from a limited set of unposed input views, characterised by discrete spatial distribution and random overlap. Accordingly, we conduct experiments using at most 16 input views. Scaling up to captures with hundreds or thousands of images—typical in large outdoor or indoor scenes—would require further optimisation to keep computation manageable. One avenue is to replace VGGT with recent memory-efficient feed-forward variants~\cite{shen2025fastvggt,liu2025vggtxvggtmeetsdense,deng2025vggtlongchunkitloop} for initialisation. Another is to avoid exhaustive pairwise matching by adopting the strategy of MAST3R-SfM~\cite{duisterhof2025mastrsfm}, running matchers only on overlapping image pairs, with overlap predicted by a feed-forward model. Additionally, our focus on discrete input views means that we do not treat video-based settings, where frames provide continuous coverage with dense overlap. In such cases, camera poses are typically obtained with classical SfM pipelines~\cite{colmap,revisitedsfm}, and dense reconstruction, given known camera parameters, is usually handled by Multi-view Stereo methods~\cite{deepmvs,zhe2023geomvsnet,izquierdo2025mvsanywhere,wang2025lightweightmvs}. The reliance on camera parameters and the small triangulation angles involved (often below $10^\circ$) places MVS methods in a different regime from ours. Nonetheless, it would be interesting to explore as a future direction whether our idea of geometry-guided 3D point transformer could also benefit dense-overlap scenarios.


\endgroup

\newcommand{\compimgwidth}{0.15\textwidth}
\newcommand{\inputheight}{0.19\textwidth}
\newcommand{\titletext}{\scriptsize}
\newcommand{\MethodA}{VGGT~\cite{wang2025vggt}}
\newcommand{\Methoda}{vggt}
\newcommand{\MethodB}{Pi3~\cite{wang2025pi3}}
\newcommand{\Methodb}{pi3}
\newcommand{\MethodC}{MapAnything~\cite{keetha2025mapanything}}
\newcommand{\Methodc}{ma}
\newcommand{\MethodD}{MAtCha~\cite{guedon2025matcha}}
\newcommand{\Methodd}{matcha}
\newcommand{\MethodE}{MASt3R-SfM~\cite{duisterhof2025mastrsfm}}
\newcommand{\Methode}{msfm}

\newcommand{\ShowComparisonb}[3]{%
\begin{tabular}{@{}m{0.16\textwidth}@{\hspace{1mm}}c@{\hspace{1mm}}c@{\hspace{1mm}}c@{\hspace{1mm}}c@{\hspace{1mm}}c@{}}

  \setlength{\tabcolsep}{5pt}
    \makebox[\linewidth][c]{\titletext Input}&
    \titletext  \MethodA &
    \titletext  \MethodB &
    \titletext  \MethodC &
    \titletext  \MethodD &
    \titletext  \MethodE \\
\multirow[t]{3}{*}{\makebox[\linewidth][c]{\raisebox{-0.12\textwidth}{%
    \includegraphics[height=#2\textwidth]{#1/input.jpg}%
}}} &
    \includegraphics[width=\compimgwidth]{#1/\Methoda-init_score.jpg} &
    \includegraphics[width=\compimgwidth]{#1/\Methodb-init_score.jpg} &
    \includegraphics[width=\compimgwidth]{#1/\Methodc-init_score.jpg} &
    \includegraphics[width=\compimgwidth]{#1/\Methodd-init_score.jpg} &
    \includegraphics[width=\compimgwidth]{#1/\Methode-init_score.jpg} \\

    \\[-2ex]

&
    \titletext  \MethodA \ + \ Ours &
    \titletext  \MethodB \ + \ Ours &
    \titletext  \MethodC \ + \ Ours &
    \titletext  \MethodD \ + \ Ours &
    \titletext  \MethodE \ + \ Ours \\

 &
    \includegraphics[width=\compimgwidth]{#1/\Methoda-ours_score.jpg} &
    \includegraphics[width=\compimgwidth]{#1/\Methodb-ours_score.jpg} &
    \includegraphics[width=\compimgwidth]{#1/\Methodc-ours_score.jpg} &
    \includegraphics[width=\compimgwidth]{#1/\Methodd-ours_score.jpg} &
    \includegraphics[width=\compimgwidth]{#1/\Methode-ours_score.jpg}   
  \end{tabular} 
\vspace{2ex}
}

\newcommand{\ShowComparisonc}[4]{%
\begin{tabular}{@{}m{0.16\textwidth}@{\hspace{1mm}}c@{\hspace{1mm}}c@{\hspace{1mm}}c@{\hspace{1mm}}c@{\hspace{1mm}}c@{}}

  \setlength{\tabcolsep}{5pt}
    \makebox[\linewidth][c]{\titletext Input}&
    \titletext  \MethodA &
    \titletext  \MethodB &
    \titletext  \MethodC &
    \titletext  \MethodD &
    \titletext  \MethodE \\
\multirow[t]{3}{*}{\makebox[\linewidth][c]{\raisebox{-0.12\textwidth}{%
    \includegraphics[height=#2\textwidth]{#1/input.jpg}%
}}} &
    \includegraphics[width=\compimgwidth]{#1/\Methoda-init_score.jpg} &
    \includegraphics[width=\compimgwidth]{#1/\Methodb-init_score.jpg} &
    \includegraphics[width=\compimgwidth]{#1/\Methodc-init_score.jpg} &
    \includegraphics[width=\compimgwidth]{#1/\Methodd-init_score.jpg} &
    \includegraphics[width=\compimgwidth]{#1/\Methode-init_score.jpg} \\
    \\[-2ex]
&
    \titletext  \MethodA \ + \ Ours &
    \titletext  \MethodB \ + \ Ours &
    \titletext  \MethodC \ + \ Ours &
    \titletext  \MethodD \ + \ Ours &
    \titletext  \MethodE \ + \ Ours \\

 &
    \includegraphics[width=\compimgwidth]{#1/\Methoda-ours_score.jpg} &
    \includegraphics[width=\compimgwidth]{#1/\Methodb-ours_score.jpg} &
    \includegraphics[width=\compimgwidth]{#1/\Methodc-ours_score.jpg} &
    \includegraphics[width=\compimgwidth]{#1/\Methodd-ours_score.jpg} &
    \includegraphics[width=\compimgwidth]{#1/\Methode-ours_score.jpg}   \\
 &
  &
  &
  &
  &
  \includegraphics[width=\compimgwidth]{fig/supp-figs/errorbar_#3.jpg} 
  \end{tabular} 
}

\begin{figure*}[p]   
\centering
\ShowComparisonb{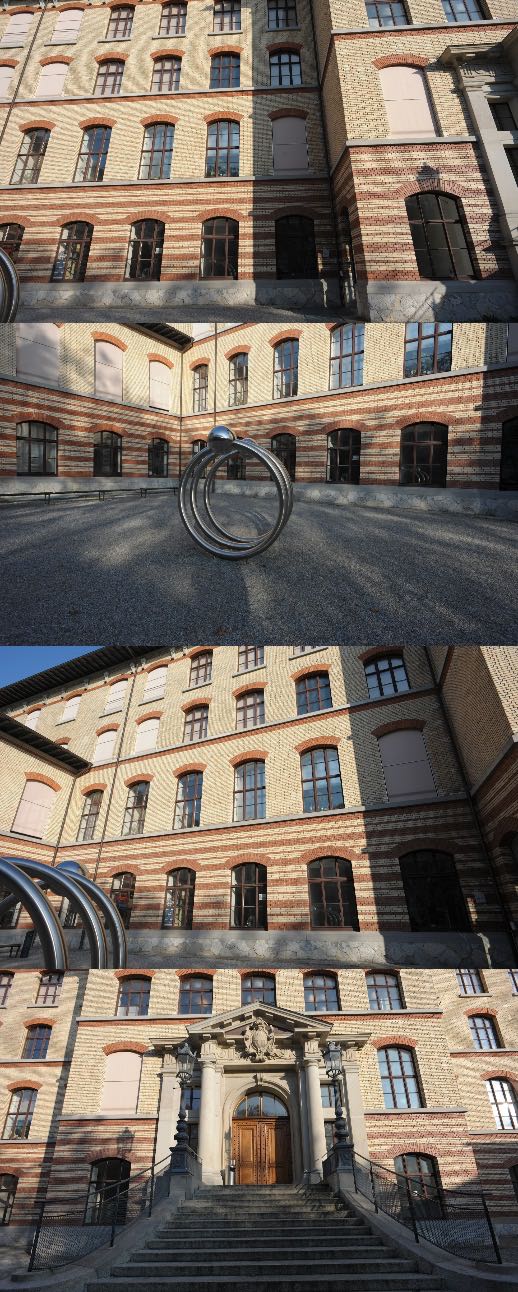}{0.19}{1}
\ShowComparisonb{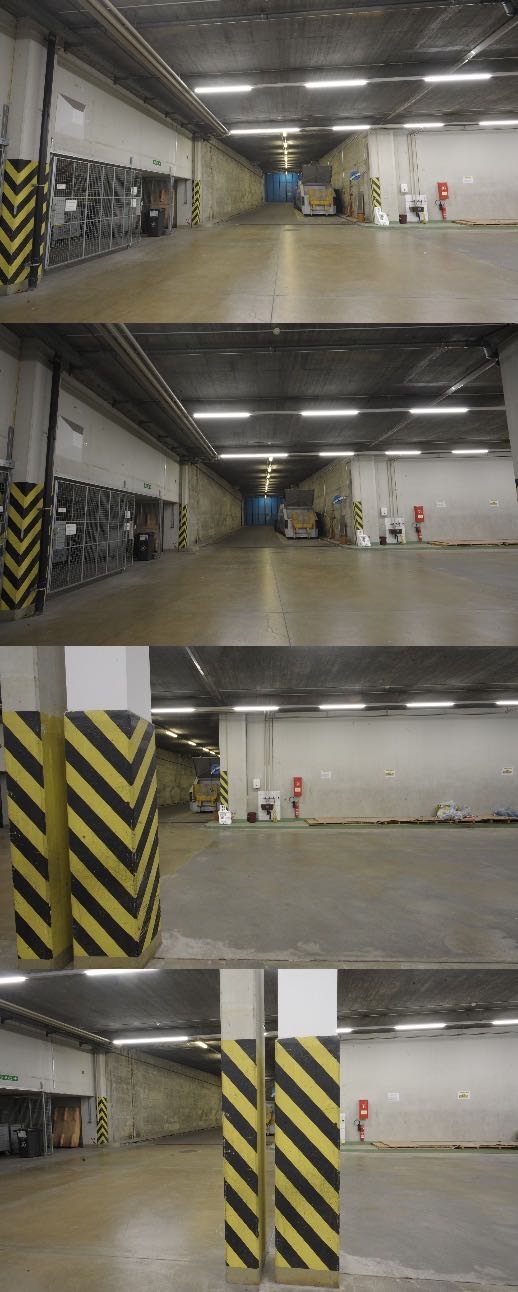}{0.19}{0.3}
\ShowComparisonb{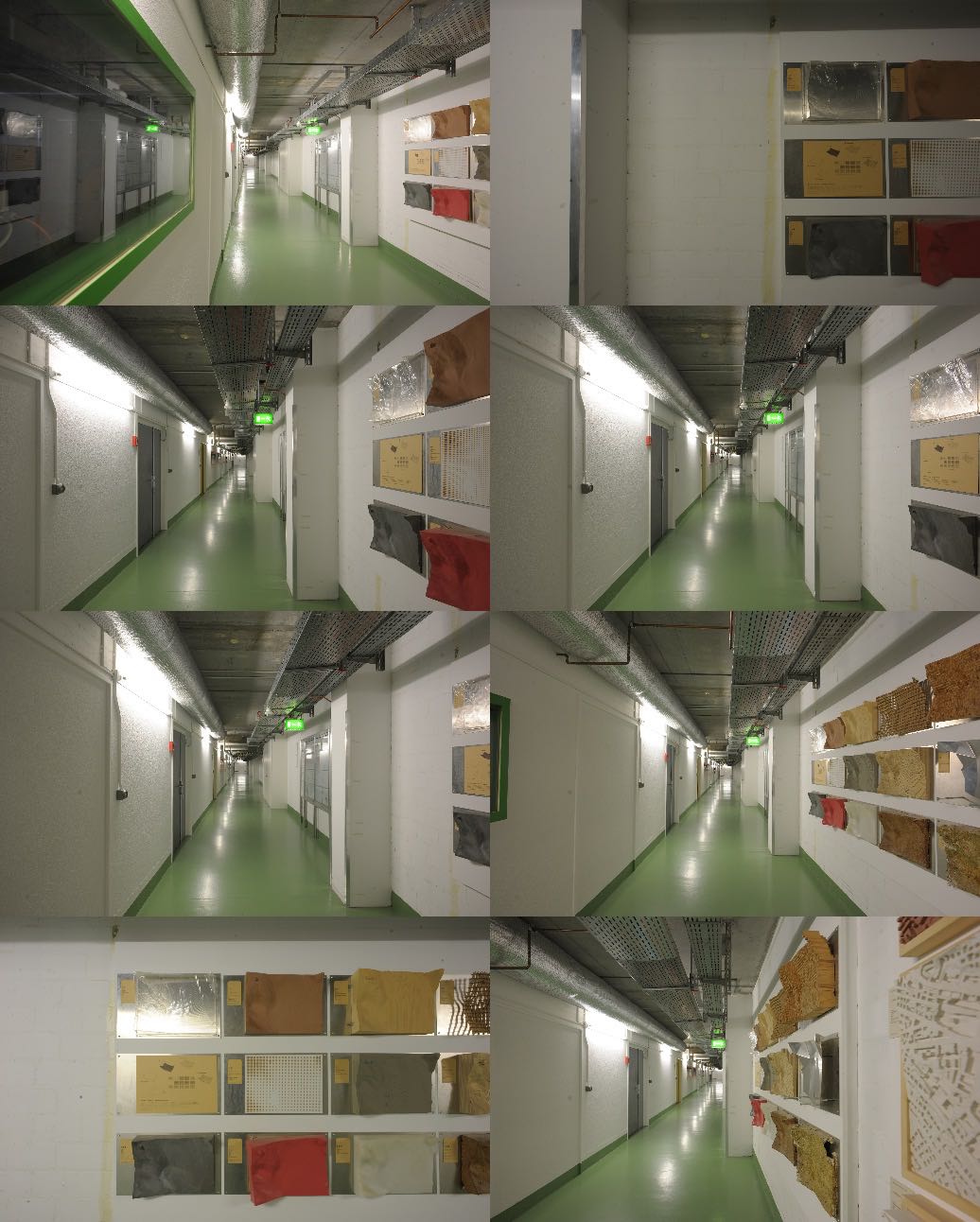}{0.19}{0.3}
\ShowComparisonc{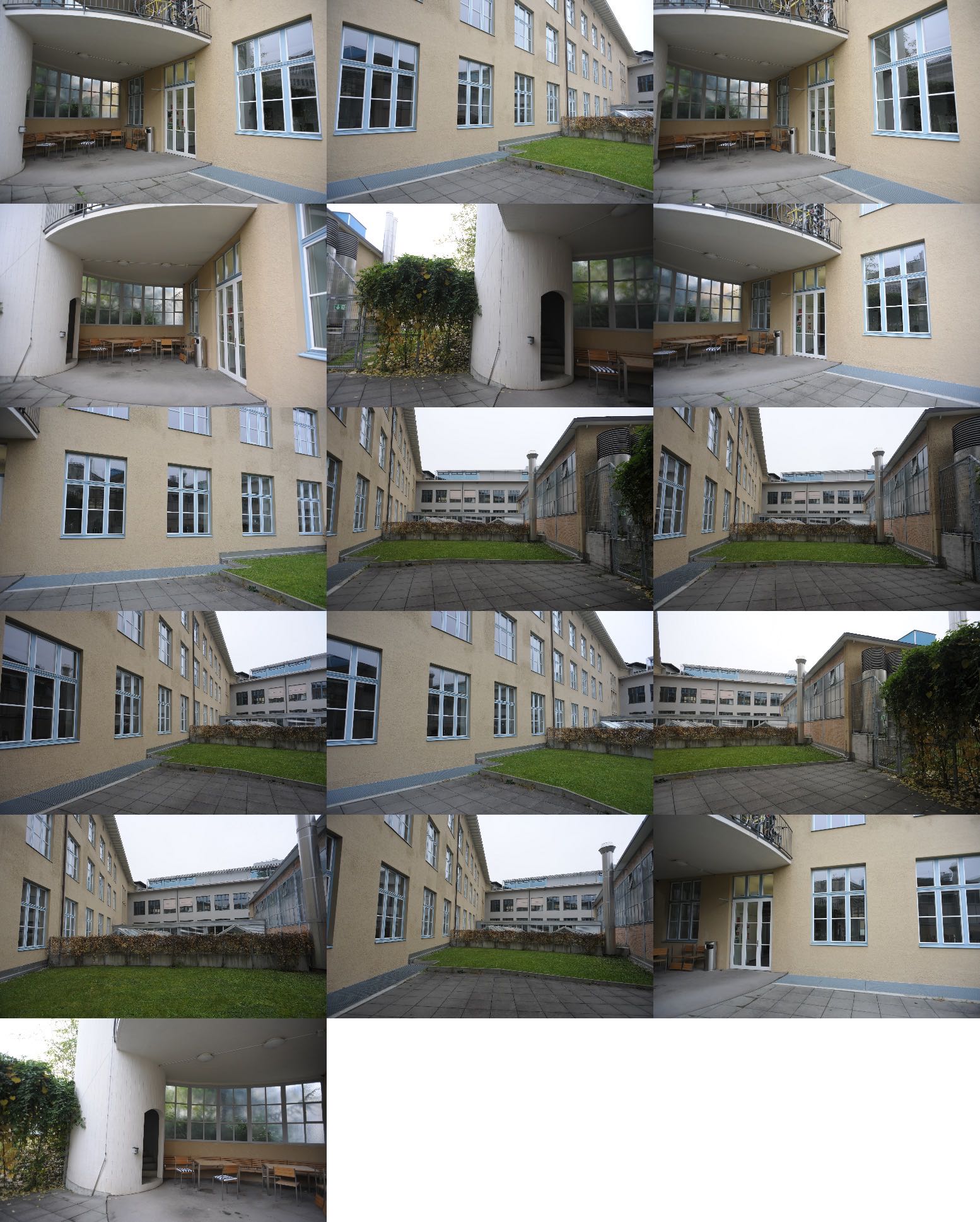}{0.19}{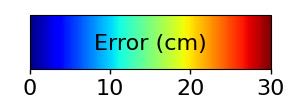}{0.3}
\caption{\textbf{Visual Results on ETH3D~\cite{schoeps2017eth3d}.} We present error maps for all baselines alongside our refined predictions and report the AUC@10 cm (\%) for each scene. Points with confidence below the 10 \% quantile are discarded. Because the laser ground truth is incomplete, some pixels in the error maps have no valid values.}
\label{fig:supp_eth3d}
\end{figure*}

\begin{figure*}[p]   
\centering
\ShowComparisonb{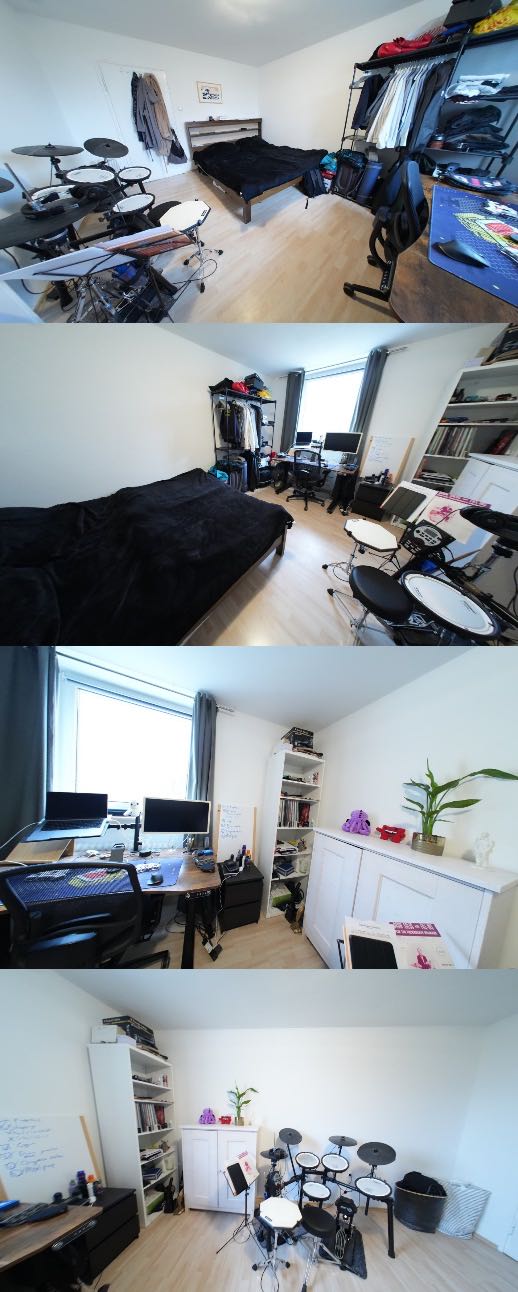}{0.19}{0.05}
\ShowComparisonb{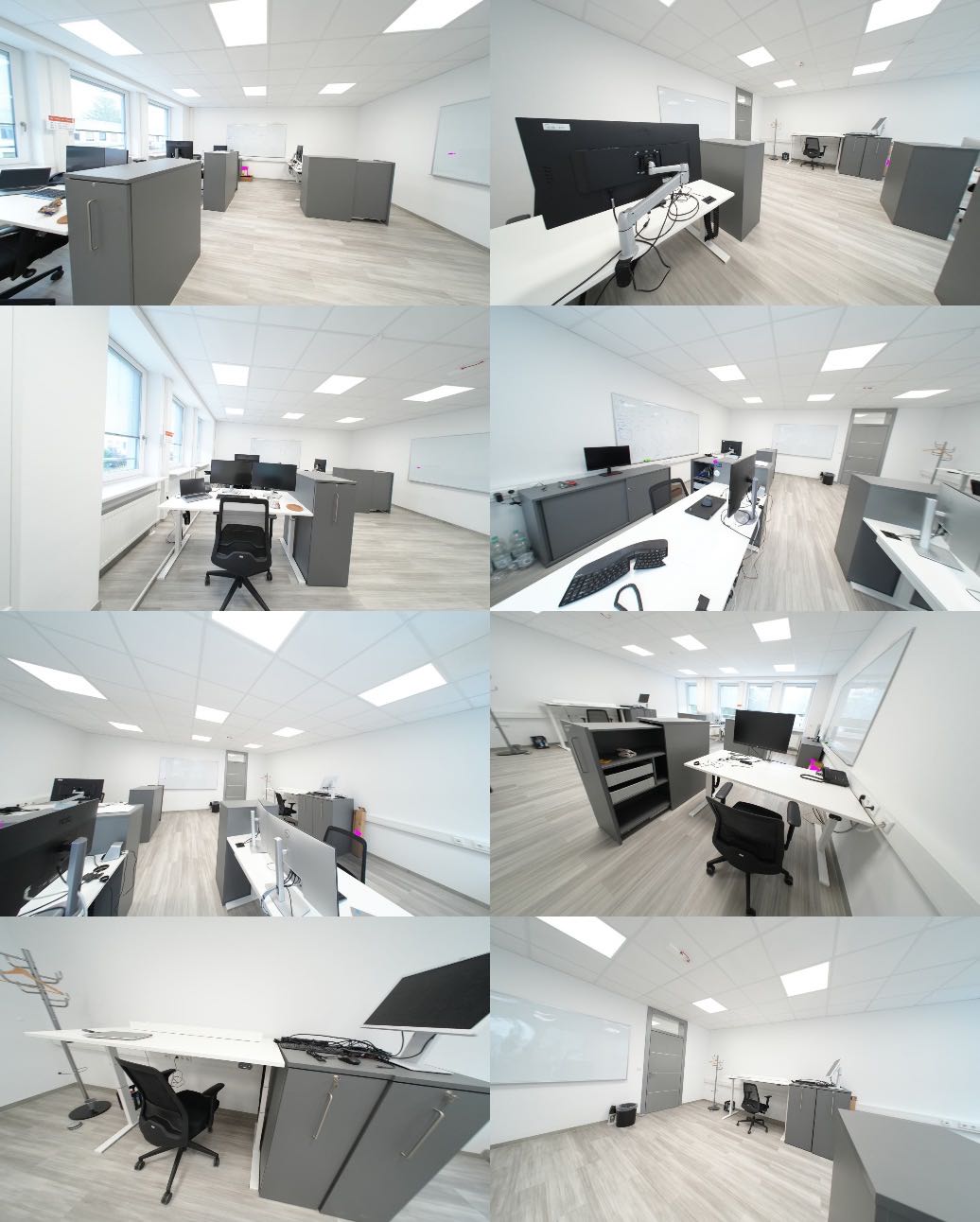}{0.19}{0.05}
\ShowComparisonb{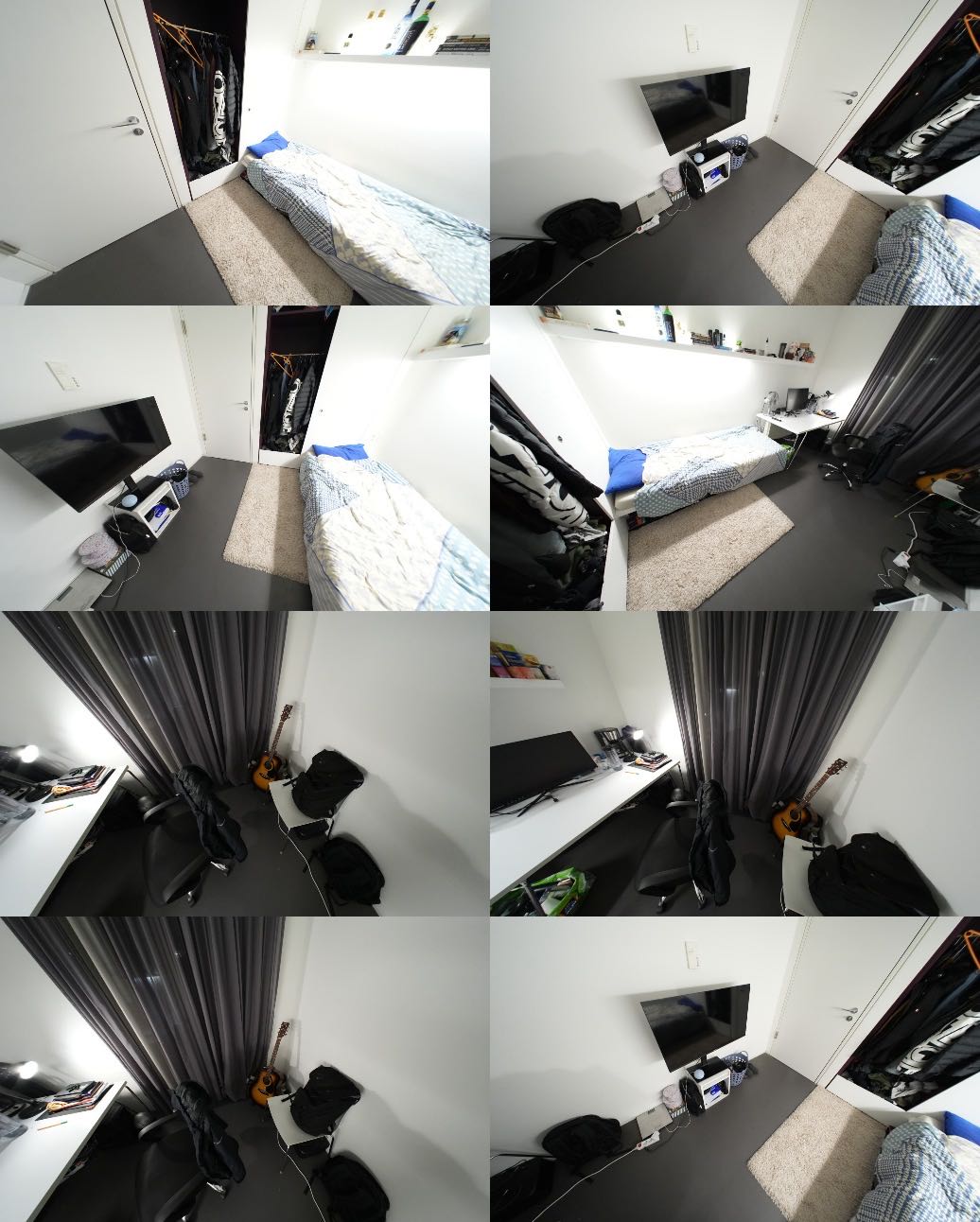}{0.19}{0.05}
\ShowComparisonc{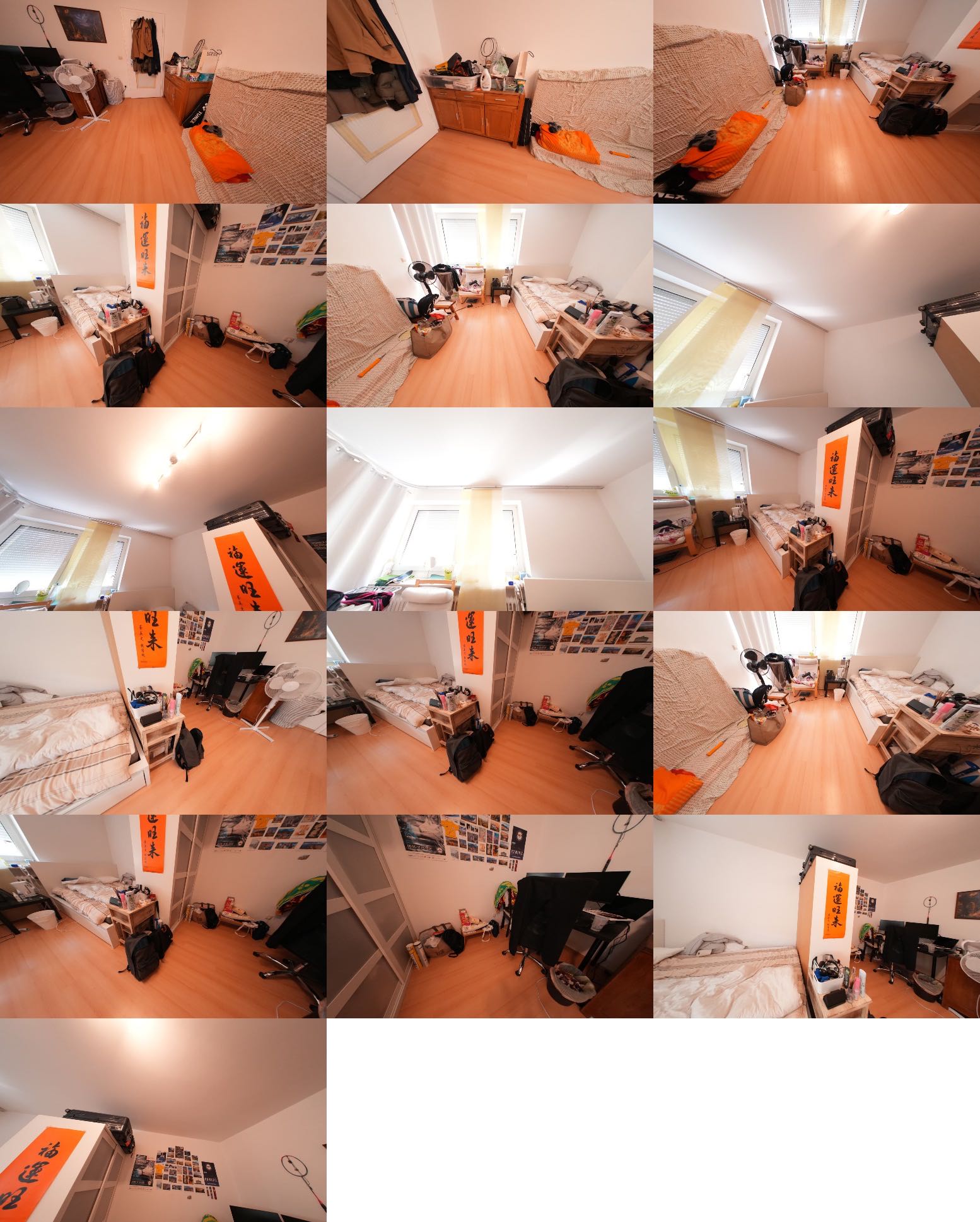}{0.19}{3datasets}{0.05}
\vspace{-3ex}
\caption{\textbf{Visual Results on ScanNet++~\cite{yeshwanth2023scannet++}.} We compare the error maps between baselines and our refinement and report the AUC@10\,cm (\%) for each scene. We filter points with predicted confidence below the 10\% quantile. On the within-domain ScanNet++, Pi3~\cite{wang2025pi3} achieves saturated performance as it is trained on a large scale of similar indoor scenes, leaving limited room for further improvement. Nevertheless, our method still enhances its performance on other datasets, as demonstrated in the accompanying figures.
}
\label{fig:supp_scannetpp}
\end{figure*}

\begin{figure*}[p]   
\centering
\ShowComparisonb{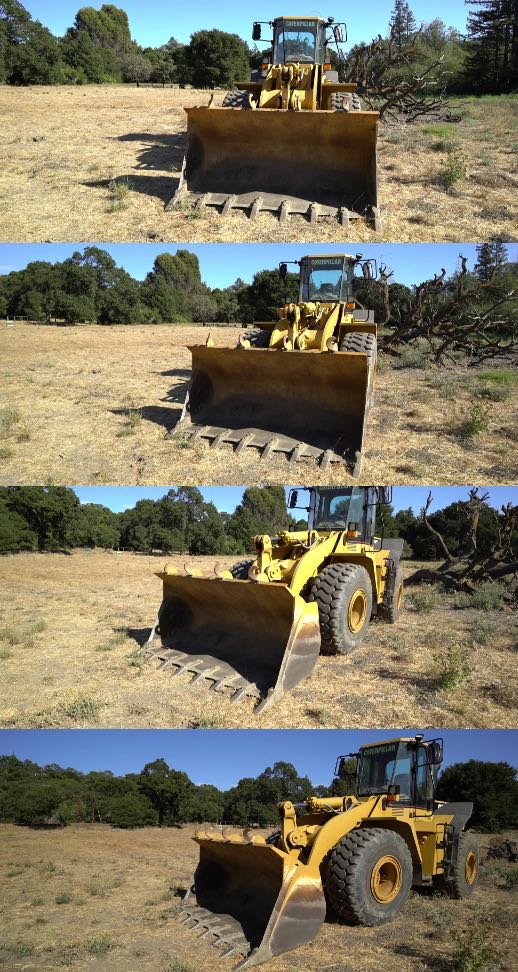}{0.19}{0.05}
\vspace{2ex}
\ShowComparisonb{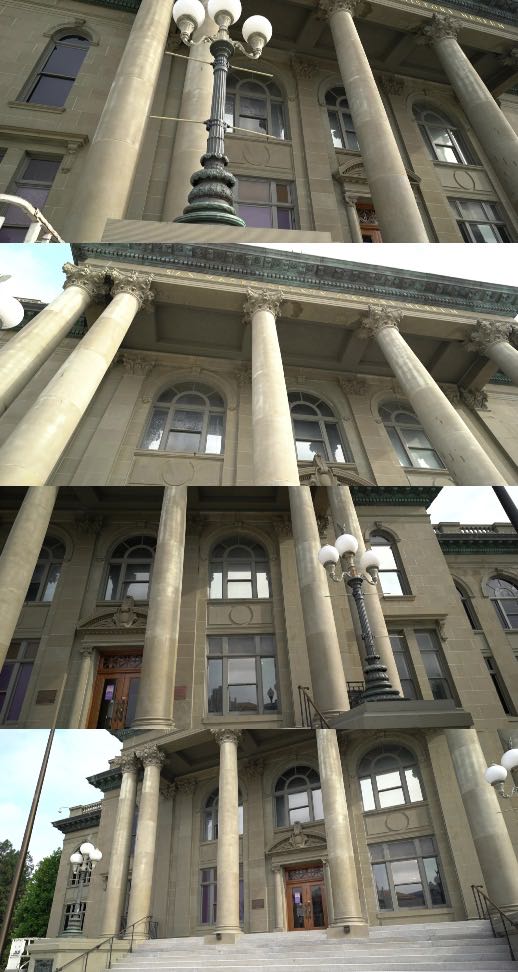}{0.19}{0.05}
\vspace{2ex}
\ShowComparisonb{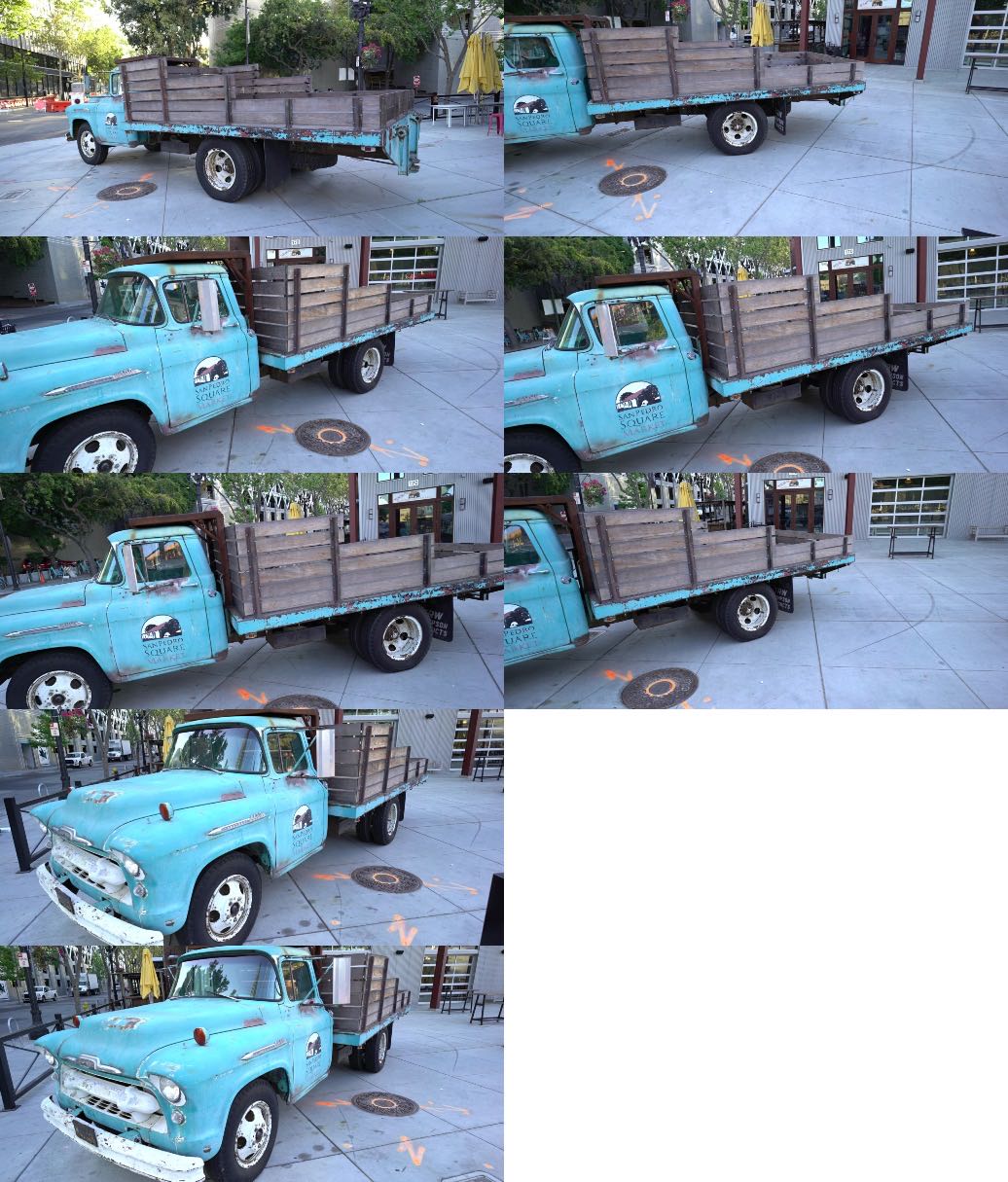}{0.17}{0.05}
\vspace{2ex}
\ShowComparisonc{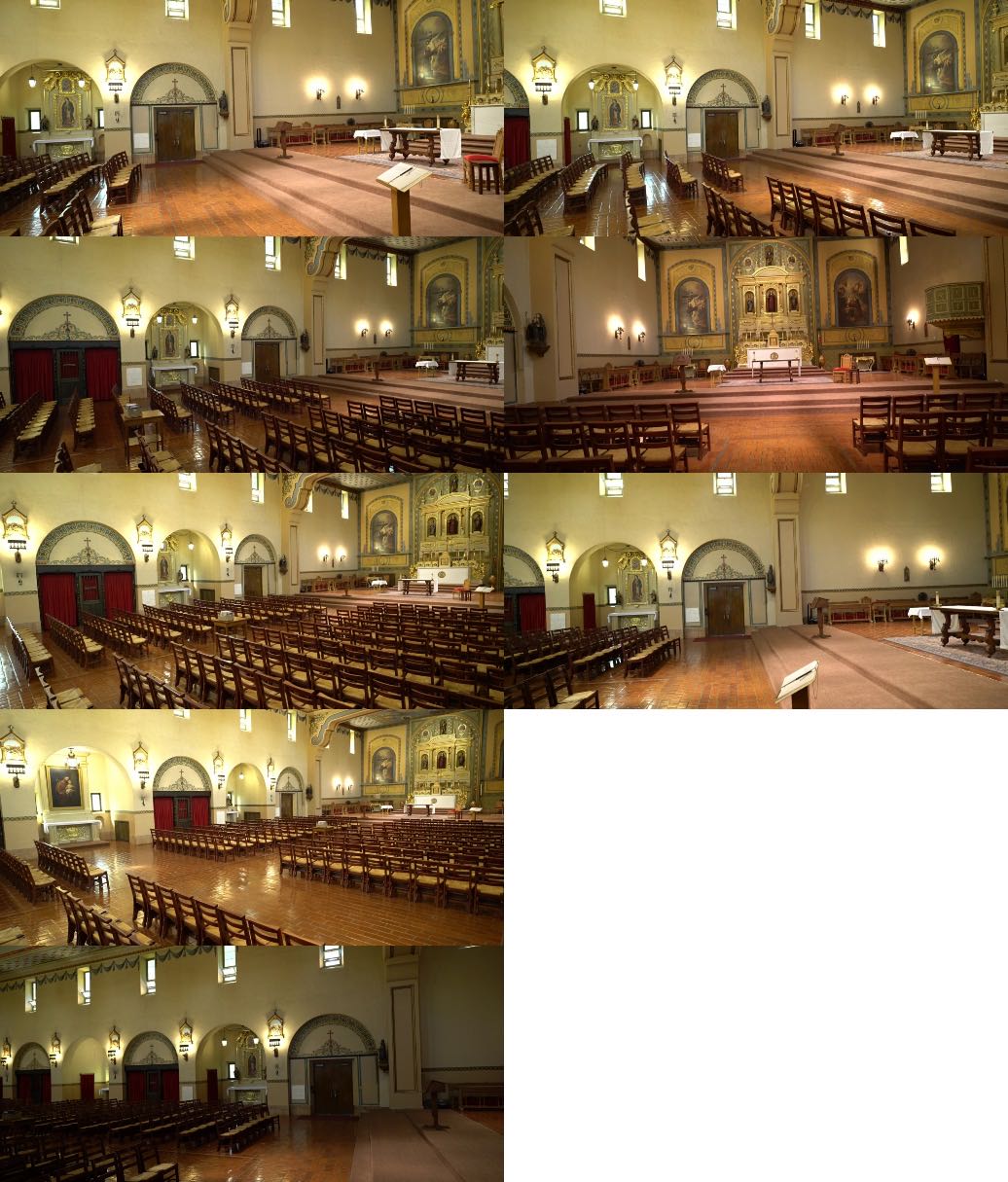}{0.17}{3datasets}{0.05}
\caption{\textbf{More Visual Results on T\&T~\cite{schoeps2017eth3d}.} We compare the error maps of the baselines with those of our refinement and report the AUC@10\,cm (\%) for each scene. Points with predicted confidence below the 10\% quantile are removed. As the laser ground truth does not cover all pixels, some regions in the error maps remain unlabelled.}
\label{fig:supp_tandt}
\end{figure*}

\begin{figure*}[p]   
\centering
\ShowComparisonb{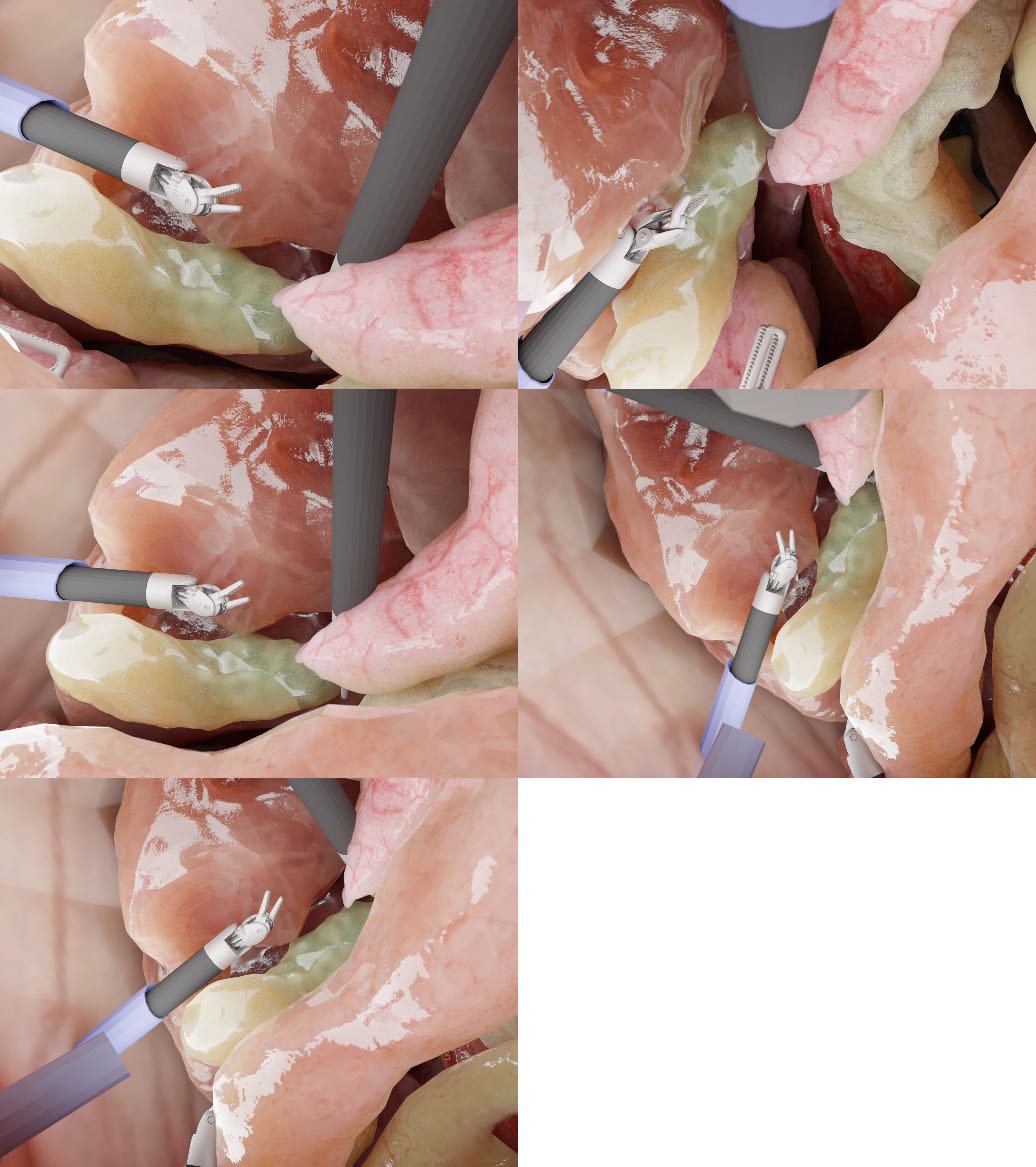}{0.16}{0.05}
\vspace{2ex}

\ShowComparisonb{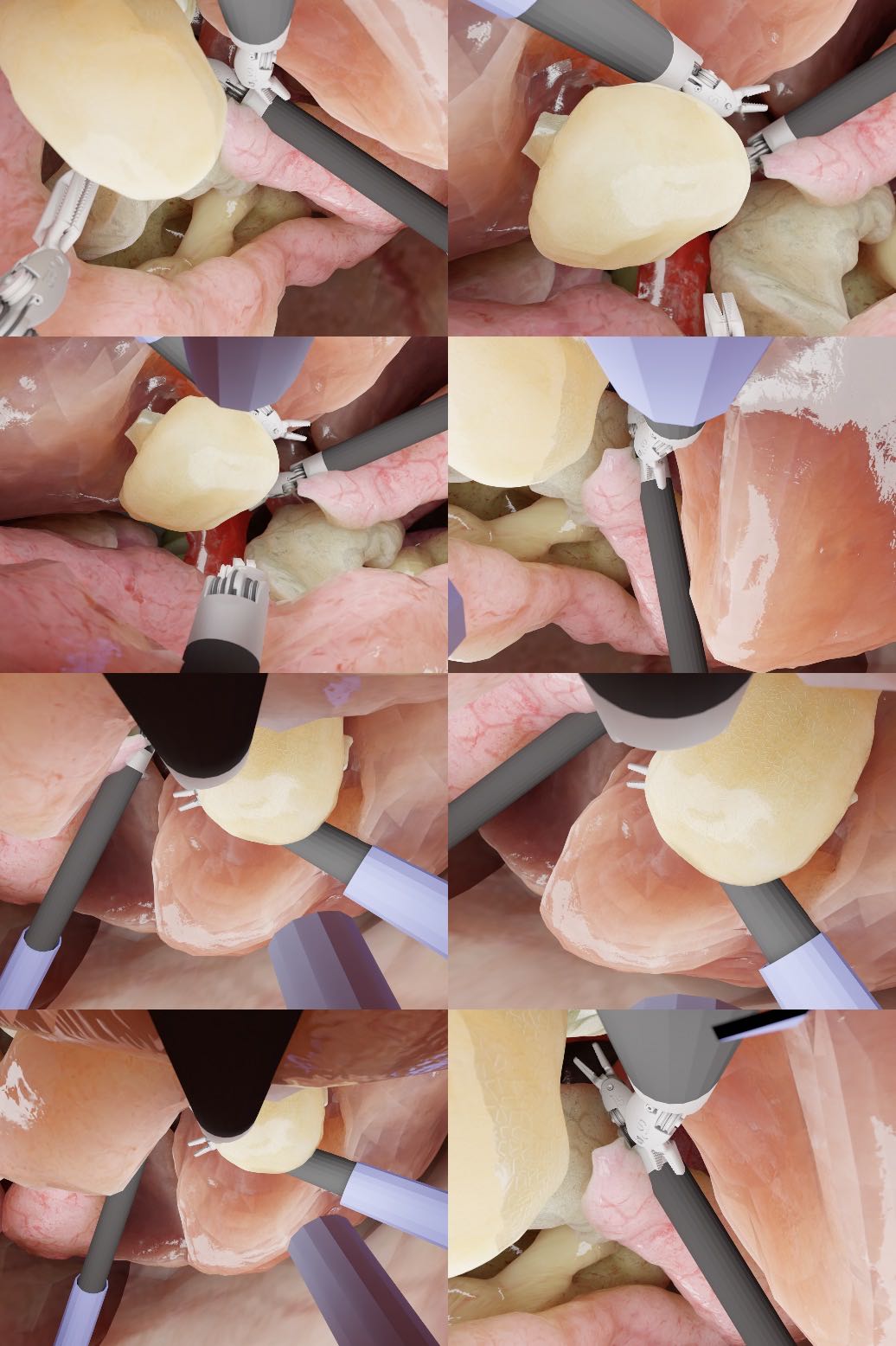}{0.19}{0.05}
\vspace{2ex}
\ShowComparisonc{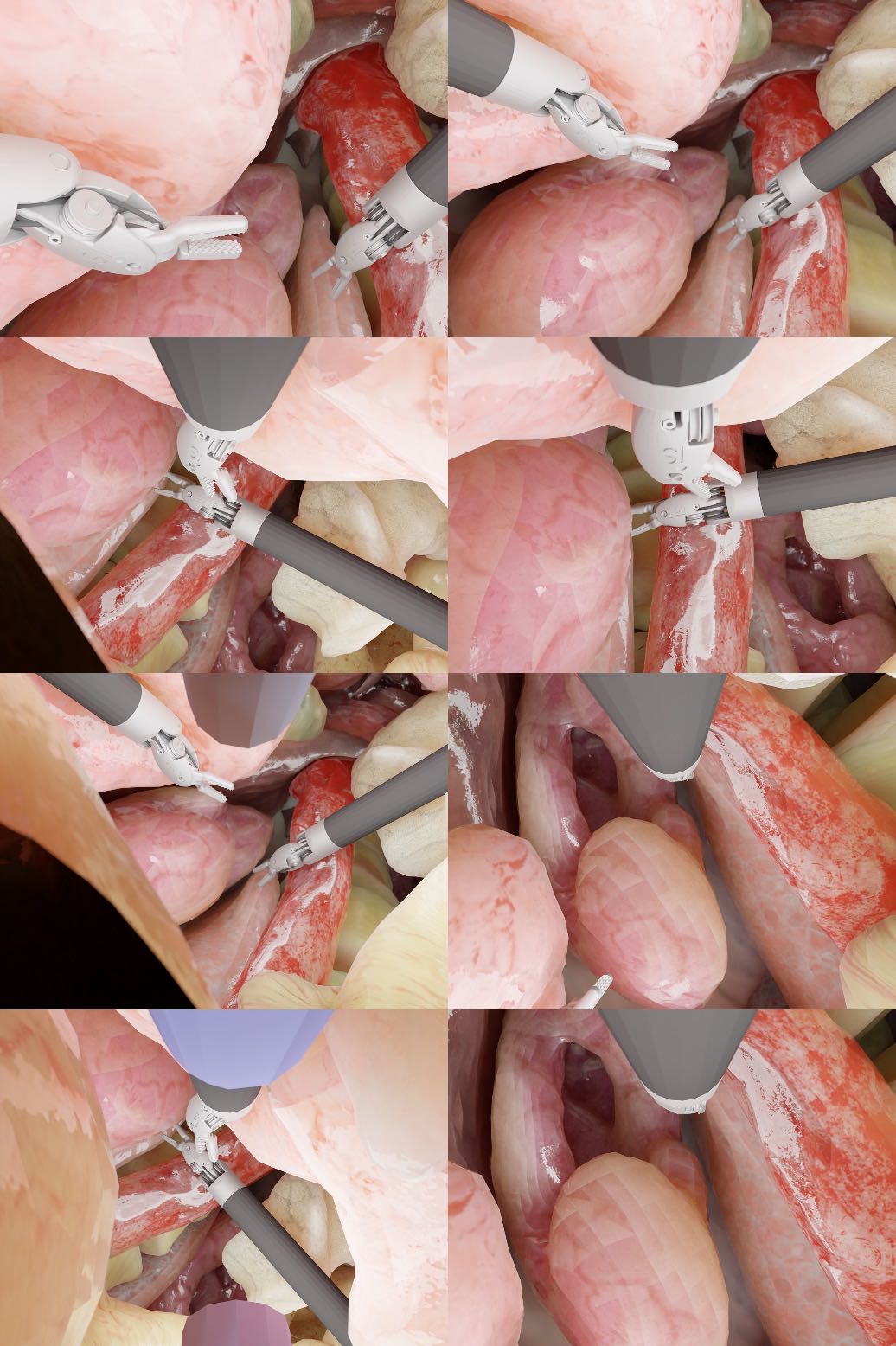}{0.19}{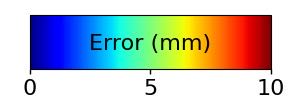}{0.05}

\caption{\textbf{More Visual Results on MV-dVRK~\cite{dvrk-smv}.} We compare the error maps of the baselines with those of our refinement and report the AUC@5\,mm (\%) for each scene. Points with predicted confidence below the 10\% quantile are discarded.}
\label{fig:supp_mvdvrk}
\end{figure*}

\begin{figure*}[p]   
\centering
\ShowComparisonb{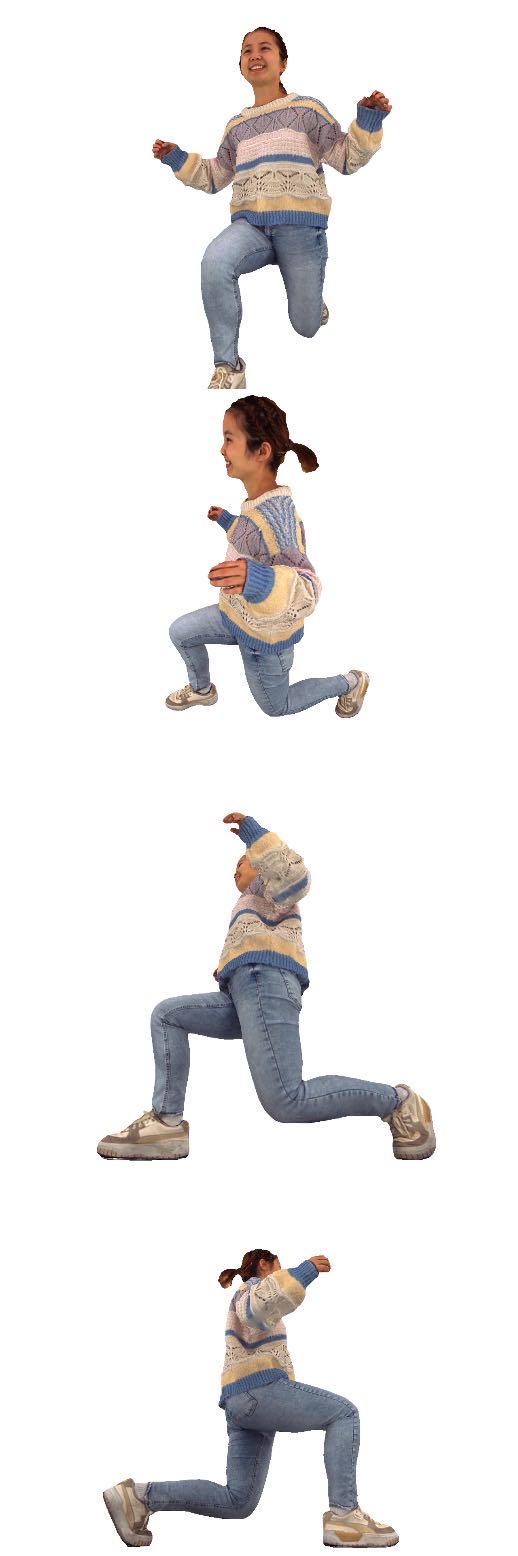}{0.22}{0.05}
\vspace{1em}
\ShowComparisonb{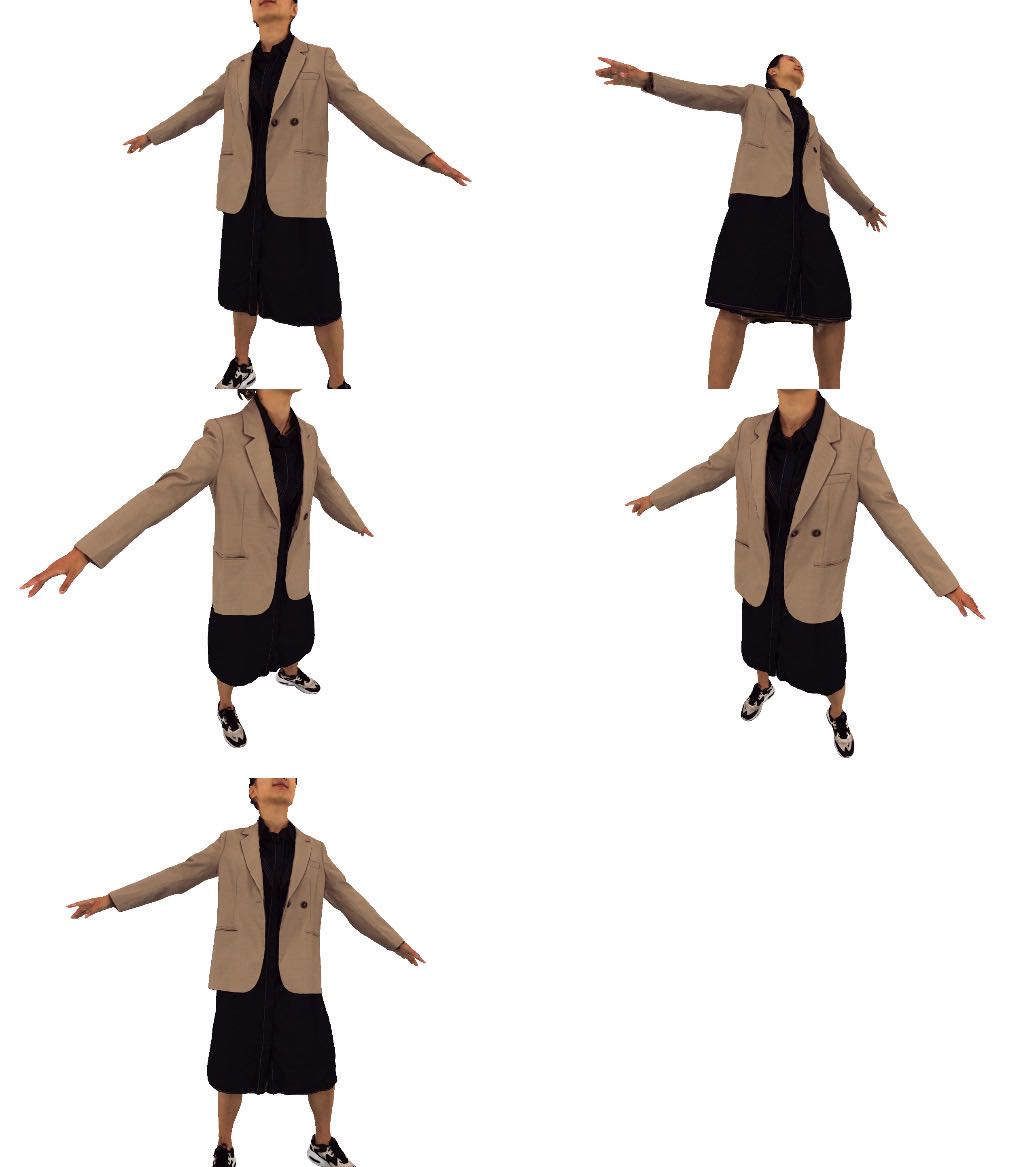}{0.2}{0.05}
\vspace{1em}
\ShowComparisonc{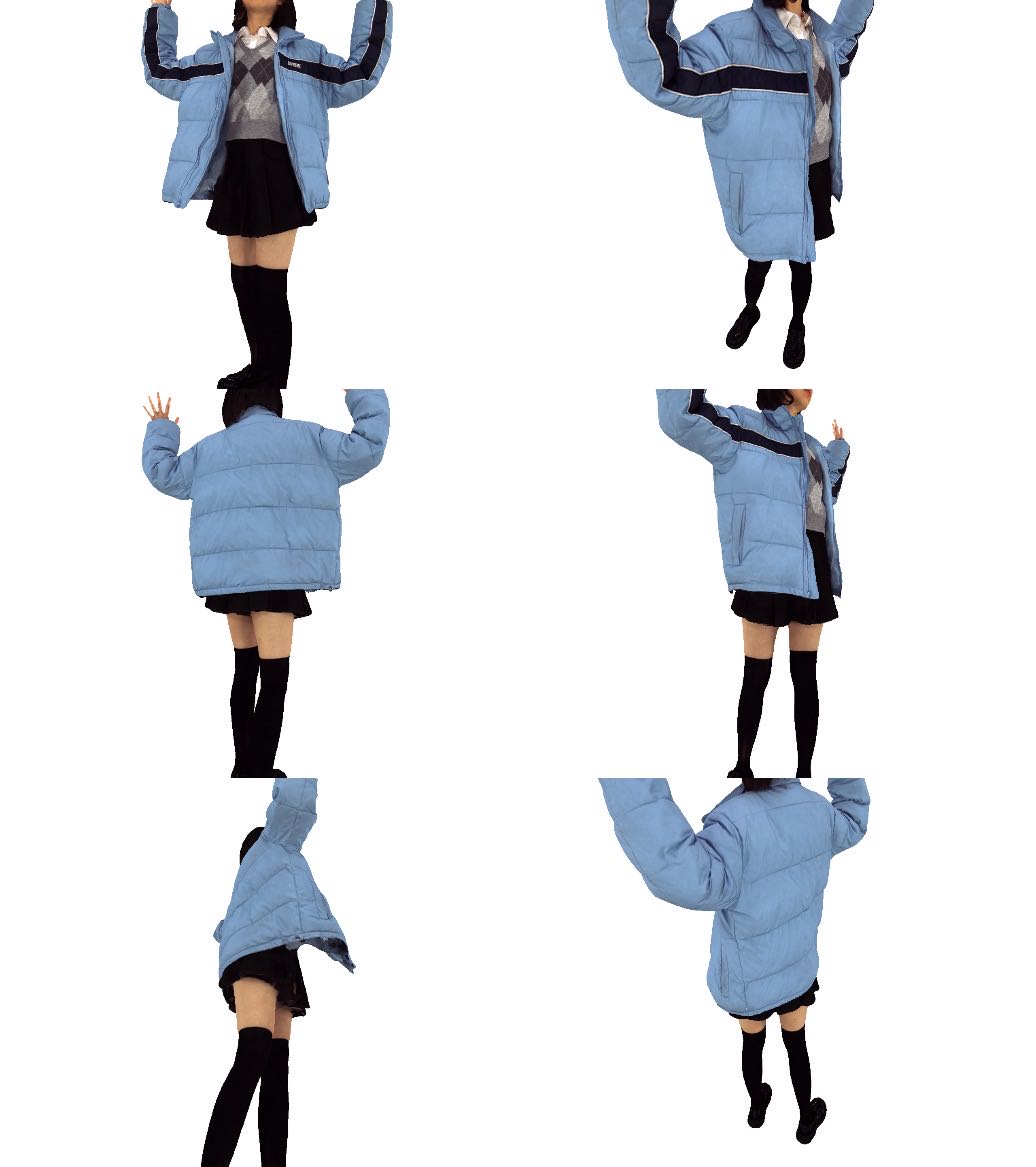}{0.2}{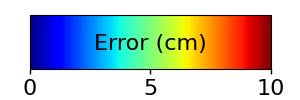}{0.05}

\caption{\textbf{More Visual Results on 4D\text{-}DRESS~\cite{wang20244ddress}.} We compare the error maps of the baselines with those of our refinement and report the AUC@5\,cm (\%) for each scene. Points with predicted confidence below the 10\% quantile are discarded.}
\label{fig:supp_4ddress}
\end{figure*}

\end{document}